\documentclass[11pt]{article}

\usepackage{mathrsfs,amsmath,amsfonts,amssymb,dsfont,amscd,
amsthm,stmaryrd,euscript,color,xcolor} %bm,bbm, pifont,
\usepackage{algorithmic}
\usepackage{algorithm}
\usepackage{url}
\usepackage{mathtools}
\usepackage[english]{babel}
\usepackage{changes}
\usepackage{xstring}

\usepackage{hyperref}  %% wants to come last...[final]

\newlength{\myleftmargin}
\usepackage{amsmath}
\usepackage{amssymb}
\usepackage{amsthm}
\DeclareSymbolFontAlphabet{\Bbb}{AMSb}

\newcommand{\ada}[1]{\emph{#1).}}

\newenvironment{proofof}[1]{\noindent{\bf Proof of #1:}}{\qed\medskip}

\newcommand{\ca}[1]{{\cal #1}}
\newcommand{\frk}[1]{\mathfrak #1}

\newcommand{\Ex}{\mathbb{E}}

\newcommand{\Z}{\mathbb{Z}}
\newcommand{\R}{\mathbb{R}}
\newcommand{\Rd}{{\mathbb{R}^d}}

\newcommand{\E}{\mathbb{E}}
\newcommand{\Px}{P_X}

\renewcommand{\a}{\alpha}

\newcommand{\g}{\gamma}

\newcommand{\D}{\Delta}
\newcommand{\e}{\varepsilon}
\newcommand{\eps}{\epsilon}

\newcommand{\vt}{\vartheta}

\newcommand{\lb}{\lambda}

\DeclareMathOperator{\id}{id}

\newcommand{\Leq}{\,\,\leq\,\,}

\newcommand{\RP}[2]{{{\cal R}_{#1,P}(#2)}}
\newcommand{\RPB}[1]{{{\cal R}_{#1,P}^{*}}}

\newcommand{\RD}[2]{{{\cal R}_{#1,D}(#2)}}

\newcommand{\RPxB}[2]{{{\cal R}_{#1,P,#2}^*}}

\newcommand{\RxB}[2]{{{\cal R}^*_{#1,#2}}}

\newcommand{\fpb}{{f_{L,P}^*}}
\newcommand{\fpd}{{f_{L,P}^\dagger}}

\newcommand{\snorm}[1]{\Vert #1 \Vert}

\newcommand{\inorm}[1]{\Vert #1 \Vert_\infty}

\newcommand{\Lx}[2]{{L_{#1}(#2)}}

\newcommand{\fO}{{f_0}}

\newcommand{\sL}[2]{\ca L_{#1}(#2)}

\newcommand{\relu}[1]{|#1|_+}

\newcommand{\Brelu}[1]{\Bigl|#1\Bigr|_+}

\newcommand{\archx}[1]{\ca A_{#1}}

\newcommand{\potmset}[1]{\mathrm {Pot}_m(#1)}
\newcommand{\potm}{\potmset X}
\DeclareMathOperator{\Image}{Im}

\newcommand{\holdernorm}[1]{|#1|_\alpha}

\newcommand{\HA}{\ca H_{\ca A}}

\newcommand{\markone}[1]{\dot{#1}} 
\newcommand{\marktwo}[1]{\ddot{#1}}

\makeatletter
\def\moverlay{\mathpalette\mov@rlay}
\def\mov@rlay#1#2{\leavevmode\vtop{%
   \baselineskip\z@skip \lineskiplimit-\maxdimen
   \ialign{\hfil$\m@th#1##$\hfil\cr#2\crcr}}}
\newcommand{\charfusion}[3][\mathord]{
    #1{\ifx#1\mathop\vphantom{#2}\fi
        \mathpalette\mov@rlay{#2\cr#3}
      }
    \ifx#1\mathop\expandafter\displaylimits\fi}
\makeatother

\newtheorem{theorem}{Theorem}[section]
\newtheorem{definition}[theorem]{Definition}
\newtheorem{lemma}[theorem]{Lemma}
\newtheorem{proposition}[theorem]{Proposition}
\newtheorem{corollary}[theorem]{Corollary}

\newtheorem{assumption}[theorem]{Assumption}

\newtheorem{example}[theorem]{Example}

\newcommand{\gh}{f_{D,s,\rho}^+}             % gh = good histo
\newcommand{\ghn}{f_{D,s_n, \rho_n}^+}    

\newcommand{\ghon}{f_{D,s_n,0}^+}

\newcommand{\bh}{f_{D,s,\rho}^-}            % bh = bad histo
\newcommand{\bhn}{f_{D,s_n,\rho_n}^-}    
  
\newcommand{\bhon}{f_{D,s_n,0}^-}

\newcommand{\gdn}{g_{D,s,\rho}^+}             % gdn = good dnn
\newcommand{\gdnn}{g_{D,s_n, \rho_n}^+}

\newcommand{\bdn}{g_{D,s,\rho}^-}             % bdn = bad dnn
\newcommand{\bdnn}{g_{D,s_n, \rho_n}^-}

\newcommand{\gbdnn}{g_{D,s_n, \rho_n}^{\pm}}

\usepackage{colortbl, color, graphicx}
\usepackage{dsfont}

\definecolor{nic}{rgb}{0.6, 0.4, 1.0} %{0.0,0.0,0.55}
\definecolor{red}{rgb}{0.75,0.0,0.0}
\definecolor{green}{rgb}{0.0,0.2,0.0}

\newcommand{\mtc}{\mathcal}

\usepackage{bm}
\newcommand{\eins}{\bm{1}}

\newcommand{\paren}[1]{ \left( #1 \right)  }

\newcommand{\norm}[2][a]{%
\IfEqCase{#1}{%
{a}{\left\lVert#2\right\rVert}%
{0}{\lVert#2\rVert}%
{1}{\big\lVert#2\big\rVert}%
{2}{\Big\lVert#2\Big\rVert}%
{3}{\bigg\lVert#2\bigg\rVert}%
{4}{\Bigg\lVert#2\Bigg\rVert}%
}[\PackageError{norm}{Undefined option to norm: #1}{}]%
}

\newcommand{\inner}[2][a]{%
\IfEqCase{#1}{%
{a}{\left\langle#2\right\rangle}%
{0}{\langle#2\rangle}%
{1}{\big\langle#2\big\rangle}%
{2}{\Big\langle#2\Big\rangle}%
{3}{\bigg\langle#2\bigg\rangle}%
{4}{\Bigg\langle#2\Bigg\rangle}%
}[\PackageError{inner}{Undefined option to inner: #1}{}]%
}

\newcommand{\Lc}{L}

\def\cA{{\mtc{A}}}
\def\cB{{\mtc{B}}}
\def\cC{{\mtc{C}}}

\def\cF{{\mtc{F}}}

\def\cH{{\mtc{H}}}

\def\cL{{\mtc{L}}}

\def\cN{{\mtc{N}}}
\def\cO{{\mtc{O}}}
\def\cP{{\mtc{P}}}

\def\cR{{\mtc{R}}}

\def\cX{\mathcal{X}}

\newcommand{\mbr}{\mathbb{R}}

\newcommand{\mbn}{\mathbb{N}}

\newcommand{\no}{ \nonumber}

\newcounter{nbnotes}
\makeatletter
\newcommand{\checknbnotes}{
\ifnum \thenbnotes > 0
\@latex@warning@no@line{**********************************************************************}
\@latex@warning@no@line{* The document contains \thenbnotes \space  note(s)}
\@latex@warning@no@line{**********************************************************************}
\fi}

\makeatother

\begin{document}
\title{Empirical Risk Minimization in the Interpolating Regime with Application to Neural Network Learning}
\author{Nicole M\"ucke  \\
Institute for Mathematical Stochastics\\
Technical University  of Braunschweig \\
\texttt{nicole.muecke@tu-braunschweig.de} 
%\and 
%Jonas Rungenhagen\\
%University of Potsdam \\ 
%\texttt{jrungenh@uni-potsdam.de}
\and 
Ingo Steinwart\\
Institute for Stochastics and Applications\\
University of Stuttgart \\ 
\texttt{ingo.steinwart@mathematik.uni-stuttgart.de}
}
\date{\today}

\maketitle

\begin{abstract}
A common strategy to 
train deep neural networks (DNNs) is to use very large architectures and 
to train them until they (almost) achieve   zero training error. Empirically observed good 
generalization performance on test data, even in the presence of lots of 
label noise, corroborate such a procedure. 
On the other hand, in   statistical learning theory it is 
known that over-fitting models may lead to poor generalization properties, occurring 
in e.g. empirical risk minimization (ERM) over too large hypotheses classes. 
Inspired by this contradictory behavior, so-called interpolation 
methods have recently received much attention, 
leading to consistent and optimally learning methods for some local 
averaging schemes with zero training error. However, there is no theoretical analysis of 
interpolating ERM-like methods so far.  
We take a  step in this direction by showing that for certain, large hypotheses classes, 
some interpolating ERMs enjoy very good statistical guarantees  
while others fail in the worst sense. 
Moreover, we show that the same phenomenon occurs  for  DNNs with zero training error
and  sufficiently large architectures.
\end{abstract}

%%%%%%%%%%%%%%%%%%%%%%%%%%%%%%%%%%%%%%%%%%%%%%%%%%%%%%%%%%%%%%%%%%%%%%%%%%%%%%%%%%%%%%%

%%%%%%%%%%%%%%%%%%%%%%%%%%%%%%%%%%%%%%%%%%%%%%
%% INTRODUCTION
%%%%%%%%%%%%%%%%%%%%%%%%%%%%%%%%%%%%%%%%%%%%%%

\section{Introduction}
\label{sec:intro}

During the last few decades
statistical learning theory (SLT) has developed powerful techniques to
analyze many variants of (regularized) empirical risk minimizers (ERMs), see e.g.~\cite{DeGyLu96,Vapnik98,vandeGeer00,GyKoKrWa02,StCh08,Tsybakov09,SSBD14}.
The resulting learning guarantees, which include finite sample bounds, oracle inequalities,
learning rates, adaptivity, and consistency,  assume in most cases
that the effective hypotheses space of the considered method is sufficiently small in terms of some notion
of capacity such as VC-dimension, fat-shattering dimension, Rademacher complexities, covering numbers, or eigenvalues.

Most training algorithms for 
DNNs also optimize an (regularized) empirical error term over a  hypotheses space, namely the 
 class of functions
that can be represented by the  architecture of the considered DNN, see \cite[Part II]{GoBeCo16}.
However, unlike for many classical ERMs, the  
hypotheses space is 
parametrized in a rather  complicated manner. Consequently, the optimization problem is, in general, harder to solve. 
A common way to address this is in practice is to use very large DNNs, since despite their 
size, training them is often easier, see e.g.~\cite{Salakhutdinov17a,MaBaBe18a} and the references therein. %  \quellen{\dots}.
Now, for sufficiently large DNNs it has been recently observed that 
common training algorithms  can achieve zero training error on randomly, or arbitrarily  labeled 
training sets, see \cite{ZhBeHaReVi16a}.
Because of this ability, their effective hypotheses space can no longer have 
a sufficiently small capacity in the sense of classical SLT, so that the usual 
techniques for analyzing learning algorithms are no longer suitable, 
see e.g.~the discussion on this in \cite{ZhBeHaReVi16a,BeHsMi18a, nagarajan2019uniform, zhou2020uniform}. 
In fact, SLT provides  well known 
examples of large hypotheses spaces  for which zero training error is possible but 
a simple ERM fails to learn. This phenomenon is  known as  over-fitting,
and common wisdom suggests that successful learning algorithms need to avoid over-fitting, see e.g.~\cite[pp.~21-22]{GyKoKrWa02}.
The empirical evidence mentioned above thus stands in stark contrast to this credo of SLT.

%\note{maybe here more/ other recent results: 

This somewhat paradoxical behavior has recently sparked interests, leading to deeper theoretical investigations of 
the so called double/ multiple-descent phenomenon for different model settings. More specifically, \cite{belkin2020two} analyzed linear regression with 
random feature selection and investigated the random Fourier feature model. This model has also been analyzed by  \cite{mei2019generalization}. 
For linear regression, where model complexity is measured in terms of the number of parameters, the authors in  \cite{bartlett2020benign, tsigler2020benign} 
show that over-parameterization is even essential for benign over-fitting. 
However, these results are highly distribution dependent and require a specific covariance structure and (sub-) Gaussian data. 
For more details we refer also to \cite{BeHsMi18a, chen2020multiple, liang2020multiple, neyshabur2018towards, allen2018learning}. 
Another line of research \cite{belkin2019does} shows for classical learning methods, namely Nadaraya-Watson  estimator with certain singular kernels,
that interpolating the training data can achieve optimal rates for problems of nonparametric
regression and prediction with square loss. Beyond empirical evidence there are therefore also theoretical results showing that
interpolating the data and good learning performance is simultaneously possible. 
So far, however, the considered interpolating learning methods do not implement an empirical risk
minimization (ERM) scheme nor do they closely resemble the learning mechanisms of DNNs. 
In this paper, 
we take a step towards closing this gap. 
%}

First, we explicitly construct, for data sets of size $n$, 
large classes of hypotheses 
$\cH_n$
for which we show that some interpolating least squares ERM algorithms over $\cH_n$ enjoy very good 
statistical guarantees, while other interpolating least squares ERM algorithms over $\cH_n$ fail 
in a   strong  sense. To be more precise,  we observe the following phenomena: There exists a universally consistent 
ERM and there exists an ERM whose predictors converge to the negative regression function
for most distributions.  In particular, the latter ERM is not consistent for most distributions, and even worse, the 
obtained risks are usually for off the best possible risk. 
We further construct modifications that enjoy minmax optimal rates of convergence up to some log factor under standard assumptions.    
In addition, there are also ERM algorithms that exhibit  
an intermediate behavior between these two extreme cases, with arbitrarily slow convergence. 
To put this in perspective, we note that 
classical SLT shows that for sufficiently small  hypotheses classes, \emph{all versions} 
of ERM enjoy good statistical guarantees. In contrast, our results demonstrate that this is no longer true for large hypotheses
classes. For such hypotheses spaces, the description ``ERM'' is thus
not sufficient to identify well-behaving learning algorithms. 
Instead, the class of algorithms described by ``ERM'' over such hypotheses spaces may encompass
learning algorithms with extremely distinct learning behavior.

Second, we show that exactly the same phenomena occur for interpolating ReLU-DNNs
of at least two hidden layers with widths growing linearly in both input dimension $d$ 
and sample size $n$. We present  DNN training algorithms that produce interpolating predictors and that 
enjoy consistency and optimal 
rates, at least up to some log factor. In addition, this
training  can be done in $\cO(d^2\cdot n^2)$-time if the DNNs are implemented as fully connected networks. 
Since the constructed predictors have a particularly sparse structure, the   
training time can actually be reduced to $\cO(d\cdot n \cdot \log n)$ by implementing the DNNs as loosely connected networks.  
Moreover, we show that there are other efficient 
and feasible training algorithms for exactly the same architectures that fail in the 
worst possible sense, and like in the ERM case, there are also a variety of training 
algorithms performing in between these two extreme cases.

The rest of the paper is organized as follows: In Section \ref{sec:classical-histo} we firstly recall classical histograms as ERMs 
that we extend then to the class of inflated histograms. We provide specific examples of interpolating predictors from that class. 
In our main theorems we derive consistency results and learning rates. In the following Section \ref{sec:ReLU-Approx} we explain how inflated histograms 
can be approximated by ReLU networks, having analogous learning properties. 

All our proofs are deferred to the Appendices \ref{app:general}, \ref{app:technical-stuff}, \ref{app-histo-consistency-proofs}, and \ref{app-dnn-proofs}. 
Finally, in the supplementary material \ref{app:histo-random} we derive general uniform bounds for histograms based on data-dependent partitions.  
This result is needed for proving our main results and is of independent interest.

%In Section \ref{sec:results} we present our main results and we discuss their consequences.  
%Section \ref{sec:prelim-histo} is devoted to constructing statistically good and bad interpolating predictors. %recalling the basic concepts about histograms 
%In Section \ref{sec:ReLU-Approx}, a similar construction is derived for DNNs.  
%All proofs are deferred to the appendices. 

%%%%%%%%%%%%%%%%%%%%%%%%%%%%%%%%%%%%%%%%%%%%%%
%% HISTOGRAM PART
%%%%%%%%%%%%%%%%%%%%%%%%%%%%%%%%%%%%%%%%%%%%%%

\section{The histogram rule revisited}
\label{sec:classical-histo}

In this section we reconsider the  histogram rule in the framework of  regression.  In more detail, we recall the \emph{classical} histogram rule 
and show how to change this appropriately in order to obtain a predictor that is able to \emph{interpolate} given data. 
To this end, let us begin by 
introducing the necessary notations. 
Throughout this work, we consider $X:=[-1,1]^d$ and $Y=[-1,1]$ 
if not specified otherwise. Moreover,   $L:Y\times \R\to [0,\infty)$ denotes 
the least squares loss $L(y,t)=(y-t)^2$. 
Given a  dataset $D:= ((x_1, y_1),...,(x_n, y_n)) \in (X\times Y)^n$ 
drawn i.i.d.~from an unknown distribution $P$ on $X \times Y$,
the aim of supervised learning is to 
build a function $f_D: X \to \R$ based on $D$ 
such that its \emph{risk} 
\begin{equation}\label{eq:risk}
\RP{L}{f_D} := \int_{X \times Y} L(y, f_D(x)) \; dP(x,y)  \;, 
\end{equation}
is close to the smallest possible 
 risk 
\begin{equation}  \label{Brisk}
\RxB{L}{P} = \inf_{f:X \to \mbr}  \RP{L}{f}   \, .
\end{equation} 
In the following, $\RxB{L}{P}$  
is called the \emph{Bayes risk} and an  $\fpb: X \to \mbr$ satisfying 
$\RP{L}{f^*_{P}} = \RxB{L}{P}$
is called \emph{Bayes decision function}. 
Recall, that for the least squares loss, $\fpb$ equals the conditional mean  
function, i.e.~$\fpb(x) = \E_P(Y|x)$ for $P_X$-almost all $x\in X$, 
where $P_X$ denotes the marginal distribution of $P$ on $X$. 
%{\notice 
In general, estimators  $f_D$ having small \emph{excess risk} 
\begin{equation}
\label{eq:excess-risk}
  \RP{L}{f_D} - \RxB{L}{P} = ||f_D - \fpb||_{L_2(P_X)}^2  \;,
\end{equation}  
where $\snorm{\cdot}_{L_2(P_X)}$ denotes the usual $L_2$-norm with respect to $P_X$, 
are considered as \emph{good} in classical statistical learning theory.

Now, to describe the class of learning algorithms we are interested in, we need
 the \emph{empirical risk} of an $f:X\to \R$, i.e.
\[  \RD{L}{f} := \frac{1}{n}\sum_{i=1}^n L( y_i, f(x_i))  \;.  \] 
Recall, that an \emph{empirical risk minimizer} (ERM) over some set $\ca F$ of 
functions $f:X\to \R$  chooses, for every data set $D$,
an $f_D \in \cF$ that satisfies 
\[ 
 \RD{L}{f_D} = \inf_{f \in \cF}\RD{L}{f} \, . 
\]
Note that the definition of ERMs implicitly requires that 
the infimum on the right hand side is  attained, namely by $f_D$. 
In general,  however,  $f_D$ does not need to be unique.
It is well-known that if we have a suitably increasing  sequence of hypotheses classes
$\cF_n$ with controlled capacity, then \emph{every} ERM $D\mapsto f_D$
that ensures $f_D \in \cF_n$ for all data sets $D$ of length $n$ learns
in the sense of e.g.~universal consistency, and under additional assumptions  it may
also enjoy minmax optimal learning rates,  see e.g.~\cite{DeGyLu96,vandeGeer00,GyKoKrWa02,StCh08}.

%%%%%%%%%%%%%%%%%%%%%%%%%%%%%%%%%%%%%%%%%%%%%%%%%%%%%%%%%%%%%%%%%%%%%%%%%%%%%%%%%%%%%%%%%%%%%%%%%%%%%%%%%%%%%%%%%%%%%%%%%%%%%%%%%%%%

\subsection{Classical Histograms}
 
Particular simple ERMs are histogram rules (HRs). To recall the latter, 
 we fix 
a finite partition $\ca A = (A_j)_{j\in J}$ of $X$
and for $x\in X$ we   write $A(x)$ for the 
unique cell $A_j$ with $x\in A_j$.
Moreover, we
define 
\begin{align}\label{hr-hypo-space}
   \cH_{\ca A}  := \biggl\{   \sum_{j\in J}  c_j\eins_{A_j} \; : \; c_j \in Y  \biggr\} \;, 
\end{align}
where $\eins_{A_j}$ denotes the \emph{indicator function} of the cell $A_j$.
Now, given a data set $D$ and a loss $L$ an  \emph{$\ca A$-histogram} 
%\fix{\note{Q: put in definition environment ?}  \noteis{The important ones should be, the others not. See fixboxes later}}
is an 
$h_{D,  \ca A} =   \sum_{j=1}^m  c_j^*\eins_{A_j} \in \cH_{\ca A}$
that satisfies 
\begin{align}\label{hr-coefficient}
  \sum_{i: x_i \in A_j} L(x_i,y_i, c^*_j) = \inf_{c\in Y}  \sum_{i: x_i \in A_j} L(x_i, y_i, c)
\end{align}
for all, so-called \emph{non-empty cells} $A_j$, that is, cells $A_j$ with   $N_j :=|\{i: x_i \in A_j\}| >0$.
Clearly, $D\mapsto h_{D, \ca A}$
is an ERM. Moreover, 
note that in general $h_{D, \ca A}$ is \emph{not uniquely determined}, since 
$c_j^*\in Y$ can take arbitrary values for empty cells $A_j$. In particular, 
\emph{there are more than one ERM over $\cH_{\ca A}$} as soon as $m,n\geq 2$.

Before we proceed, let us consider the specific example of the least squares loss in more detail. 
Here, a simple calculation shows, see Lemma \ref{result:hr-is-erm}, that 
for all non-empty cells $A_j$, the coefficient 
$c_j^*$ in \eqref{hr-coefficient} is uniquely determined by 
\begin{equation}
\label{eq:HRR}
 c_j^* := \frac 1 {N_j}   \sum_{i: x_i \in A_j} y_i \;  
\end{equation} 
provided that $Y$ is convex.
In the following, we call every resulting $D\mapsto h_{D,  \ca A}$ with 
\[ h_{D,  \ca A}: =   \sum_{j=1}^m  c_j^*\eins_{A_j} \in \cH_{\ca A} \] 
an \emph{empirical HR for regression} with respect to the least-squares loss $L$. 
For later use we also introduce an infinite sample version of a classical histogram 
\begin{equation}\label{eq:inf-hist}
 h_{P,\cA}: =  \sum_{j\in J}   c^*_j \eins_{A_j} \;, \qquad\qquad\mbox{ where }\quad  c^*_j := \frac{1}{P_X(A_j)} \int_{A_j}\fpb(x) dP_X(x) \;
\end{equation} 
for all cells $A_j$ with $P_X(A_j)>0$. 
Similarly to empirical histograms one has 
\[ \cR_{L,P}(  h_{P,\cA}) = \inf_{h \in \cH_\cA} \cR_{L,P}(h) \; .\]

We are mostly interested in HRs on $X=[-1,1]^d$ whose underlying partition 
essentially consists of cubes with a fixed width. To rigorously deal with boundary effects, 
 we first say that a partition $(B_j)_{j\geq 1}$
of $\R^d$ is a \emph{cubic partition of width $s>0$}, if each cell $B_j$ is a translated version of $[0,s)^d$, 
i.e.~there is an $x^\dagger \in \R^d$ called \emph{offset} such that 
for all $j\geq 1$ there exist $k_j := (k_1,\dots,k_d)\in \Z^d$ with
\begin{align}\label{cubic-def}
 B_j =  x^\dagger + sk_j + [0,s)^d \, .
\end{align}
Now, a partition $\ca A = (A_j)_{j\in J}$ of $X=[-1,1]^d$ is called a cubic partition of width $s>0$,
if there is a cubic partition $\cB=(B_j)_{j\geq 1}$ of $\R^d$ with width $s>0$ such that 
$J= \{j\geq 1: B_j \cap X \neq \emptyset\}$ and $A_j = B_j\cap X$ for all $j\in J$.
If $s\in (0,1]$, then, up to reordering, this $(B_j)_{j\geq 1}$ is uniquely determined by $\ca A$.

If the hypotheses space \eqref{hr-hypo-space} is based on a 
cubic partition of $X=[-1,1]^d$ with 
width $s>0$, then the resulting HRs are well understood. 
For example, universal consistency and learning rates 
have been established, see e.g.~\cite{DeGyLu96, GyKoKrWa02}.  
In general, these results only require a suitable choice 
for the widths $s=s_n$  for $n\to \infty$ but no specific choice of the cubic 
partition of width $s$. For this reason we write $\ca H_s := \bigcup \ca H_{\ca A}$,
where the union runs over all cubic partitions $\ca A$ of $X$ with fixed width $s\in (0,1]$.

%%%%%%%%%%%%%%%%%%%%%%%%%%%%%%%%%%%%%%%%%%%%%%%%%%%%%%%%%%%%%%%%%%%%%%%%%%%%%%%%%%%%%%%%%%%%%%%%%%%%%%%%%%%%%%%%%%%%%%%%%%%%%%%%%%%%

\subsection{Interpolating Predictors and Inflated Histograms}
\label{sec:inflated-histo}

In this section we construct particular interpolating ERMs. 
In a nutshell, the basic idea is to first consider classical histogram rules, and then to 
inflate their hypotheses space so that we can find interpolating ERMs in these inflated  hypotheses spaces.

\begin{definition}[Interpolating Predictor]
We say that an  
$f:X\to Y$ \emph{interpolates} $D$, if 
\begin{displaymath}
   \RD Lf = \RxB LD := \inf_{\tilde f:X\to \R} \RD L{\tilde f}\, ,
\end{displaymath}
where we emphasize that the infimum is taken 
over all $\R$-valued functions, while  $f$ is required to be $Y$-valued. 
\end{definition}
Clearly, an $f:X\to Y$ interpolates $D$ if and only if
\begin{align}\label{interpol-function}
   \sum_{k: x_k =x_i^*} L(x_i,y_i, f(x_i^*)) = \inf_{c\in \R}  \sum_{k: x_k =x_i^*} L(x_i, y_i, c)\, ,
\qquad\qquad
i=1,\dots,m,
\end{align}
where  $x_1^*,\dots, x_m^*$ are the elements of $D_X := \{x_i: i=1,\dots,n\}$.

It is easy to check that for  the least squares loss $L$
and all data sets $D$ there exists an $f_D^*$ interpolating $D$. 
Moreover, we have $\RxB LD > 0$ if and only if
$D$ contains \emph{contradicting samples}, i.e.~$x_i = x_k$ but $y_i \neq y_k$.
Finally, if $\RxB LD = 0$, then any interpolating $f_D^*$ needs to satisfy 
$f_D^*(x_i) = y_i$ for all $i=1,\dots,n$.

\begin{definition}[Interpolatable Loss]
We say that $L$ is \emph{interpolatable  for $D$}
if there exists an 
 $f:X\to Y$ that \emph{interpolates} $D$, i.e.~$\RD Lf = \RxB LD$. 
\end{definition}

Note that \eqref{interpol-function} in particular ensures that the infimum over $\R$ on the right is 
attained at some $c^*_i\in Y$.
Many common  losses including the
least squares, the hinge, and  the classification loss
interpolate all $D$, and for the latter three losses 
we have $\RxB LD > 0$ if and only if
%\fix{\noteis{Dieser Absatz wiederholt sich etwas mit dem Text for der Definition}} 
$D$ contains \emph{contradicting samples}, i.e.~$x_i = x_k$ but $y_i \neq y_k$.
Moreover, for the least squares loss, $c^*_i$ can be easily computed by averaging over all labels 
$y_k$ that belong to some sample $x_k$ with $x_k = x_i$.

\vspace*{1ex}
Let us now describe more precisely the \emph{inflated versions of $\ca H_s$}. 
For 
  $r,s>0$ and   
$m\geq 0$ we want to consider functions 
\begin{equation}\label{proper}
f =  h  +  \sum_{i=1}^m b_i \eins_{x_i^* + t B_\infty  }
\end{equation}
with $ h\in \ca H_s,\,  b_i \in 2Y ,\,   x_i^* \in X,$ and  $t \in [0,r]$, 
where $B_\infty := [-1,1]^d$. 
In other words, for $m\geq 1$, such an $f$ 
changes a classical histogram $h \in \ca H_s$ 
on at most  $m$ small neighborhoods of some arbitrary points $x_1^*,\dots,x_m^*$ in $X$. Such changes are useful for finding interpolating predictors. In general, 
these small neighborhoods $x_i^* + t B_\infty$ however may intersect and may be contained in more than one 
cell $A_j$ of the considered partition $\ca A$ with $h \in \cH_\cA$. 
To avoid undesired boundary effects we restrict the class of all admissible cubic partitions $\cA$ of $X$ associated with $h$.  
An additional technical difficulty arises in particular when constructing interpolating predictors since the set of points $\{x_1^*, ..., x_m^*\}\subset X$ 
are naturally the random input variables. As a consequence, the admissible cubic partitions become data-dependent. 
As a next step, we introduce the notion of a \emph{partitioning rule}. 
To this end, we write
\begin{align*}
 \potm := \bigl\{A\subset X: |A| = m   \bigr\}
\end{align*}
for the set of all subsets of $X$ having cardinality $m$. Moreover, 
we denote  the set of all finite partitions of $X$ by $\cP(X)$.

\begin{definition}
Given an integer $m\geq 1$, an \emph{$m$-sample partitioning rule for $X$} is a map 
$\pi_m: \potm \to \cP(X)$, i.e.~a map that associates to every subset 
$\{x_1^*, ..., x_m^*\}\subset X$  of cardinality $m$ a finite partition $\ca A$. 
Additionally, we will call an $m$-sample partitioning  
rule that assigns to any such $\{x_1^*, ..., x_m^*\}\in \potm$ 
a cubic partition with fixed width $s \in (0,1]$ an 
\emph{$m$-sample cubic partitioning rule} and write $\pi_{m,s}$. 
\end{definition}

Next we explain in more detail which particular partitions are considered as \emph{admissible}.

\begin{definition}[Proper Alignment]
Let $\cA$ be a cubic partition of $X$ with width $s\in (0,1]$, $\cB$ be the partition of $\mbr^d$ that defines $\ca A$, and $r\in (0,s)$.
We say that $\cA$ is \emph{properly aligned   to the set of points $\{x_1^*, ..., x_m^*\}\in \potm$ with parameter $r$},  if for all $i,k=1,\dots,m$
we have 
\begin{align}\label{proper-1}
    x_i^* + r B_\infty  & \subset B(x_i^*)\, , \\ \label{proper-2}
x_i^* + r B_\infty  \cap x_k^* + r B_\infty  & = \emptyset \qquad \qquad \mbox{ whenever $i\neq k$,}
\end{align}
where $B(x_i^*)$ is the unique cell \footnote{Note that this gives 
$A(x_i^*) = B(x_i^*)\cap X$.} of $\cB$ that contains $x_i^*$. 
\end{definition}

Clearly, if  $\cA$ is properly aligned with parameter $r> 0$, then it is also properly aligned for any parameter $t \in [0, r]$ for the same 
set of points $\{x_j^*\}_{j=1}^m$ in $\potm$. Moreover, any 
cubic partition $\cA$ of $X$ with width $s>0$ is properly aligned with the parameter $r=0$ for any set of points $\{x_j^*\}_{j=1}^m$ in $\potm$.

In what follows, we establish the existence of  cubic partitions $\cA$ that are properly aligned to a given set of points with  
parameter $r>0$ being sufficiently small. In other words,  we construct a special $m$-sample cubic partitioning rule  $\pi_{m,s}$. 
We call  henceforth any such rule $\pi_{m,s}$ an \emph{$m$-sample properly aligned cubic partitioning rule}.   
To this end, let  $D_X:= \{x_1^*, ..., x_m^*\} \in \potm$ be a set points and 
note that \eqref{proper-2} holds for all  $r>0$ satisfying
\[ r<\frac 12\min_{i,k:i\neq k}\inorm{x_i^*-x_k^*} \;. \] 
Clearly, a brute-force algorithm finds such an $r$ in $\ca O(dm^2)$-time. However,
a smarter approach is to first sort the first coordinates $x_{1,1}^*,\dots, x_{m,1}^*$ and to 
determine the smallest \emph{positive} distance $r_1$ of two consecutive, non-identical ordered coordinates.
This approach is then repeated for the remaining $d-1$-coordinates, so at the end we have 
$r_1,\dots,r_d>0$. Then 
\begin{equation}\label{eq:tstar}
r^* := r^*_{D_X} := \frac 13 \min\{r_1,\dots,r_d\}
\end{equation}
 satisfies \eqref{proper-2}
and the used algorithm is $\ca O(d\cdot m \log m)$ in time. 
Our next result shows that we can also ensure \eqref{proper-1} by jiggling the cubic partitions. 
Being rather technical, the proof is deferred to the Appendix \ref{app:technical-stuff}.

\begin{theorem}[Existence of Properly Aligned Cubic Partitioning Rule]
\label{ruckel-theorem}
For all $d\geq 1$,
$s\in (0,1]$, and $m\geq 1$ there exist an $m$-sample cubic partitioning rule $\pi_{m,s}$  with $|\Image(\pi_{m,s})|\leq (m+1)^d$
that assigns to each 
set of points 
$D_X:=\{x_1^*, ..., x_m^*\} \in \potm$ a cubic partition $\ca A$ that is properly aligned to $\{x_1^*, ..., x_m^*\}$  
with parameter $r:= r_{D_X} := \min\{r^*,\frac s{3m+3}\}$, where 
$r^*= r^*_{D_X} $ is defined in \eqref{eq:tstar}. 
\end{theorem}

The construction of an $m$-sample cubic partitioning rule $\pi_{m,s}$ basically relies on the representation \eqref{cubic-def} of cubic partitions $\cB$ 
of $\mbr^d$. In fact, the proof of Theorem \ref{ruckel-theorem} shows that there exists a finite set $x_1^\dagger , ..., x^\dagger_K \in \mbr^d$ 
of candidate offsets, with $K=(m+1)^d$. While at first glance this number seems to be prohibitively large 
for an efficient search, it turns out that the proof of Theorem \eqref{ruckel-theorem}
actually provides  a simple algorithm that is $\ca O(d\cdot m)$ in time for identifying coordinate-wise the $x_\ell^\dagger$ that 
leads to $\pi_{m,s}(\{x_1^*, ..., x_m^*\})$ .

\vspace*{1ex}

Being now well prepared, we introduce the class of \emph{inflated histograms}.

\begin{definition}
\label{def:inflatedHR}
 Let $s \in (0,1]$ and $m\geq 1$. Then a function $f:X\to Y$ is 
 called an $m$-inflated histogram of width $s$, if 
 there exist a subset $\{x_1^*, ..., x_m^*\} \in \potm$ and a cubic partition $\ca A$ of 
  width $s$ that is properly aligned to $\{x_1^*, ..., x_m^*\}$ with parameter $r\in [0,s)$
  such that 
  \begin{align*}
   f =  h  +  \sum_{i=1}^m b_i \eins_{x_i^* + t B_\infty  }\, , 
  \end{align*}
where $h\in \cH_\cA$, $t\in [0,r]$, and $b_i\in 2Y$ for all $i=1,\dots,m$. 
  We denote the set of all $m$-inflated histograms of width $s$ by $\cF_{s,   m}$.
  Moreover, for $n\geq 1$ we write 
  \begin{align*}
   \ca F^*_{s,n} := \ca F_{s,1}\cup \dots \cup \ca F_{s,n}\, .
  \end{align*}
\end{definition}

Note that the condition $t\leq r < s$ ensures that the representation $f=  h  +  \sum_{i=1}^m b_i \eins_{x_i^* + t B_\infty  }$
of any $f\in \cF_{s,  m}$ is unique.In addition, given an $f\in \ca F^*_{s,n}$,
the number $m$ of inflation points 
$\{x_1^*, ..., x_m^*\}$ is uniquely determined, too, and hence so is the 
representation of $f$.

\begin{figure}[t]
\begin{center}
\includegraphics[width=0.55\textwidth, height=0.4\textheight]{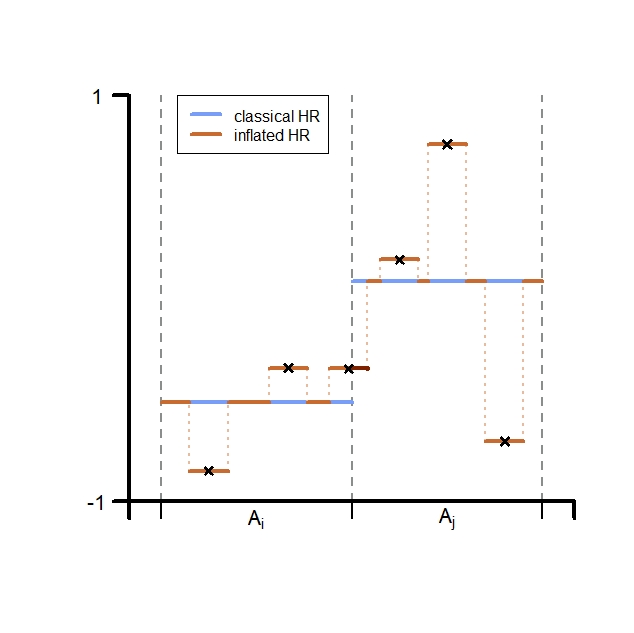}
%\hspace*{0.1\textwidth}
\hspace{-1.8cm}
\includegraphics[width=0.55\textwidth, height=0.4\textheight]{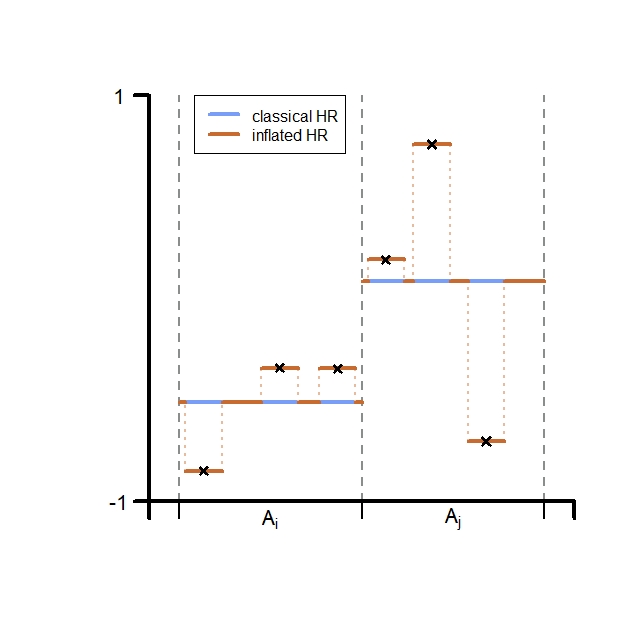} 
\vspace*{-4ex}
\end{center}
\caption{Left. Depiction of an inflated histogram for regression for a cubic partition $\cA=(A_j)_{j \in J}$ that is not properly aligned to the data 
(black crosses). The predictions  $c_i^*$ and $c_j^*$ on the associated cells $A_i$ and $A_j$ are calculated according to \eqref{eq:HRR}, i.e. by a local average. 
Mispredicted samples are corrected according to \eqref{def-bi} on a $tB_\infty$-neighborhood for some small $t > 0$. Note that one sample is too close to the 
cell boundary, i.e. \eqref{proper-1} is violated.  
Right.  An  inflated histogram that is properly aligned to the same data set. Note that \eqref{proper-1} ensures that boundary effects as for the left HR do not take place. 
For inflated histograms these effects seem to be a negligible technical nuisance. For their DNN counterparts considered
in Section \ref{sec:ReLU-Approx}, however, such effects may significantly complicate the constructions of interpolating predictors, see Figure \ref{figure:dnn-approx}.
}
\label{figure:histo-inflated}
\end{figure}

\vspace*{3ex}
So far we have formalized the notion of \emph{interpolation} and defined an appropriate inflated  hypotheses class 
for  modified histograms. In our next result we go a step further by providing a sufficient condition for the existence of interpolating predictors in $\cF_{s,  m}$.

\begin{proposition}\label{suff-interpol-erm}
Let $L$ be a loss that is interpolatable for  $D=((x_1,y_i),\dots,(x_n,y_n))$ 
and let $x_1^*,\dots, x_m^*$ be as in \eqref{interpol-function}.
Moreover, for $s\in (0,1]$ and $r>0$ we fix an $f^*\in  \cF_{s,  m}$  with representation as given in Definition \ref{def:inflatedHR}. 
For $i=1,\dots,m$ let $j_i$ be the index such that $x_i^* \in A_{j_i}$.
Then $f^*$ interpolates $D$,
if for all $i=1,\dots,m$ we have 
\begin{align}\label{def-bi}
   b_i = - c_{j_i} +  \arg\min_{c\in Y}  \sum_{k: x_k = x_i^*} L(x_k, y_k, c)  \, .
\end{align}
\end{proposition}

\begin{proofof}{Proposition \ref{suff-interpol-erm}} 
By our assumptions we   have 
\begin{displaymath}
 c_i^* :=   b_i + c_{j_i} 
\in  \arg\min_{c\in Y}  \sum_{k: x_k = x_i^*} L(x_k, y_k, c)  
= \arg\min_{c\in \R}  \sum_{k: x_k = x_i^*} L(x_k, y_k, c) \, , 
\end{displaymath}
where the last equality is a consequence of the fact that there is an $f:X\to Y$
satisfying \eqref{interpol-function}. Moreover, since 
\eqref{proper-1} and \eqref{proper-2} hold, we find 
$f^*(x_i^*) = h(x_i^*) + b_i =    c_{j_i} + b_i = c_i^*$, and therefore 
$f^*$ interpolates $D$ by \eqref{interpol-function}.
\end{proofof}

Note that for all $c_{j_i} \in Y$ the value $b_i$ given by \eqref{def-bi}
satisfies $b_i \in 2Y $ and we have $b_i=0$ if $c_{j_i}$ is contained in the $\arg\min$ in \eqref{def-bi}. 
Consequently, \emph{defining} $b_i$ by \eqref{def-bi} always gives an interpolating 
 $f^*\in  \cF_{s,  m}$.
Moreover, \eqref{def-bi} shows that an interpolating 
$f^*\in  \cF_{s,  m}$ can have an \emph{arbitrary} histogram part $h\in \ca H_{\ca A}$, that is, 
the behavior of $f^*$ 
outside the 
 small $tB_\infty$-neighborhoods around the samples of $D$ can be arbitrary.
In other words, as soon as 
we have found a properly aligned cubic partition $\ca A$ in the sense of $\cF_{s,m}$,
we can pick an arbitrary histogram  $h\in \ca H_{\ca A}$ and compute 
the $b_i$'s by \eqref{def-bi}. Intuitively, if the chosen $tB_\infty$-neighborhoods
are sufficiently small, then the prediction capabilities of the resulting interpolating predictor  are (mostly) determined 
by the chosen histogram part $h\in \ca H_{\ca A}$.
Based on this observation,
we can now construct different, interpolating $f^*_D\in  \cF_{s,  m}$ 
that have particularly good and bad learning behaviors.

\begin{example}[Good interpolating histogram rule]\label{good-guy}
Let $L$ be the least squares loss, $s\in (0,1]$ be a cell width,
$\rho \geq 0$ be an \emph{inflation parameter},
and $D=((x_1,y_i),\dots,(x_n,y_n))$ be a data set. By $D_X=\{x_1^*,...,x_m^*\}$ we denote the set of all 
covariates $x_j \in X$ with $(x_j, y_j)$ belonging to the data set. 
For $m=|D_X|$, Theorem \ref{ruckel-theorem} ensures the existence of a cubic partition $\cA_D=\pi_{m,s}(D_X)$ with width $s\in (0,1]$, 
being properly aligned to $D_X$ with the \emph{data-dependent} parameter $r$.
Based on this \emph{data-dependent} cubic partition $\cA_D$ we fix an empirical histogram for regression %$h_{D,\ca A}^+ \in \ca H_{\ca A}$ as given in \eqref{eq:HRR} 
\begin{equation}
\label{eq:histo-good}
    h_{D,\ca A_D}^+ :=  \sum_{j\in J} c_j^+ \eins_{A_j} \in \ca H_{s}
\end{equation}
with coefficients $(c_j^+)_{j \in J}$ precisely given in \eqref{eq:HRR}. 
Applying now Proposition \ref{suff-interpol-erm} 
gives us an $\gh  \in \cF_{s,  m} \subset \ca F^*_{s,n}$,
which interpolates $D$ and has the representation 
\[  \gh  :=  h_{D,\ca A_D}^+  + \sum_{i=1}^m b^+_i \eins_{x_i^* + t B_\infty  } \;, \]
where the $b^+_1,\dots,b_m^+$ are calculated according to the rule \eqref{def-bi},   
and $t := \min\{r, \rho\}$ is again \emph{data-dependent}. %  needs to satisfy $0\leq t\leq r$.
We call the map $D\mapsto \gh $ a \emph{good interpolating histogram rule}. 
\end{example}

\begin{example}[Bad interpolating histogram rule]\label{bad-guy}
Let $L$ be the least squares loss, $s\in (0,1]$ be a cell width,
$\rho \geq 0$ be an \emph{inflation parameter},
and $D=((x_1,y_i),\dots,(x_n,y_n))$ be a data set. Consider again a 
cubic partition $\cA_D=\pi_{m,s}(D_X)$ with width $s\in (0,1]$, that is properly aligned to $D_X$ with parameter $r$ and fix 
an empirical histogram $h_{D,\ca A_D}^+ \in \ca H_{s}$ as in \eqref{eq:histo-good}. 
Setting $t := \min\{r, \rho\}$, we define 
a predictor  $\bh \in  \cF_{s,  m}$  by 
\[ \bh :=    h_{D,\ca A_D}^- + \sum_{i=1}^m b^-_i \eins_{x_i^* + t B_\infty  }  \;,\]
with  $\ca H_{\ca A}$-part  $ h_{D,\ca A_D}^- := -  h_{D,\ca A_D}^+$. The $b^-_1,\dots,b_m^-$  are calculated according to \eqref{def-bi} and satisfy 
\[  b_i^- = b_i^+ + 2c_{j_i}^+ \;, \]
for any $i=1,...,m$ and where $j_i$ denotes the index such that $x_i^* \in A_{j_i}$.  
By writing 
\begin{align}\label{def:dxpt}
 D_X^{+t} := \bigcup_{i=1}^m \bigl( x_i^* + t B_\infty \bigr) 
\end{align}
we easily see that the definition of $\bh$ gives $\bh  \in \cF_{s,  m} \subset \ca F^*_{s,n}$ and
\begin{align}\label{eq-diff-good-bad-hr}
 \bh (x)
 = 
 \begin{cases}
  \gh (x) & \mbox{ if } x\in D_X^{+t}\\
  -\gh(x) & \mbox { if } x\not\in D_X^{+t}\, ,
 \end{cases}
\end{align}
while Proposition \ref{suff-interpol-erm} ensures that $\bh$ 
interpolates $D$. We call the map $D\mapsto \bh$ a 
\emph{bad interpolating histogram rule} and remark that $t$ is, like for  good interpolating histogram rules, data-dependent.
\end{example}

Our main results below show that the description \emph{good/ bad interpolating histogram rule} from the 
above Examples \ref{good-guy}/ \ref{bad-guy}, respectively, 
is  indeed justified, provided the inflation parameter is chosen appropriately.
Here we recall that good learning algorithms can be described by a small excess risk,
or equivalently, a small distance to the Bayes decision function $\fpb$,
see \eqref{eq:excess-risk}. To describe bad learning behavior, 
we denote the \emph{point spectrum} of $P_X$ by 
\begin{equation}
\label{eq:delta}
 \Delta:=\{ x \in X\; : \; P_X(\{x\}) > 0  \}  \;, 
\end{equation} 
see \cite{HoJo17}.  One easily verifies that $\Delta$ is  at most countable, since $P_X$ is finite.   
Moreover, for an arbitrary but fixed version $\fpb$ of the Bayes decision function, we write
\begin{align*}
 \fpd :=  \eins_\D \fpb - \eins_{X\setminus \D }\fpb 
 \qquad \mbox{ and }\qquad 
 \cR^\dagger_{L,P} &:= \RP L \fpd \, , 
\end{align*}
where we note that $\cR^\dagger_{L,P}$ does, of 
course, not depend on the choice of $\fpb$. 
Moreover, note that for $x\in \Delta$ the value $\fpb(x)$ is also independent of 
the choice of $\fpb$
and it holds $f^\dagger_{L,P} (x) = \fpb(x)$. In contrast, for $x\in X\setminus \Delta$
with $\fpb(x) \neq 0$ we have 
$f^\dagger_{L,P}(x) \neq  \fpb(x)$. In fact, a quick calculation using
\eqref{eq:excess-risk} shows 
\begin{align}\label{eq:fpd-vs-fpb}
 \cR^\dagger_{L,P}  - \RPB L 
 =
 \snorm{\fpd - \fpb}_{\Lx 2 {P_X}}^2
 = 
4  \snorm{\eins_{X\setminus \D}\fpb}_{\Lx 2 {P_X}}^2\, , 
\end{align}
and consequently we have $\cR^\dagger_{L,P}  - \RPB L>0$ whenever $P_X(\D) < 1$
and $\fpb$ does not almost surely vanish on $X\setminus \D$. It seems fair to say that 
the overwhelming majority of ``interesting'' $P$ fall into this category. 
Finally, note that in general we do not have 
an equality of the form  \eqref{eq:excess-risk}, when we replace 
$\RPB L$ and $\fpb$ by 
$\cR^\dagger_{L,P}$ and $\fpd$. However, 
for $y,t,t'\in Y=[-1,1]$ we have $|L(y,t) - L(y,t')| \leq 4 |t-t'|$, and consequently
we find 
\begin{align}\label{eq:excess-dagger-risk}
 \bigl|\RP Lf - \cR^\dagger_{L,P}\bigr| \leq 4\, \snorm{f - \fpd}_{\Lx 2 {P_X}}
\end{align}
for all $f:X\to Y$. For this reason, we will investigate the bad interpolating 
histogram rule only with respect to its $L_2$-distance to $\fpd$.

Before the state our  main result of this section we need to introduce one more assumption
that will be required for parts of our results.

\begin{assumption}
\label{ass:proba-2}
There exists a non-decreasing continuous map $\varphi: \mbr_+ \to \mbr_+$ with $\varphi(0)=0$  
such that for any $t \geq 0$
and $x \in X$ one has  
$P_X(x + tB_\infty ) \leq \varphi( t) $.  
\end{assumption}

Note that this assumption implies $P_X(\{x\})=0$ for any $x \in X$. Moreover, it is satisfied 
for the uniform distribution $P_X$, if we consider $\phi(t) := 2^d t^d$, and a simple argument shows
that modulo the constant appearing in $\phi$ the same is true if $P_X$ only has a bounded 
Lebesgue density. The latter is, however, not necessary. Indeed, for $X=[-1,1]$ and $0<\beta< 1$ it is easy to 
construct unbounded Lebesgue densities that satisfy Assumption \ref{ass:proba-2}
for $\phi$ of the form $\phi(t) = ct^\beta$, and higher dimensional analogons are also easy to construct.
Moreover, in higher dimensions Assumption 
\ref{ass:proba-2} also applies to various distributions living on sufficiently smooth low-dimensional 
manifolds.

With these preparations we can now establish the following theorem that shows that for $\rho=0$ the good interpolating 
histogram rule is universally consistent while the bad interpolating histogram rule fails to be consistent in a stark sense. 
It further shows  consistency, respectively non-consistency for $\rho=\rho_n>0$ with $\rho_n\to 0$.

\begin{theorem}
\label{results:main-erm}
Let $L$ be the least-squares loss and let $D \in (X \times Y)^n$ be an i.i.d. sample of size $n \geq 1$. 
Let $D \mapsto \gh $ denote the good interpolating histogram rule from Example \ref{good-guy}. 
Similarly, let $D \mapsto \bh $ denote the bad interpolating histogram rule from Example \ref{bad-guy}. 
Assume that $(s_n)_{n \in \mbn}$ is a sequence with 
$s_n \rightarrow 0$ as well as $\frac{\ln (n s_n^d)}{n s_n^d}\to 0$ as $n \to \infty$.  
\begin{enumerate} 
\item (Non)-consistency for $\rho_n = 0$. We have in probability for $|D|\to \infty$ 
\begin{align}
\label{consistence-good}
\snorm{\ghon - \fpb}_{\Lx 2 {P_X}} & \to 0 \,, \\
\label{consistence-bad}
\snorm{\bhon - \fpd}_{\Lx 2 {P_X}} & \to 0 \, .
\end{align}

\item (Non)-consistency for $\rho_n >0$.
Let $(\rho_n)_{n \in \mbn}$ be a non-negative sequence with 
$\rho_n \to 0$ as $n \to \infty$. 
Then for all distributions $P$ that satisfy 
Assumption \ref{ass:proba-2} for a function $\varphi$ with $n\varphi(\rho_n) \to 0$ for $n\to \infty$, we have 
\begin{align}
\label{consistence-good-2}
||\ghn  - \fpb||_{L_2(P_X)} \to 0   \,, \\
\label{consistence-bad-2}
||\bhn  - \fpd||_{L_2(P_X)} \to 0 \,,
\end{align}
in probability for $|D|\to \infty$.
\end{enumerate}
\end{theorem}

The proof of Theorem \ref{results:main-erm} is provided in Appendix \ref{app-proof-consistency-histos}. 
Our second main result, whose proof is provided in Appendix \ref{app-proof-rates-histos}, refines the above theorem and establishes learning rates  for the good and bad interpolating histogram rules, provided 
the width $s_n$ and 
the inflation parameter $\rho_n$ decrease sufficiently fast as $n \to \infty$.

\begin{theorem}[Learning Rates]
\label{results:main-erm2}
Let $L$ be the least-squares loss and let $D \in (X \times Y)^n$ be an i.i.d. sample of size $n \geq 1$. 
Let $D \mapsto \gh $ denote the good interpolating histogram rule from Example \ref{good-guy}. 
Similarly, let $D \mapsto \bh $ denote the bad interpolating histogram rule from Example \ref{bad-guy}. 
Suppose that $\fpb$ is $\a$-H\"older continuous with $\alpha \in (0,1]$
and that $P$ satisfies Assumption \ref{ass:proba-2} for some function $\varphi$.
Assume further that $(s_n)_{n \in \mbn}$ is a sequence with 
\[   s_n = n^{-\gamma}\;, \quad \gamma = \frac{1}{2\alpha + d}   \]
and that $(\rho_n)_{n\geq 1}$ is  a non-negative 
sequence with $n \varphi(\rho_n) \leq \ln(n) n^{-2/3}$ for all $n\geq 1$.
Then there exists a constant 
$c_{d,\alpha }>0$   only depending on $d$, $\alpha$, and $|f^*_{L,P}|_\alpha$, such that
for all $n\geq 1$ the good interpolating histogram rule satisfies 
\begin{align}\label{rates-good}
||\ghn  - \fpb||_{L_2(P_X)} &\leq c_{\alpha,d} \sqrt{\ln(n)}\paren{\frac{1}{n}}^{\alpha\gamma }  \;, 
\end{align}
with probability $P^n$ not less than $1- 2^dn^{1+d} e^{-n^{d \gamma}}$. 
Furthermore, for all 
$n\geq 1$, the bad interpolating histogram rule satisfies 
\begin{align}  \label{rates-bad}
||\bhn  - \fpd||_{L_2(P_X)}&\leq c_{\alpha,d} \sqrt{\ln(n)}\paren{\frac{1}{n}}^{\alpha\gamma }\;.
\end{align}
with probability $P^n$ not less than $1- 2^dn^{1+d} e^{-n^{d \gamma}}$.
\end{theorem}

To set the results above in context, let us first recall that 
even for a fixed hypotheses class, ERM is, in general, \emph{not} a single algorithm, but  a collection of algorithms. In fact, this ambiguity appears, as soon as 
the ERM-optimization problem has not a unique solution for certain data sets, and as 
Lemma \ref{result:hr-is-erm} shows, this non-uniqueness may even occur 
for strictly convex loss functions such as the least squares loss.
Now, the standard techniques of statistical learning theory are capable of 
showing that for sufficiently small hypotheses classes, \emph{all versions} of ERM enjoy good statistical guarantees. In other words, the non-uniqueness of ERM does not affect 
its learning capabilities as long as the hypotheses class is sufficiently small. 
In addition, it is folklore that in some large hypotheses classes, there may be heavily 
overfitting 
ERM solutions, leading to the usual conclusion that   such hypotheses classes 
should be avoided.

In contrast to this common wisdom, however,
Theorem \ref{results:main-erm} demonstrates that for large hypotheses classes, the situation 
may be substantially more complicated: 
First, it shows that there exist ERMs, whose predictors converge to a function $\fpd$, see 
\eqref{consistence-bad}, that in 
almost all interesting cases is far off the target regression function, see 
\eqref{eq:fpd-vs-fpb}, confirming that 
the overfitting issue is indeed present for the chosen hypotheses classes.
Moreover, this strong overfitting may actually take place with fast convergence, see 
\eqref{rates-bad}.
Despite this negative result, however, 
we can also find ERMs that enjoy a good learning behavior in terms of consistency \eqref{consistence-good} and almost optimal learning rates \eqref{rates-good}.
In other words, both the expected overfitting and standard learning guarantees may be 
realized by suitable versions of ERM over these hypotheses classes. 
In fact, these two different behaviors are just extreme examples, and a variety of 
intermediate behaviors are possible, too: Indeed, as the training error can be solely 
controlled by the corrections on the inflating  parts, the behaviour
of the histogram part $h$ can be arbitrarily chosen. 
For our theorems above, we have chosen a particular good and bad $h$-part, repectively, 
but of course, a variety of other choices leading to intermediate behavior are also possible.
As a consequence, the ERM property of an algorithm working with a large hypotheses class
is, in general, no longer a sufficient notion for describing its learning behavior. Instead, 
additional assumptions are required to determine its learning behavior. In this respect 
we also note that for our inflated hypotheses classes, other learning algorithms that 
do not (approximately) minimize the empirical risk may also enjoy good learning properties.
Indeed, by setting the inflating parts to zero, we recover standard histograms, which 
in geneneral do not have close-to-zero training error, but for which the 
guarantees of our good interpolating predictors also hold true.

Of course, the chosen hypotheses classes may, to some extent, appear artificial. 
Nonetheless, in the following section they will be key for showing
 that for sufficiently large DNN architectures exactly the same phenomena occur 
 for some of their global minima.

\section{Approximation of histograms with ReLU networks}
\label{sec:ReLU-Approx}

The goal of this section is to build  neural networks of suitable depth and width that  mimic the learning properties of 
inflated histogram rules. To be more precise, we aim to construct a particular class of \emph{inflated networks} that contains 
\emph{good} and \emph{bad} interpolating predictors, similar to the good and bad interpolating  histogram rules from 
Example \ref{good-guy} and Example \ref{bad-guy}, respectively.

We begin with describing in more detail the specific networks that we will consider. 
Given an activation function $\sigma: \mbr \to \mbr$ and $b \in \mbr^p$ 
we define the {\it shifted activation function} $\sigma_b: \mbr^p \to \mbr^p$ as 
\begin{equation}
\label{eq:shifted-activation}
  \sigma_b (y) := ( \sigma(y_1+b_1), ..., \sigma(y_p+b_p) )^T \;. 
\end{equation} 
A {\it hidden layer} with activation $\sigma$, of {\it width} $p \in \mbn$ and with input dimension $q \in \mbn$ is a function $H_\sigma:\R^q \to \R^p$
of the form
\begin{align}\label{eq:layer-repres}
 H_\sigma (x) := (\sigma_b \circ A )( x) \;,  \quad x\in \R^q, 
\end{align}
where $A$ is a $p \times q$ {\it weight matrix} 
%\fix{\noteis{Can we live with the notational conflict of $P$? It also denotes a distribution. Possible alternative $q$?}}
and $b \in \mbr^{p}$ is a {\it shift vector} or {\it bias}. 
Clearly, each pair $(A,b)$ describes a layer, but in general, a layer, if viewed as a \emph{function}, can be described by more than one such pair. 
The class of networks we consider is given in the following definition. 

\begin{definition}
Given an activation function $\sigma: \mbr \to \mbr$ and an integer $L\geq 1$, 
a neural network with architecture $p \in \mbn^{L+1}$ is a function 
$f: \mbr^{p_0} \to \mbr^{p_{L}}$,  having a representation of the form 
\begin{align}
\label{eq:basic-rep}
  f(x) &= H_{\mbox{id} , L}\circ H_{\sigma ,L-1}\circ \dots\circ H_{\sigma, 1}(x)  \;, \quad x \in \mbr^{p_0} \;,%\no \\
%  &= A^{(L)}\sigma_{b^{(L-1)}}A^{(L-1)}\sigma_{b^{(L-2)}} \cdot ...\cdot A^{(2)}\sigma_{b^{(1)}}A^{(1)} (x)  +  b^{(L)} \;, 
\end{align} 
where $H_{\sigma , l}: \mbr^{p_{l-1}} \to \mbr^{p_l}$ is a hidden layer of width $p_l \in \mbn$ and input dimension 
$p_{l-1} \in \mbn$, $l=1,...,L-1$. Here, the last layer  $ H_{\mbox{id} , L}: \mbr^{p_{L-1}} \to \mbr^{p_L} $ is associated to the 
identity  $\mbox{id}: \mbr \to  \mbr$. 
\end{definition} 

A network architecture is therefore described by an activation  function $\sigma$ and a {\it width vector} $p = (p_0, ...,p_L) \in \mbn^{L+1}$.  
The positive integer $L$ is the number of {\it layers}, $L-1$ is the number of  {\it hidden layers} or the 
{\it depth}. Here, $p_0$ is the input dimension and $p_{L}$ is the output dimension. 
In the sequel, we confine ourselves to the ReLU-activation function $|\cdot |_+ : \mbr \to [0,\infty)$ defined by 
\[ |t|_+:=\max\{0,t\} \;, \quad t \in \mbr \;.  \] 
Moreover, we consider networks with fixed input dimension $p_0=d$ and output dimension $p_L=1$, that is, 
%the last layer is a single neuron without activation function, that is, 
\[ H_{\mbox{id} , L}(x) = \langle a,x \rangle + b \;, \quad x\in \mbr^{p_{L-1}} \;.\]
Thus, we may parameterize the (inner) architecture by the width vector $(p_1,...,p_{L-1}) \in \mbn^{L-1}$ of the hidden layers only. 
In the following, we denote the set of all such neural networks by $\cA_{p_1,...,p_{L-1}}$.

\subsection{$\e$-Approximate Inflated Histograms}\label{subsec:eps-inflate-histo-dnn}

Motivated by the representation \eqref{hr-hypo-space} 
%\eqref{proper} 
for histograms, the first step of our construction 
approximates the indicator function of an multi-dimensional interval
by a small part of a possibly large DNN. This will be our main building block. 
We emphasize that the ReLU activation function is particularly suited for this approximation and
it thus  plays a key role in our entire construction.

For the formulation of the corresponding result we fix some notation. 
For $z_1, z_2 \in \Rd$ we write $z_1\leq z_2$ if each coordinate satisfies $z_{1,i}\leq z_{2,i}$, $i=1,\dots,d$. 
We define $z_1 < z_2$ analogously. 
In addition, if $z_1\leq z_2$, then the
 multi-dimensional interval is $[z_1, z_2] := \{ z\in \Rd: z_1\leq z\leq z_2    \}$,
and we similarly define $(z_1, z_2)$ if  $z_1 < z_2$. 
Finally, for $s\in \R$, we let $z_1 + s := (z_{1,1}+s,\dots,z_{1,d}+s)$.

\begin{definition}[$\e$-Approximation]
\label{def:eps-approx}
Let $A\subset X$,  $z_1, z_2\in \Rd$ with $z_{1} < z_{2}$ and $\e>0$ with  $\e < \min\{z_{2,i}-z_{1,i}: i=1,\dots,d \}$.  
Then a network   $\eins_A^{(\e)} \in \cA_{2d,1}$ is called an \emph{$\e$-Approximation of the indicator function $\eins_A: X \to [0,1]$} if 
\[ \{ \eins_A^{(\e)} = \eins_{A}\} = [z_{1}+\e, z_{2}-\e]   \cup \bigl( X\setminus A \bigr) \;, \]
and if 
\[ \{\eins_A^{(\e)} >1 \} = \emptyset \;, \quad \{\eins_A^{(\e)} <0 \} = \emptyset  \;. \] 
\end{definition}
%\fix{\notice{I introduced '$\e$-Approximation' since we are using this notion very often}
%\noteis{Gute idee. Brauchen wir evtl.~auch dass $\{\eins_A^{(\e)} <0 \} = \emptyset$?}}
%\fix{\noteis{Achtung: Die Existenz-Proposition \ref{result:DNN-aprox-HR} hat sich ganz leicht ge\"andert, siehe fixboxes dort!}}

The next lemma ensures the existence of such approximations. 
The full construction is elementary calculus and is provided in Appendix \ref{app-neuron-lego}, in particular in Lemma \ref{result:one-bump-d-v2}.  
Lemma \ref{lemma:aprox-error-bump} provides then the desired properties.

\begin{lemma}[Existence of $\e$-Approximations]
\label{result:one-bump-d}
Let $z_1, z_2\in \Rd$ and $\e>0$ as in Definition \ref{def:eps-approx}. 
Then for all $A\subset X$ with $[z_{1}+\e, z_{2}-\e] \subset A \subset [z_{1}, z_{2}]$ there exists 
an $\e$-Approximation $ \eins_A^{(\e)}$ of $\eins_A$. 
\end{lemma}

\begin{figure}[t]
\begin{center}
\includegraphics[width=0.49\textwidth, height=0.26\textheight]{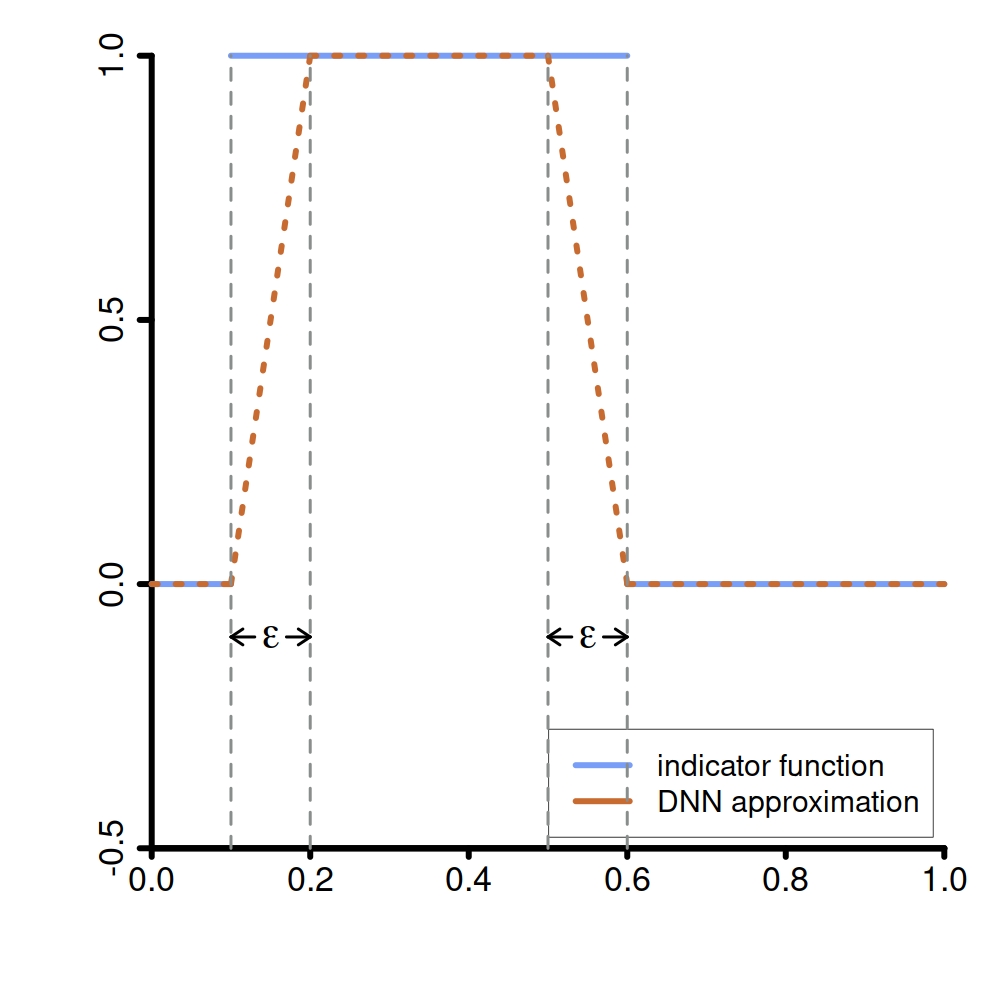}
%\hspace*{0.1\textwidth}
\includegraphics[width=0.49\textwidth, height=0.26\textheight]{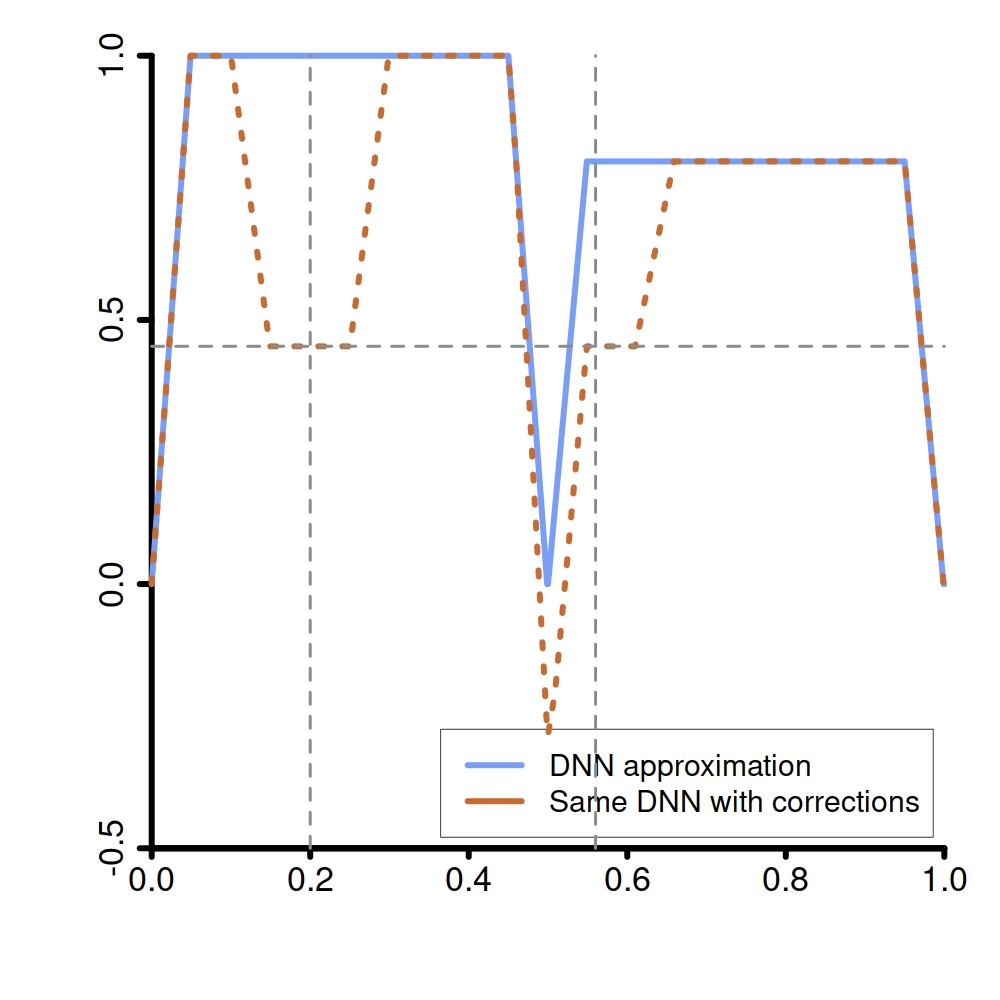} 
\vspace*{-4ex}
\end{center}
\caption{Left. Approximation $\eins_A^{(\e)}$ (orange) of the indicator function $\eins_A$ for $A =[0.05, 0.45]$ (blue)
according to Lemma \ref{result:one-bump-d} for $\e = 0.1$ on $X=[0,1]$. The construction of $\eins_A^{(\e)}$
ensures that $\eins_A^{(\e)}$ coincides with $\eins_A$ modulo a small set that is controlled by $\e>0$.
Right. A DNN (orange) for regression that approximates the histogram $\eins_{[0,0.5)} + 0.8 \cdot \eins_{[0.5,1)}$
and a DNN (green) that additionally  tries to interpolate two samples $x_1 = 0.15$ and $x_2= 0.975$ 
(located at the two vertical dotted lines)
with $y_i = -0.5$. The  label
$y_1$ is correctly interpolated since the alignment condition \eqref{proper-1} is satisfied for $x_1$ with $t=0.15$
and $\e=\delta = t/3 = 0.05$ as in Example \ref{good-DNN}.
In contrast, $y_2$ is not correctly interpolated since condition \eqref{proper-1} is violated for this $t$
and hence $\e$ and $\delta$ are too large.
}\label{figure:dnn-approx}
\end{figure}

Figure \ref{figure:dnn-approx} illustrates  $\eins_A^{(\e)}$ for $d=1$. 
Moreover, the proof of Lemma \ref{result:one-bump-d-v2} shows that out of the $2d^2$ weight parameters of the first layer, only $2d$ are non-zero.
In addition, the $2d$ weight parameters of the neuron in the second layer are all identical.
In order to approximate inflated histograms we need to know how 
to combine several functions of the form provided by Lemma \ref{result:one-bump-d} into a single neural network. 
An appealing feature of our DNNs  
is that the concatenation of layer structures is very easy.

\begin{lemma}
If  $c \in \mbr$, $(p_1, p_2) \in \mbn^2$, and 
$g \in \cA_{p}$, $g' \in \cA_{p'}$, then $cg \in \cA_{p}$ and $g + g' \in \cA_{p+p'}$. 
\end{lemma}
For an extended version of this result, see Lemma \ref{result:dnn-algebra-2}.  
In particular, our constructed DNNs have a particularly sparse structure and 
the number of required neurons behaves in a very controlled  and natural fashion.

With these insights, we are now able to  find a representation similar to \eqref{hr-hypo-space}.  
To this end, we choose a cubic partition $\cA=(A_j)_{j \in J}$ of $X$ with width $s>0$ and define for $\e \in (0, \frac s3]$
\begin{align*}%\label{hr-DNN}
   \cH^{(\e)}_{\ca A}  := \biggl\{   \sum_{j \in J}  c_j  \; \eins_{A_j}^{(\e)} \; : \; c_j \in Y  \biggr\}\, , 
\end{align*}
where $\eins_{A_j}^{(\e)} := (\eins_{B_j}^{(\e)})_{|A_j}$ is the restriction of $\eins_{B_j}^{(\e)}$ to $A_j$  
and $\eins_{B_j}^{(\e)}$ is an $\e$-approximation of $\eins_{B_j}$
of  Lemma \ref{result:one-bump-d}. Here, $B_j$ is the cell with $A_j = B_j\cap X$, see
the text around \eqref{cubic-def}. 
We call any function in $\cH^{(\e)}_{\ca A}$ an \emph{$\e$-approximate histogram}.

Our considerations above show that 
we have   $  \cH^{(\e)}_{\ca A}  \subset \cA_{p_1,p_2}$ with $p_1 = 2d|J|$ and $m_2 = |J|$. %=(2d|J|, |J|)\in \mbn^2$.  
Thus, any  $\e$-approximate histogram can be represented by a  neural network with $2$ hidden layers. 
Inflated versions are now straightforward.

\begin{definition}
\label{eq:class-dnn}
Let $s \in (0,1]$, $m\geq 1$, and $\e \in (0, s/3]$.  Then a function $f: X \to Y$ is called an $\e$-approximated $m$-inflated histogram 
of width $s$ if  
there exist a subset $\{x_1^*, ..., x_m^*\} \in \potm$ and a cubic partition $\ca A$ of 
width $s$ that is properly aligned to $\{x_1^*, ..., x_m^*\}$ with parameter $r\in [0,s)$
such that 
%such that 
\[ f^{(\e)} = h^{(\e)}  +  \sum_{i=1}^m b_i  \eins^{(\delta )}_{x_i^* + t B_\infty} \;, \]
where $h^{(\e)} \in \cH_\cA^{(\e)}$, $t \in (0,r]$, $\delta \in (0, t/3] $, 
$b_i \in 2Y$ and where $I^{(\delta )}_{x_i^* + t B_\infty  }$ 
is a $\delta $-approximation of $\eins_{x_i^* + t B_\infty}$ for all $i=1,...,m$. 
We denote the set of all $\e$-approximated $m$-inflated histograms of width $s$ by $\cF^{(\e)}_{s, m}$. 
\end{definition}

A short calculation shows that $\cF^{(\e)}_{s, m} \subset \cA_{p_1,p_2}$ 
with $p_1 = 2d(m+|J|)$,  $p_2 =  m+|J|$  and $|J| \leq (2/s)^d$. 
With these preparations, we can now introduce good and bad interpolating DNNs.

\begin{example}[Good and bad interpolating DNN]
\label{good-DNN} 
Let $L$ be the least squares loss, $s \in (0,1]$ be a cell width and let $\rho > 0$ be an inflation 
parameter.  For a data set $D=((x_1,y_i),\dots,(x_n,y_n))$ we consider again a cubic partition 
$\cA_D=\pi_{m,s}(D_X)$, with $m=|D_X|$, being properly aligned to $D_X$ with parameter $r$. Set  
$t:=\min\{r, \rho\}$. According to Example \ref{good-guy}, a \emph{good interpolating HR} is given by  
\begin{displaymath}
  \gh  :=  \sum_{j\in J} c_j^+ \eins_{A_j}  + \sum_{i=1}^m b^+_i \eins_{x_i^* + t B_\infty  } \;, 
\end{displaymath}
where the $(c_j^+)_{j \in J}$ are given in \eqref{eq:HRR} and $b^+_1,\dots,b_m^+$ are from \eqref{def-bi}.  
For $\e := \delta := t/3$ 
we then define the \emph{good interpolating DNN} by
\begin{displaymath}
  \gdn =  \sum_{j\in J} c_j^+ \eins^{(\e)}_{A_j}  + \sum_{i=1}^m b^+_i \eins^{(\delta)}_{x_i^* + t B_\infty  } \, .
\end{displaymath}
Clearly, we have  $\gdn \in \cF^{(\e)}_{s, m}$. 
We call the map $D\mapsto \gdn$ a \emph{good interpolating DNN} and it is easy to see that this network indeed interpolates $D$. 
Finally, the \emph{bad interpolating DNN} $\bdn$ is defined analogously
using the bad interpolating HR from Example \ref{bad-guy}, instead.
\end{example}

Similarly to our inflated histograms from the previous section, the next theorem shows that the good interpolating DNN is 
consistent while the bad interpolating DNN fails to be. The proof of this result is given in Appendix \ref{sec:proof-thm2}.

\begin{theorem}[(Non)-consistency]
\label{results:main-dnn}
Let $L$ be the least-squares loss and let $D \in (X \times Y)^n$ be an i.i.d. sample of size $n \geq 1$. 
Let $D \mapsto \gdn$ denote the good interpolating DNN from Example \ref{good-DNN}. 
Similarly, let $D \mapsto \bdn $ denote the bad interpolating DNN from Example \ref{good-DNN}. 
Assume that $(s_n)_{n \in \mbn}$ is a sequence with 
$s_n \rightarrow 0$, $\frac{\ln (n s_n^d)}{n s_n^d}\to 0$ as $n \to \infty$ as well as $s_n > 2n^{-1/d}$.  
Additionally, let $(\rho_n)_{n \in \mbn}$ be a non-negative sequence with $\rho_n \leq 2n^{-1/d}$. 
%as $n \to \infty$. 
Then $\gbdnn \in \cA_{4dn, 2n}$. 
Moreover, for all distributions $P$ that satisfy 
Assumption \ref{ass:proba-2} for a function $\varphi$ with $\rho_n^{-d} \varphi( \rho_n ) \to 0$  for $n\to \infty$, we have 
\begin{align}
\label{consistence-good-dnn}
||\gdnn  - \fpb||_{L_2(P_X)} \to 0   \,, \\
\label{consistence-bad-dnn}
%\cR_{L,P}(\bho) &\to \cR_{L,P}(-\fpb)  \,.
||\bdnn  - \fpd||_{L_2(P_X)} \to 0 \,,
\end{align}
in probability for $|D|\to \infty$.
\end{theorem}

The above result can further be refined to establishing rates of convergence if the width $s_n$ and the inflation parameter $\rho_n$ converge 
to zero sufficiently fast as $n \to \infty$. The proof is provided in Appendix \ref{sec:proof-thm3}.

\begin{theorem}[Learning Rates]
\label{results:main-dnn-2}
Let $L$ be the least-squares loss and let $D \in (X \times Y)^n$ be an i.i.d. sample of size $n \geq 1$. 
Let $D \mapsto \gdn$ denote the good interpolating DNN from Example \ref{good-DNN}. 
Similarly, let $D \mapsto \bdn $ denote the bad interpolating DNN from Example \ref{good-DNN}. 
Suppose that $\fpb$ is $\a$-H\"older continuous with $\alpha \in (0,1]$
and that $P$ satisfies Assumption \ref{ass:proba-2} for some function $\varphi$.
Assume further that $(s_n)_{n \in \mbn}$ is a sequence with 
\[   s_n = n^{-\gamma}\;, \quad \gamma = \frac{1}{2\alpha + d}   \]
and that $(\rho_n)_{n\geq 1}$ is  a non-negative 
sequence with $\rho_n \leq 2n^{-1/d}$ and $\rho_n^{-d} \varphi(\rho_n) \leq \ln(n) n^{-2/3}$ for all $n\geq 1$.
Then there exists a constant 
$c_{d,\alpha }>0$   only depending on $d$, $\alpha$, and $|f^*_{L,P}|_\alpha$, such that
for all $n\geq 2$ the good interpolating histogram rule satisfies 
\begin{align}
\label{rates-good-dnn}
||\gdnn  - \fpb||_{L_2(P_X)} &\leq c_{\alpha,d} \sqrt{\ln(n)}\paren{\frac{1}{n}}^{\alpha\gamma }  \;, 
\end{align}
with probability $P^n$ not less than $1- 2^dn^{1+d} e^{-n^{d \gamma}}$. 
Furthermore, for all 
$n\geq 2$, the bad interpolating histogram rule satisfies 
\begin{align}
\label{rates-bad-dnn}
||\bdnn  - \fpd||_{L_2(P_X)}&\leq c_{\alpha,d} \sqrt{\ln(n)}\paren{\frac{1}{n}}^{\alpha\gamma }\;.
\end{align}
with probability $P^n$ not less than $1- 2^dn^{1+d} e^{-n^{d \gamma}}$. Finally, there exists a natural number $n_{d, \alpha} > 0$ such that for 
any $n \geq n_{d, \alpha}$ we have $\gbdnn \in \cA_{4dn, 2n}$. 
\end{theorem}

Note that 
the  rates of convergence in  
\eqref{rates-good-dnn} and \eqref{rates-bad-dnn} remain true if we consider 
a sequence $s_n$ with $c^{-1} n^{-\g} \leq s_n \leq cn^{-\g}$ 
for some constant $c$ independent of $n$. In fact, the only reason why we have formulated Theorem \ref{results:main-dnn-2} with $s_n = n^{-\g}$ is to avoid another constant appearing in the statements.
Moreover, if we choose $s_n := 2 a \lfloor n^{-\g}\rfloor^{-1}$
with $a:= 3^{1/d}/(3^{1/d}-2)$, then we have 
$|J| \leq  (2 s_n^{-1} + 2)^d \leq (a^{-1}n^{1/d} + 2)^d \leq n$ 
for all $n\geq 3$. Consequently, for $m:= n$, we can choose 
$n_{d, \alpha} := 3$, and hence we have $\gbdnn \in \cA_{4dn, 2n}$ for all $n\geq 3$ while
\eqref{rates-good-dnn} and \eqref{rates-bad-dnn} hold true modulo a change in the constant
$c_{\a,d}$.

\vspace{0.2cm}

\noindent
{\bf Discussion of results.} To fully appreciate 
Theorems \ref{results:main-dnn} and \ref{results:main-dnn-2} as well as their underlying construction let us discuss its various consequences:

First, the good interpolating DNN predictors $\gdnn$ show that  
its is actually possible to train sufficiently  large, over-parameterized DNNs such that they become consistent and enjoy
optimal learning rates up to a logrithmic factor without adapting the network size to the 
particular smoothness of the target function.
In fact, it suffices to consider DNNs with two hidden layers and $4dn$, respectively $2n$ neurons in the first, respectively second, hidden layer.
In other words,  Theorems \ref{results:main-dnn} and \ref{results:main-dnn-2}  already apply to moderately 
over-parameterized DNNs, and by the particular properties of the ReLU-activation function, also 
for all larger network architectures.
In addition, when  using architectures of minimal size, 
 training, that is constructing $\gdnn$, can be done in $\ca O(d^2\cdot n^2)$-time
if the DNNs are implemented as fully connected networks. Moreover, the constructed 
DNNs have a particularly sparse structure and exploiting this can actually reduce 
the training time to $\ca O(d \cdot n\cdot \log n)$.
While we believe that this is one of the very first statistically sound 
end-to-end\footnote{By ``end-to-end'' we mean the explicit 
construction of an efficient, feasible, and implementable  training algorithm and the 
rigorous statistical analysis of this very particular algorithm under minimal assumptions.}
proofs of consistency and optimal rates  for DNNs, we also need to admit that our
training algorithm is mostly interesting from a theoretical point of view, but useless
 for  practical purposes. 
 
 Second, 
Theorems \ref{results:main-dnn} and \ref{results:main-dnn-2}  also have its consequences for DNNs trained by variants of 
stochastic gradient descent (SGD) if the resulting predictor is interpolating. Indeed, these
theorems  show that  ending in a global minimum 
may result in either a very good learning behavior or an extremely 
overfitting, bad behavior. In fact, all the observations made for histograms at the end 
of Section \ref{sec:classical-histo}  
apply to DNNs, too. In particular, since for $n\geq n_{d,\a}$ the
$\cA_{4dn, 2n}$-networks can $\e$-approximate all functions in $\ca F_{s,n}^*$  
for all $\e\geq 0$ and all $s\in [n^{-1/d}, 1]$, we can, for example, find, for each 
polynomial learning rate slower than $n^{-\a\g}$, an interpolating learning method
$D\mapsto f_D$ with 
$f_D\in \cA_{4dn, 2n}$ that learns with  this rate. 
Similarly, we can find interpolating $f_D\in \cA_{4dn, 2n}$ 
with various degrees of bad learning behavior. In summary, the optimization landscape 
induced by $\cA_{4dn, 2n}$ contains a wide variety of global minima whose 
learning properties range somewhat continuously from essentially optimal to extremely poor.
Consequently, an optimization guarantee for (S)GD, that is, a guarantee that (S)GD 
finds a global minimum in the optimization landscape, is useless for learning guarantees
unless more information about the particular nature of the minimum found  is provided.
Moreover, it becomes clear that considering (S)GD without the initialization 
of the weights and biases is a meaningless endeavor: For example, constructing 
$\gbdnn$ can be viewed as a very particular form of initilization for which 
(S)GD won't change the parameters anymore. More generally, when initializing 
the parameters randomly in the attraction basin of $\gbdnn$ then GD will converge to 
$\gbdnn$ and therefore the behavior of GD is completely determined by the 
initialization.
In this respect note that so 
 far there is no statistically sound way to distinguish between good and bad 
interpolating DNNs on the basis of the training set alone, and hence 
the only way to identify good interpolating DNNs obtained by SGD is to use 
a validation set. Now, for the good interpolating DNNs of Theorem \ref{results:main-dnn} 
it is actually possible to construct a finite set of candidates
such that the one with the best validation error achieves the optimal 
learning rates without knowing $\a$. For DNNs trained by SGD, however, we do not have 
this luxury anymore. Indeed,
while we can still identify the best predicting DNN from a finite set of SGD-learned 
interpolating DNNs
 we no longer have  any theoretical understanding  of
 whether there is any useful candidate among them, or whether they all behave like a $\bdnn$.

Third, for both  consistency and learning with essentially optimal rates 
it is by no means necessary to find a global minimum, or at least a   local minimum,
in the optimization landscape. For example, the positive learning rates \eqref{rates-good}
also hold for ordinary cubic histograms with widths $s_n:= n^{-\g}$, and the latter can, of course, 
also be approximated by $\cA_{4dn, 2n}$. Repeating the proof of 
Theorem \ref{results:main-dnn-2} it is easy to verify that these approximations 
also enjoy the good learning rates \eqref{rates-good-dnn}. Moreover, these 
approximations $f_D$ are almost never global minima, or more precisely, 
$f_D$ is not a global minimum as soon as there exist a cubic cell $A$ containing two samples
$x_i$ and $x_j$
with different labels, i.e.~$y_i\neq y_j$. In fact, in this case, $f_D$ is not even a 
local minimum. To see this, assume without loss of generality 
that $x_i$ is one of the samples in $A$ with
$y_i \neq f_D(x_i)$. Considering $f_{D,\lb} := f_D + \lb b_i^+ \eins_{x_i+tB_\infty}^{(t/3)}$
for all $\lb\in [0,1]$ and $t:= \min\{r,\rho\}$ we then see that there is a continuous path in the parameter space
of $\cA_{4dn, 2n}$ that corresponds to the $\inorm\cdot$-continuous 
path $\lb\mapsto f_{D,\lb}$ in the set of functions
$\cA_{4dn, 2n}$ for which we have $\RD L{f_{D,\lb}} <  \RD L{f_D}$ for all $\lb\in (0,1]$.
In other words, $f_D$ is not a local minimum.
In this respect we note that this phenomenon also occurs to some extend in 
under-parameterized DNNs, at least for $d=1$. 
Indeed, if we consider $m:= 1$ and $s_n:= n^\g$, then 
$f_D, f_{D,\lb}\in \cA_{4dn^{\g d}, 2 n^{\g d}}$ for all sufficiently large $n$.
Now, the functions in $\cA_{4dn^{\g d}, 2 n^{\g d}}$ have ${\cal O}(d^2 n^{2\g d})$ many 
parameters and for $2\g d = \frac{2d}{2\a+d}< 1$, that is $\a > d/2 = 1/2$, we then see
that we have strictly less than ${\cal O}(\sqrt n)$ neurons with 
${\cal O}(n)$ parameters, while all the observations made so far still hold.

% \[\]
% \[\]

% \note{Note about uniform convergence and complexity based approaches for DNNs: 
% Finally, we compare our findings with recent complexities based approaches. }

%%%%%%%%%%%%%%%%%%%%%%%%%%%%%%%%%%%%%%%%%%%%%%%%%%%%%%%%%%%%%%%%%%%%%%%%%%%%%%
% References 
%%%%%%%%%%%%%%%%%%%%%%%%%%%%%%%%%%%%%%%%%%%%%%%%%%%%%%%%%%%%%%%%%%%%%%%%%%%%%%

\bibliographystyle{plain}
\bibliography{steinwart-article,steinwart-books,steinwart-mine,steinwart-proc,steinwart-preprint, bib_nicole}

%%%%%%%%%%%%%%%%%%%%%%%%%%%%%%%%%%%%%%%%%%%%%%
%% APPENDICES 
%%%%%%%%%%%%%%%%%%%%%%%%%%%%%%%%%%%%%%%%%%%%%%

%Appendices should be provided before Acknowledgements.

%%%%%%%%%%%%%%% APPENDIX %%%%%%%%%%%%%%%%%%%%%%%%%%%%%%%%%

%\newpage 

\appendix

%\begin{appendix} 
\section{Characterization of empirical risk minimizers}
\label{app:general}

In this section we briefly provide a full characterization of empirical risk minimizers that we use several times for proving our main results.  

\begin{lemma}[Characterization of ERMs] 
\label{result:hr-is-erm}
Let $Y$ be convex, $A\subseteq X$ be non-empty, $\ca A = (A_j)_{j\in J}$ be a finite
partition of $A$, and
 \begin{displaymath}
\HA := \biggl\{ \sum_{j\in J} c_j \eins_{A_j} : c_j \in Y \biggr\} \;.
 \end{displaymath}
Moreover, let $D=((x_1,y_1),\dots,(x_n,y_n)) \in (X\times Y)^n$
be a data set and let $L_A(x, y, t)=\eins_A(x)L(y,t)$, with $L$ being the least squares loss. 
Furthermore, denote   the number of samples whose covariates fall into cell $A_j$ by $N_j$, that is $N_j :=|\{i: x_i \in A_j\}|$. 
Then, for every $f^*\in \HA$ with representation $f^* =
\sum_{j\in J} c_j \eins_{A_j}$, the following statements are
equivalent:
\begin{enumerate}
\item The function $f^*$ is an empirical risk minimizer, that is 
\begin{displaymath}
  \RD {L_A} {f^*} = \min_{f\in \HA} \RD {L_A} f\, .
\end{displaymath}
\item
For all $j\in J$ satisfying $N_j \neq 0$ 
we have 
\begin{align}
\label{result:hr-is-erm-coeff-R}
 c_j =  \frac{1}{N_j}\sum_{i: x_i\in A_j} y_i  \;.
\end{align}
\end{enumerate}
\end{lemma}

\vspace*{1ex}

\begin{proofof}{of Lemma \ref{result:hr-is-erm}}
We first note that for an $f^*\in \HA$ with representation $f^*= \sum_{j\in J} c_j \eins_{A_j}$ we have
\begin{align*}
 \RD {L_A} {f^*}
= \frac 1n \sum_{i=1}^n \eins_A(x_i) L\bigl(y_i, f^*(x_i)\bigr) = \frac 1 n \sum_{j\in J} \sum_{i: x_i\in A_j} L(y_i, c_j)\, .
\end{align*}
Consequently, $f^*$ is an empirical risk minimizer, if and only
if
$c_j$ minimizes $\sum_{i: x_i\in A_j} L(y_i, \cdot )$ for all
$j\in J$. Now, if $N_j = 0$, the sum is empty, and hence there is nothing to consider. For $j\in J$ with $N_j$ we 
observe that 
\[  \sum_{i: x_i\in A_j} L(y_i, c_j )  
= N_j c_j^2 - 2c_j\biggl( \sum_{i: x_i\in A_j} y_i  \biggr) + \sum_{i: x_i\in A_j} y_i^2\,  ,
\]
which is minimized for $c_j$ given by \eqref{result:hr-is-erm-coeff-R}.
\end{proofof}

\section{Existence of properly aligned cubic partitioning rule}
\label{app:technical-stuff}

%\subsection{Proof of Theorem \ref{ruckel-theorem}}%Additional Results}

In this section we prove the existence of a properly aligned cubic partitioning rule.

\vspace*{1ex}

\begin{proofof}{Theorem \ref{ruckel-theorem}} 
Recall that 
cubic partitions $\ca B$ of $\R^d$ have a representation of the form \eqref{cubic-def}.
Now, to construct $\pi_{m,s}$ 
we will consider a finite set of candidate offsets 
$x_1^\dagger, \dots, x_K^\dagger\in \R^d$. For the construction of these offsets  
we write  $\delta := s/(m+1)$ and 
for $j \in \{0,\dots,m\}$  we  further define  
\begin{displaymath}
 z_{j}^\dagger := \Bigl( j  + \frac 12  \Bigr) \,  \delta \, .
\end{displaymath}
Now, our candidate offsets $x_1^\dagger, \dots, x_K^\dagger\in \R^d$ are exactly those 
vectors whose coordinates are taken from  $ z_{0}^\dagger, \dots, z_{m}^\dagger$. Clearly, 
this gives $K=(m+1)^d$.
Now let $\{x_1^*,\dots, x_m^*\}\in \potmset{[-1,1]^d}$. In the following, we will identify the 
offset $x_\ell^\dagger$ that leads to $\pi_{m,s}(\{x_1^*,\dots, x_m^*\})$
coordinate-wise. 
We begin by determining 
 its first 
coordinate $x_{\ell,1}^\dagger$.  To this end,  we   define 
\begin{displaymath}
   I_j :=  \bigcup_{k\in \Z} \bigr[k s + j\delta , \, k s +(j+1)\delta \bigl) \, .
\end{displaymath}
Our first goal is to show that $I_0,\dots,I_m$ are a partition of $\R$.
To this end, we fix an $x\in \R$. Then there exists a \emph{unique} $k\in \Z$ with 
$k s \leq x < (k +1) s$. Moreover, for $y:= x- k s\in [0,s)$, there exists a \emph{unique} $j\in \{0,\dots,m\}$
with $j\delta \leq y < (j+1)\delta$. Consequently, we have found 
$x\in [k s + j\delta , \, k s +(j+1)\delta )$. This shows $\R\subset I_0\cup\dots\cup I_m$, and the converse 
inclusion is trivial. Let us now fix
some $j,j'\in \{0,\dots,m\}$ and assume that there is 
an $x\in I_j \cap I_{j'}$.
Then there
exist $k,k'\in \Z$ such that 
\begin{align}\label{ruckel-lemma-h1}
   x \in \bigr[k s + j\delta, \, k s +(j+1)\delta \bigl) \, \cap \, \bigr[k' s + j'\delta, \, k' s +(j'+1)\delta \bigl)\, .
\end{align}
Since $(j+1)\delta \leq s$ and $(j'+1)\delta \leq s$, we conclude that 
$k s \leq x < (k +1) s$ and $k' s \leq x < (k' +1) s$. As observed above this implies $k = k'$.
Now consider $y:= x-k s\in [0,s)$. Then \eqref{ruckel-lemma-h1} implies 
\begin{displaymath}
   y \in \bigr[j\delta , \, (j+1)\delta \bigl) \, \cap \, \bigr[ j'\delta , \, (j'+1)\delta \bigl)\, ,
\end{displaymath}
and again we have seen above that this implies $j=j'$. This shows $I_j \cap I_{j'} =\emptyset$
 for all $j\neq j'$.

Let us now denote the first coordinate of $x_i^*$ by $x_{i,1}^*$. Then
 $D_{X,1}^* := \{x_{i,1}^*: i=1,\dots,m\}$ satisfies $|D_{X,1}^*|\leq m$
and since we have $m+1$ cells $I_j$, we conclude that there exists a $j_1^*\in \{0,\dots,m\}$ with 
$D_{X,1}^* \cap I_{j_1^*} = \emptyset$. We define  
\begin{displaymath}
   x_{\ell,1}^\dagger := z_{j_1^*}^\dagger =    \Bigl( j_1^* + \frac 12  \Bigr)    \, \delta \, .
\end{displaymath}
Next we repeat this construction for the remaining $d-1$ coordinates, so that we finally obtain 
$x_\ell^\dagger := (z_{j_1^*}^\dagger, \dots, z_{j_d^*}^\dagger)\in \R^d$ for indices $j_1^*,\dots,j_d^*\in \{0,\dots,m\}$ found by the above reasoning.

% We then define $\ca A = (A_j)_{j\in J}$ by the cubic partition given by \eqref{cubic-def} for the offset 
% $x^\dagger$ we have just found.

It remains to show that \eqref{proper-1} holds the cubic partition \eqref{cubic-def} with offset $x_\ell^\dagger$
and all $t>0$ with $t\leq \frac s{3m+3} = \delta /3$.
% Our next goal is to show that  \eqref{proper-1} holds for $t_1 := \d/3$. 
To this end, we fix an $x_i^*$.
Then its cell $B(x_i^*)$ is described by a unique $k := (k_1,\dots,k_d)\in \Z^d$, namely 
% \begin{displaymath}
%    A(x_i^*) = X \cap \bigl(x^\dagger + sk + [0,s)^d \bigr)\, ,
% \end{displaymath}
% see \eqref{cubic-def}. Let $B(x_i^*)$ be the cell of the partition  \eqref{cubic-def}
% that satisfies $B(x_i^*) \cap X = A(x_i^*)$. Then we have 
\begin{displaymath}
   B(x_i^*) 
= \bigl[x_{\ell,1}^\dagger+  k_1s,\,  x_{\ell,1}^\dagger+ (k_1+1)s\bigr) \times \dots\times  \bigl[x_{\ell,d}^\dagger+  k_ds, \, x_{\ell,d}^\dagger+ (k_d+1)s\bigr)\, .
\end{displaymath}
Let us now consider the first coordinate $x_{i,1}^*$. By construction we   know that $x_{i,1}^* \not\in I_{j_1^*}$
and 
\begin{align}\label{ruckel-lemma-h2}
    \Bigl( j_1^* + \frac 12  \Bigr)\cdot \delta   +  k_1s \,\leq\, x_{i,1}^* \,<\,        \Bigl( j_1^* + \frac 12  \Bigr) \cdot \delta + (k_1+1)s\, .
\end{align}
Now, $x_{i,1}^* \not\in I_{j_1^*}$ implies 
\begin{displaymath}
x_{i,1}^* \not\in  \bigr[(k_1 +1) s + j_1^*\delta, \, (k_1+1) s +(j_1^*+1)\delta\bigl) 
\end{displaymath}
Since the right hand side of \eqref{ruckel-lemma-h2} excludes the case
 $x_{i,1}^* \geq (k_1+1) s +(j_1^*+1)\delta $,   we  hence  find 
\begin{displaymath}
   x_{i,1}^*  < (k_1 +1) s + j_1^*\delta = x_{\ell,1}^\dagger +   (k_1 +1) s  - \delta/2 \, .
\end{displaymath}
This shows $x_{i,1}^* + r < x_{\ell,1}^\dagger +   (k_1 +1) s$ for all $r\in [-t,t]$.
To show that $x_{i,1}^* + r > x_{\ell,1}^\dagger +   k_1  s$ holds for all $r\in [-t,t]$
we first observe that $x_{i,1}^* \not\in I_{j_1^*}$ also implies 
\begin{displaymath}
x_{i,1}^* \not\in  \bigr[k_1 s + j_1^*\delta, \, k_1 s +(j_1^*+1)\delta\bigl) \, .
\end{displaymath}
Now, the left hand side of \eqref{ruckel-lemma-h2} excludes the case $x_{i,1}^* < k_1 s + j_1^*\delta$.
Consequently, we have 
\begin{align*}
 x_{i,1}^* \geq k_1 s +(j_1^*+1)\delta = x_{\ell,1}^\dagger +   k_1  s + \delta/2
\end{align*}
and this yields $x_{i,1}^* + r > x_{\ell,1}^\dagger +   k_1  s$   for all $r\in [-t,t]$.
% Analogously, we can show $x_{i,1}^* + r > x_{\ell,1}^\dagger +   k_1  s$ for all $r\in [-t,t]$.
Finally, by repeating these considerations for the remaining $d-1$ coordinates, we conclude that 
$ x_i^* + t B_\infty   \subset B(x_i^*)$.
\end{proofof}

\section{Learning properties of inflated histograms}
\label{app-histo-consistency-proofs}

In this section we provide the proofs of the results 
for the good and bad interpolating histogram rules from Section \ref{sec:inflated-histo}. To this end let us introduce some more notation. 
For a measurable set $A$ and a loss $L:  Y\times \mbr \to [0,\infty)$ we therefore introduce the loss $L_A: X\times Y \times \mbr \to [0,\infty)$ by
\begin{equation}   \label{eq:loss-dec}
L_A(x,y,t)) = \eins_A(x)L(y,t)\;.
\end{equation} 
Obviously, for any measurable function $f:X \to \mbr$ it holds 
\begin{equation}
\label{eq:chi}
\RP{L_{A}}{f} = \RP{L_{A}}{\eins_{A}f} \; . 
\end{equation}
Moreover, by linearity,  for every measurable sets $B\subset A$, the risk then decomposes as
\begin{equation}
\label{eq:LriskAB}
 \RP{L_A}{f}= \cR_{L_{A \setminus B}, P}(f) + \cR_{L_B, P}(f) \;.
\end{equation}
The next result shows that also the Bayes risk enjoys a similar decomposition.  

\begin{lemma}\label{result:bayes-part2}
Let $A,B\subset X$ be non-empty, disjoint, and measurable with $A\cup B = X$. Then we have 
\begin{displaymath}
  \RPB L = \RPB {L_A} + \RPB {L_B}\, .
\end{displaymath}
\end{lemma}

\begin{proofof}{Lemma \ref{result:bayes-part2}}
Basically, this is a consequence of the presence of the indicator functions $\eins_A, \eins_B$ in the definition of $L_A, L_B$, see \eqref{eq:loss-dec}. 
More precisely, 
there is a sequence of functions $f_n^A$ with $\{f_n^A \neq 0\}  \subset A$ such that 
$$ \RP{L_A}{f_n^A} \to \RPB {L_A} \; ,$$
as $n \to \infty$, and similarly for $A$ replaced by $B$.  Thus, for $f_n:= f_n^A + f_n^B$, one has
$$  \RPB L  \leq \lim_{n \to \infty} \RP{L}{f_n}= \lim_{n \to \infty}   \RP{L_A}{f_n^A}     + 
\lim_{n \to \infty}  \RP{L_B}{f_n^B} =  \RPB {L_A} + \RPB {L_B}\, .$$
Since the converse inequality is trivial, this proves the lemma.
\end{proofof}

%%%%%%%%%%%%%%%%%%%%%%%%%%%%%%%%%%%%%%%%%%%%%%%%%%%%%%%%%%%%%%%%%%%%%%%%%%%%%%%%%%%%%%%%%%%%%%%%%%%%%
%%%%%%%%%%%%%%%%%%%%%%%%%%%%%%%%%%%%%%%%%%%%%%%%%%%%%%%%%%%%%%%%%%%%%%%%%%%%%%%%%%%%%%%%%%%%%%%%%%%%%

\subsection{Preparatory Lemmata}

The next lemma provides a bound on the difference of the risks of two measurable functions.

\begin{lemma}\label{result:classifier-diff}
Let  $Y=[-1,1]$ and let $f_1, f_2: X \to Y$ be measurable 
functions. For  $A\subset X$ measurable and non-empty we define $L_A(x,y,t)=\eins_A(x)L(y, t)$ with $L$ being the least square loss. 
Then the following two inequalities hold: 
\begin{align*}
   \bigl|  \RP{L_A}{f_1}- \RP{L_A}{f_2} \bigr| &\leq 4\;P_X\bigl(A \cap \{ f_1\neq f_2 \}\bigr) \;, \\
|| \eins_A(f_1 - f_2)||^2_{L_2(P_X)} &\leq 4\;P_X\bigl(A \cap \{ f_1\neq f_2 \}\bigr) \;.
\end{align*} 
\end{lemma}

\vspace*{1ex}

\begin{proofof}{Lemma \ref{result:classifier-diff}}
We begin by proving the first inequality. To this end, we note that the definition of $L$ yields
\begin{align*}
 \RP{L}{f_1}- \RP{L}{f_2}   
 &=  \int_A \int_Y  (y-f_1(x))^2-(y-f_2(x))^2 \,   P(dy|x)dP_X(x)   \\
 &= \int_{A \cap \{ f_1\neq f_2 \}} \int_Y  (y-f_1(x))^2-(y-f_2(x))^2    P(dy|x)dP_X(x)  \; .
\end{align*}
Now observe that  $y, f_i(x) \in [-1,1]$ implies $(y-f_i(x))^2 \leq 4$. Moreover, we also have $(y-f_i(x))^2\geq 0$, and hence we conclude that 
 \[ \Bigl|  (y-f_1(x))^2-(y-f_2(x))^2  \Bigr| \leq 4  \;.    \]
Combining these considerations we find 
\begin{align*}
 \bigl| \RP{L}{f_1}- \RP{L}{f_2} \bigr|
 &\leq 
 \int_{A \cap \{ f_1\neq f_2 \}} \int_Y \Bigl| (y-f_1(x))^2-(y-f_2(x))^2  \Bigr|  P(dy|x)dP_X(x) \\
 &\leq 
 4 P_X\bigl( A \cap \{ f_1\neq f_2 \}\bigr)\; .
\end{align*}
The second inequality can be show similarly. Namely, we have 
\begin{align*}
 || \eins_A(f_1 - f_2)||^2_{L_2(P_X)}  &= \int_{A}(f_1(x) - f_2(x))^2 dP_X(x) \\
 &=\int_{A \cap \{ f_1\neq f_2 \}}(f_1(x) - f_2(x))^2 dP_X(x) \\
 &\leq 4P_X\bigl( A \cap \{ f_1\neq f_2 \}\bigr) \;,
\end{align*}
where we again used $f_i(x) \in [-1,1]$.
\end{proofof}

%%%%%%%%%%%%%%%%%%%%%%%%%%%%%%%%%%%%%%%%%%%%%%%%%%%%%%%%%%%%%%%%%%%%%%%%%%%%%%%%%%%%%%%%%%%%%%%

\begin{lemma}
\label{lem:useful}
Let $h: X \to \mbr$ be measurable, $A \subseteq X$, and  $L$ be the least-squares loss. Then we have the identity 
\[ \RP {L_A}{-h} - \RP {L_A}{-\fpb} 
    = \RP {L_A}{h} - \RPB{L_A} +   4\langle \eins_A f^*_{L,P}\; , h -  f^*_{L,P}  \rangle_2  \;.  \]
\end{lemma}

\begin{proof}[Proof of Lemma \ref{lem:useful}]
Given $x \in \cX$ and using 
\[  (a+b)^2 - (a-b)^2=   4ab \;, \]
we obtain for the difference of inner risks
\begin{align*}
& \int_Y (y+h(x))^2 \; P(dy|x) - \int_Y (y+f^*_{L,P}(x))^2 \; P(dy|x) \\ 
&= \int_Y (y+h(x))^2 - (y-h(x))^2 + (y-h(x))^2    \\
 & \qquad  -   (y+f^*_{L,P}(x))^2 + (y-f^*_{L,P}(x))^2 - (y-f^*_{L,P}(x))^2\; P(dy|x) \\
&=  \int_Y   (y-h(x))^2  -  (y-f^*_{L,P}(x))^2   + 4yh(x) - 4y f^*_{L,P}(x) \; P(dy|x) \;.
\end{align*}
Thus, since 
\[  f^*_{L,P}(x) = \int_Y y\; P(dy|x) \;, \]
we arrive at 
\begin{align*}
 &\RP {L_A}{-h} - \RP {L_A}{-\fpb} \\
 &= \RP {L_A}{h} - \RPB{L_A} + 4\int_A \int_Y  yh(x) - y f^*_{L,P}(x)   \; P(dy|x) P_X(dx) \\
 &= \RP {L_A}{h} - \RPB{L_A} + 4\langle \eins_A  f^*_{L,P}\; , h -  f^*_{L,P}\rangle_2  \;,
\end{align*}
i.e., we have shown the assertion.
\end{proof}

%%%%%%%%%%%%%%%%%%%%%%%%%%%%%%%%%%%%%%%%%%%%%%%%%%%%%%%%%%%%%%%%%%%%%%%%%%%%%%%%%%%%%%%%%%%%%%%%%%%%%

With these preparations we can now present the following key lemma
that shows that it suffices to understand the behavior of the good and bad interpolating histogram rules on $\D$
and the behavior of 
$h_{D,\ca A_D}^+$.

\begin{lemma}\label{lem:segments-of-gh}
 Let $L$ be the least squares loss, $P$ be a distribution on $X\times Y$
 with point spectrum $\D$, see \eqref{eq:delta}
 , and $D \in (X\times Y)^n$
 be a data set. Then for all $s\in (0,1]$ and all $\rho \geq 0$ the good 
 interpolating histogram rule satisfies
 \begin{align*}
  \RP L \gh - \RPB L 
 & \leq \RP {L_{\D}} \gh - \RPB {L_\D}  
 + 4 P_X(D_X^{+t}\setminus \D) \\
 &\quad +
   \RP {L} {h_{D,\ca A_D}^+} - \RPB {L} 
 \end{align*}
    where $D_X^{+t}$ is defined by \eqref{def:dxpt}. Moreover,  for all $s\in (0,1]$ and all $\rho \geq 0$ 
    the bad interpolating histogram rule satisfies
 \begin{align*}
     \snorm{\bh - \fpd}_{\Lx 2 \Px}^2 
    &\leq  \RP {L_{\D}} \bh - \RPB {L_\D}  +   4 P_X(D_X^{+t}\setminus \D)\\
  & \quad + \RP {L} {h_{D,\ca A_D}^+} - \RPB {L}\, .
 \end{align*}
\end{lemma}

\begin{proof}[Proof of Lemma \ref{lem:segments-of-gh}]
To simplify notation, we write $A:= D_X^{+t}\setminus \D$ and 
$B:= X \setminus(D_X^{+t}\cup \D)$. Note that this 
yields the partition $X = \D \cup A \cup B$. 
In addition, we have $\gh(x) = h_{D,\ca A_D}^+(x)$ for all $x\in X\setminus D_X^{+t}$. Using this in combination with $B\subset X\setminus D_X^{+t}$
as well as the risk decomposition formula \eqref{eq:LriskAB} and 
Lemma \ref{result:bayes-part2} we then  find
 \begin{align*}
  \RP L \gh - \RPB L 
 & = \RP {L_\D} \gh - \RPB {L_\D} \\
 & \quad +
   \RP {L_A} \gh - \RPB {L_A} \\
  & \quad +
   \RP {L_B} {h_{D,\ca A_D}^+} - \RPB {L_{B}} 
    \, .
 \end{align*}
 Moreover, Lemma \ref{result:classifier-diff} applied to $f_1 := \gh$ and
 $f_2 := \fpb$ 
 implies 
 \begin{align*}
  \RP {L_A} \gh - \RPB {L_A}
  \leq 4 P_X \bigl(A \cap \{\gh \neq \fpb  \}\bigr)
  \leq 4 P_X (A)\, .
 \end{align*}
In addition, we have 
\begin{align}
 \RP {L_B} {h_{D,\ca A_D}^+} - \RPB {L_B} \nonumber \label{lem:segments-of-gh-h1}
 & \leq 
 \RP {L_B} {h_{D,\ca A_D}^+} - \RPB {L_B} 
 + 
 \RP {L_{X\setminus B}} {h_{D,\ca A_D}^+} - \RPB {L_{X\setminus B}} \\
 & = \RP {L} {h_{D,\ca A_D}^+} - \RPB {L} \, ,
\end{align}
where again we used \eqref{eq:LriskAB} and 
Lemma \ref{result:bayes-part2}. Combining these 
estimates we then obtain the assertion for the good interpolating ERM.

  To prove the  inequality for the bad interpolating histogram rule, we consider the
  decomposition
   \begin{align*}
     \snorm{\bh - \fpd}_{\Lx 2 \Px}^2 
     & =  \int_{\D} \bigl( \bh - \fpb \bigr)^2\, d\Px \\
    &\quad +
     \int_{A} \bigl( \bh - \fpd \bigr)^2\, d\Px \\
    &\quad  +
     \int_{B} \bigl( \bh - \fpd \bigr)^2\, d\Px \, ,
   \end{align*}
    where in the first integral we  used $\fpd(x) = \fpb(x)$ for all $x\in \D$.
 Now, $\bh(x)\in [-1,1]$ and $\fpd(x) \in [-1,1]$ for all $x\in X$ gives 
 \begin{align*}
  \int_{A} \bigl( \bh - \fpd \bigr)^2\, d\Px \leq 4 P_X(A)\, .
 \end{align*}
     Moreover, by \eqref{eq-diff-good-bad-hr} we find 
 $\bh(x) = -\gh(x) = - h_{D,\ca A_D}^+(x)$ for all $x\in X\setminus D_X^{+t}$
 and thus also for all $x\in B$. In addition, $B\subset X\setminus \D$
 shows $\fpd(x) = -\fpb(x)$ for all $x\in B$. 
 Together, these considerations give 
    $ \bh(x) - \fpd(x) = -h_{D,\ca A_D}^+(x) + \fpb(x)$ for all $x\in B$, and consequently
    we obtain
    \begin{align*}
     \int_{B} \bigl( \bh - \fpd \bigr)^2\, d\Px 
      \leq  \int  \bigl( h_{D,\ca A_D}^+ - \fpb \bigr)^2\, d\Px  
     = \RP L{ h_{D,\ca A_D}^+} - \RPB L\, .
    \end{align*}
Combining these considerations finishes the proof.
\end{proof}

%%%%%%%%%%%%%%%%%%%%%%%%%%%%%%%%%%%%%%%%%%%%%%%%%%%%%%%%%%%%%%%%%%%%%%%%%%%%%%%%%%%%%%%%%%%%%%%%%%%%%
%%%%%%%%%%%%%%%%%%%%%%%%%%%%%%%%%%%%%%%%%%%%%%%%%%%%%%%%%%%%%%%%%%%%%%%%%%%%%%%%%%%%%%%%%%%%%%%%%%%%%

\subsection{Proof of Theorem \ref{results:main-erm}}
\label{app-proof-consistency-histos}

Throughout this section we assume that the general  assumptions of Theorem \ref{results:main-erm} are satisfied. In particular, 
 $D \in (X \times Y)^n$ is an i.i.d.~sample of size $n \geq 1$ and 
 $D_X:=\{x_1^*, ..., x_{m_n}^*\}\in \potm$ is  the set of input observations.
Moreover,  $(s_n)_{n \in \mbn}$ is a sequence with 
$s_n \rightarrow 0$ as well as $\frac{\ln (n s_n^d)}{n s_n^d}\to 0$ as $n \to \infty$.

%%%%%%%%%%%%%%%%%%%%%%%%%%%%%%%%%%%%%%%%%%%%%%%%%%%%%%%%%%%%%%%%%%%%%%%%%%%%%%%%%%%%%%%%%%%%%%%%%%%%%%%%%%%%%%%%%%%%%%%%%%%%%
%%%%%%%%%%%%%%%%%%%%%%%%%%%%%%%%%%%%%%%%%%%%%%%%%%%%%%%%%%%%%%%%%%%%%%%%%%%%%%%%%%%%%%%%%%%%%%%%%%%%%%%%%%%%%%%%%%%%%%%%%%%%%

\subsubsection{The good interpolating histogram rule}
\label{sec:good-consistency}

We begin by introducing the basic strategy of our proof. 
To this end, consider the good interpolating histogram rule from Example \ref{good-guy} 
with representation 
\[  \gh =  h_{D,\ca A_D}^+  + \sum_{i=1}^{m} b_i \eins_{\{x_i^* + tB_\infty\}} \in \cF_{s,m}\;. \]
In view of 
\eqref{eq:excess-risk} it suffices to consider the excess risk of $\gh$.
Now observe that in the case \emph{i)}, i.e.~for $\rho=0$, we have $t=0$ and thus  $D_X^{+t} = D_X$.
Since $P_X(D_X\setminus \D )= 0$ by the definition of $\D$, we then 
find by 
 Lemma \ref{lem:segments-of-gh} that
 \begin{align*}
  \RP L \ghon - \RPB L 
  \leq  \cR_1 ( h_{D,\ca A_D}^+) +\cR_2(\ghon)\, ,
 \end{align*}
 where 
 \begin{align*}
   \cR_1 ( h_{D,\ca A_D}^+) & := \RP {L} {h_{D,\ca A_D}^+} - \RPB {L} \\
   \cR_2(\ghon) &:=  \RP {L_{\D}} \ghon - \RPB {L_\D} \, .
 \end{align*}
Moreover, in the case  \emph{ii)}, 
i.e.~for $\rho=\rho_n>0$ the distribution $P$
satisfies Assumption \ref{ass:proba-2}, which ensures $\Delta = \emptyset$.
The latter implies $\cR_2(\ghon) = 0$, and therefore  we
find by 
 Lemma \ref{lem:segments-of-gh} that
 \begin{align*}
  \RP L \gh - \RPB L 
  \leq   
  4 P_X (D_X^{+t}\setminus \D) +
  \cR_1 ( h_{D,\ca A_D}^+) \, .
 \end{align*}
 Moreover, by Assumption \ref{ass:proba-2} and $t\leq \rho = \rho_n$ we obtain 
 \begin{align}\label{ineq-inflat-bound}
 P_X (D_X^{+t}\setminus \D) 
 \leq 
 P_X (D_X^{+\rho_n})
 \leq 
 \sum_{i=1}^n P_X (x_i + \rho_n B_\infty) 
 \leq
 n \varphi(\rho_n) \to 0 \, ,
 \end{align}
 and consequently, it suffices to bound $\cR_1 ( h_{D,\ca A_D}^+)$. 
Therefore, 
the rest of this subsection is devoted to bounding 
$\cR_1 ( h_{D,\ca A_D}^+)$  and   $\cR_2(\ghon)$ individually.

\vspace*{2ex}
\noindent
{\bf Bounding $\cR_1 (h_{D,\ca A_D}^+)$.} 
Thanks to Proposition \ref{prop:cons-reg},  we already know that 
\begin{align*}
 \cR_1 (h_{D,\ca A_D}^+)= \RP {L}{h_{D,\ca A_D}^+} - \RPB{L} \to 0
\end{align*}
 in probability
for $n\to \infty$.

\vspace*{2ex}
\noindent
{\bf Bounding $\cR_2(\ghon)$.}
If $\D = \emptyset$ we obviously have $\cR_2(\ghon) = 0$, and hence 
we assume $\D \neq \emptyset$ in the following.
In this case, $\D$ can be at most countable, and 
therefore we fix an at most countable enumeration $(\tilde x_j)_{j\in J}$ 
of $\D$, i.e.
\begin{displaymath}
\D =   \bigcup_{j \in J} \{\tilde x_j \} \;.
\end{displaymath}
% where $(\tilde x_j)_{j\in J}$ is an at most countable enumeration of $\D$.
Let us further fix an $\eps>0$ and a finite subset $\D_0 \subset \D$ such that $P_X(\D\setminus \Delta_0) \leq \eps$.
With the help of \eqref{eq:LriskAB} and Lemma \ref{result:bayes-part2} we then observe 
that
\begin{equation}
\label{eq:dec-00}
 \RP{L_\Delta}{\ghon} - \RPB{L_\Delta} 
 = \RP{L_{\Delta_0}}{\ghon} - \RPB{L_{\Delta_0}}
     \; +  \;  \RP{L_{\D\setminus \Delta_0}}{\ghon} - \RPB{L_{\D\setminus \Delta_0}} \;.
\end{equation}
Since $Y=[-1,1]$ is bounded the second difference can be bounded by 
\begin{align}
\label{eq:part401}
 \bigl|\RP{L_{\D\setminus \Delta_0}}{\ghon} - \RPB{L_{\D\setminus \Delta_0}} \bigr|
& = \int_{X \times Y} \eins_{\D\setminus \Delta_0} (x) (\ghon(x)- f^*_P(x))^2 \; dP(x,y) \no \\
& \leq 4 \; P_X(\D\setminus \Delta_0) \no \\ 
&<  4 \eps \, .
\end{align}
% for some $c<\infty$ if we choose $|\D_0|$ sufficiently large.  
Our next step is to bound the first difference in \eqref{eq:dec-00}.  To this end, 
we write 
\[ J_0 := \{j\in J: \tilde x_j\in \D_0\}  \;,  \]
$C_j := \{\tilde x_j\}$ for  $j\in J_0$, and $\cC := (C_j)_{j\in J_0}$. 
Then $\cC$ is a finite  partition of $\D_0$, and we set 
\begin{align}\label{cf_cC_on_D0}
  \cF_\cC := \biggl \{ \sum_{j \in J_0} c_j \eins_{C_j} : c_j \in Y   \biggr \}  \;.
\end{align}
Since all $C_j$ are singletons, 
every measurable function $f:X\to Y$ satisfies $\eins_{\D_0} f \in \cF_\cC$.  
We thus conclude that 
$f_D := \eins_{\D_0}\ghon \in \cF_\cC$, too. Moreover, by \eqref{eq:chi} we know
\begin{align} 
\label{fd_eq_ghon}
 \RP{L_{\Delta_0}} {f_D} = \RP{L_{\Delta_0}} \ghon \, .
\end{align}

Our next goal is to  show that $f_D$ minimizes the empirical risk over $\cF_\cC$ with respect to $L_{\Delta_0}$.
To this end, we fix a $j \in J_0$ for which we have    $N_j :=|\{i: x_i \in C_j\}| > 0$. 
Since $\ghon$ interpolates $D$ by construction, Proposition \ref{suff-interpol-erm} then gives

\begin{equation}
\label{eq:ERMD0}
  f_D(\tilde x_j)  = \eins_{\D_0}(x)\ghon(\tilde x_j) =\ghon(\tilde x_j)   =  \frac{1}{N_j}\sum_{i: x_i \in C_j} y_i \, .
\end{equation}
Thus, Lemma \ref{result:hr-is-erm} shows that $f_D$ is indeed an empirical risk minimizer with respect 
to $L_{\D_0}$ and $\cF_\cC$.

Our next goal is to apply Theorem \ref{theo:generic-oracle},
which holds for \emph{all} ERM with respect 
to $L_{\D_0}$ and $\cF_\cC$,
to our specific ERM  $f_D$. To this end, 
we first observe, as in the proof of  Corollary \ref{prop:oracle-partitioning}, that 
since $L$ is the least squares loss, the assumptions
\eqref{sup-bound} and \eqref{var-bound} 
of Theorem \ref{theo:generic-oracle} are satisfied for $L_{\D_0}$ with $\vartheta  = 1$, $B=4$, and $V=16$. 
Moreover, our assumption $Y= [-1,1]$ ensures that $L_{\D_0}$ is locally Lipschitz 
continuous with $|L_{\D_0}|_{1,1} \leq 4$. In addition, we have 
\begin{equation*}
  \cN(\cF_\cC, ||\cdot||_\infty, \e) \leq (2/\e)^{|\D_0|}  \;. 
\end{equation*}  
Applying Theorem \ref{theo:generic-oracle}  and optimizing the resulting oracle inequality
with respect to $\e$ like at the end of the proof of Corollary \ref{prop:oracle-partitioning},
we then see that,  for all $n\geq 1$ and $\tau >0$,
\begin{align*}
% \sup_{i=1,...,K}(
\RP {L_{\D_0}} {f_D} - \RPB {L_{\D_0}}
\leq 4 \bigl( \RPxB {L_{\D_0}} {\cF_\cC}-\RPB {L_{\D_0}}\bigr) 
+
1024 \,\frac{ \tau   }n  +  512 \,\frac{  |\D_0|  }n\left( 1+\ln \left(\frac{n}{|\D_0|} \right) \right) 
\end{align*}
holds with probability $P^n$ not less than $1- e^{-\tau}$. 
Now, to bound the approximation error term, 
we note that 
\begin{align*}
% \bigl\{f_{|\D_0}: f\in \cF_\cC\bigr\} = \bigl\{  f\, \bigl|\, f:\D_0\to Y \mbox{ measurable }\bigr\}
 \cF_\cC = \bigl\{ f:X\to Y \,\bigl|\, f \mbox{ measurable and } f(x) = 0 \mbox{ for all } x\not\in \D_0\bigr\}  
 \, ,
\end{align*}
and hence we easily find 
\begin{align*}%\label{eq:fclass}
   \RPB{L_{\Delta_0}} 
   = \inf_{f:X\to Y}\RP {L_{\Delta_0}}f   
   = \inf_{f:X\to Y} \RP {L_{\Delta_0}}{\eins_{\D_0}f}    
    = \RPxB {L_{\Delta_0}}{\cF_\cC}\, .
\end{align*}
% Combining these considerations with \eqref{fd_eq_ghon} and 
Setting $\tau:= \ln (n)$ we conclude that 
\begin{align}\label{erm_orac_ineq_D0}
% \sup_{i=1,...,K}(
\RP {L_{\D_0}} {f_D} - \RPB {L_{\D_0}}
\leq 
1024 \,\frac{ \tau   }n  +  512 \,\frac{  |\D_0|  }n\left( 1+\ln \left(\frac{n}{|\D_0|} \right) \right) 
\end{align}
holds with probability $P^n$ not less than $1- 1/n$. 
For later use note that this oracle inequality actually holds 
for \emph{all} ERM respect 
to $L_{\D_0}$ and $\cF_\cC$, since so does Theorem \ref{theo:generic-oracle}
and we have not used any property of our specfic ERM $f_D$ to derive \eqref{erm_orac_ineq_D0}.
Finally, combining this with \eqref{eq:dec-00}, \eqref{eq:part401}, \eqref{fd_eq_ghon}, and the obvious 
$\RP {L_{\D_0}} {\ghon} - \RPB {L_{\D_0}}\geq 0$,
 we conclude that 
 \begin{align*}
  \cR_3(\ghon) =  \RP{L_\Delta}{\ghon} - \RPB{L_\Delta}  \to 0
 \end{align*}
in probability for $n\to \infty$.

%%%%%%%%%%%%%%%%%%%%%%%%%%%%%%%%%%%%%%%%%%%%%%%%%%%%%%%%%%%%%%%%%%%%%%%%%%%%%%%%%%%%%%%%%%%%%%%%%%%%%%%%%%%%%%%%%%
%%%%%%%%%%%%%%%%%%%%%%%%%%%%%%%%%%%%%%%%%%%%%%%%%%%%%%%%%%%%%%%%%%%%%%%%%%%%%%%%%%%%%%%%%%%%%%%%%%%%%%%%%%%%%%%%%%

\subsubsection{The bad interpolating histogram rule}
\label{sec:bad-consistency}

In this subsection
we consider  the   bad interpolating histogram rule from Example \ref{bad-guy} 
with representation 
\[  \bh =  h_{D,\ca A_D}^-  + \sum_{i=1}^{m} b_i \eins_{x_i^* + t B_\infty  } \in \cF_{s, m}\;. \]
Now observe that in the case \emph{i)} of Theorem \ref{results:main-erm}, i.e.~for $\rho=0$, we have $t=0$ and thus  $D_X^{+t} = D_X$.
Since $P_X(D_X\setminus \D )= 0$ by the definition of $\D$, we then 
see by Lemma \ref{lem:segments-of-gh} and Proposition \ref{prop:cons-reg} 
that it suffices to show that
\begin{align}\label{bad-erm-on-delta-h1}
 \RP{L_\Delta}{\bhon } - \RPB {L_\D} \to 0
\end{align}
in probability for $n\to \infty$. 
To this end, we 
fix an $\eps>0$ and a  finite $\D_0\subset \D$ with $P_X(\D\setminus \D_0 ) \leq \eps$.
Then we note that the decomposition \eqref{eq:dec-00} and the estimate \eqref{eq:part401} 
for $\ghon$
also holds for $\bhon$. Consequently, it suffices to bound the term
\begin{align*}
 \RP {L_{\D_0}} {\bhon} - \RPB {L_{\D_0}}\, .
\end{align*}
To this end, recall 
that $\ghon$ and $\bhon$ are both interpolating predictors, and hence
we have 
\begin{align}\label{ghon_eq_bhon_on_samples}
 \bhon(x_i) = \ghon(x_i)
\end{align}
for all samples $(x_i, y_i)$ of $D$, and thus in particular for all samples $(x_i, y_i)$ of $D_0$ with 
$x_i \in \D$. 
Let us define 
$f_D := \eins_{\D_0}\bhon$.
Combining \eqref{ghon_eq_bhon_on_samples} with  
\eqref{eq:ERMD0} we see that $f_D$ is an empirical risk minimizer 
over the hypotheses set $\cF_\cC$ defined \eqref{cf_cC_on_D0}  with respect to $L_{\Delta_0}$.
Since 
\eqref{erm_orac_ineq_D0} has been shown for \emph{all} ERM 
respect 
to $L_{\D_0}$ and $\cF_\cC$ we thus find
\begin{align*}
 \RP {L_{\D_0}} {\bhon} - \RPB {L_{\D_0}} =  \RP {L_{\D_0}} {f_D} - \RPB {L_{\D_0}}   \to 0
\end{align*}
in probability for $n\to \infty$. 
This finishes the proof in the case \emph{i)} of Theorem \ref{results:main-erm}.
Moreover, in the case  \emph{ii)}, 
i.e.~for $\rho=\rho_n>0$ the distribution $P$
satisfies Assumption \ref{ass:proba-2}, which ensures $\Delta = \emptyset$.
In combination with 
 Lemma \ref{lem:segments-of-gh} the  latter implies 
 \begin{align*}
     \snorm{\bh - \fpd}_{\Lx 2 \Px}^2 
    \leq      4 P_X(D_X^{+t}\setminus \D)
   + \RP {L} {h_{D,\ca A_D}^+} - \RPB {L}\, .
 \end{align*}
Now, the first term has already been bounded in 
\eqref{ineq-inflat-bound} and the excess risk of 
$h_{D,\ca A_D}^+$ can again be bounded by Proposition \ref{prop:cons-reg}.

%%%%%%%%%%%%%%%%%%%%%%%%%%%%%%%%%%%%%%%%%%%%%%%%%%%%%%%%%%%%%%%%%%%%%%%%%%%%%%%%%%%%%%%%%%%%%%%%%%%%%%%
%%%%%%%%%%%%%%%%%%%%%%%%%%%%%%%%%%%%%%%%%%%%%%%%%%%%%%%%%%%%%%%%%%%%%%%%%%%%%%%%%%%%%%%%%%%%%%%%%%%%%%%

\subsection{Proof of Theorem \ref{results:main-erm2} (Learning Rates)}
\label{app-proof-rates-histos}

In the following  we suppose that all assumption of Theorem \ref{results:main-erm2}  are satisfied. 

Let us first prove the assertions for the good interpolating histogram rule. 
To this end, we first recall that Assumption \ref{ass:proba-2}
implies $\D = \emptyset$.
By \eqref{eq:excess-risk} and 
Lemma \ref{lem:segments-of-gh} we then  obtain 
 \begin{align*}
 ||\ghn  - \fpb||_{L_2(P_X)}^2
 = \RP L \gh - \RPB L 
  \leq 
  4 P_X(D_X^{+t}\setminus \D) 
+
   \RP {L} {h_{D,\ca A_D}^+} - \RPB {L} \, .
 \end{align*}
Now, \eqref{ineq-inflat-bound} shows 
\begin{align}\label{ineq-inflat-bound-2}
  P_X (D_X^{+t}\setminus \D) 
 \leq 
 n \varphi(\rho_n)
 \leq 
 \ln(n) \, n^{-2/3}
 \leq 
 \ln(n) \, n^{-2\alpha \gamma}
 \, .
\end{align}
Moreover, by Theorem \ref{ruckel-theorem} we know that 
$|\Image(\pi_{m,s})|\leq (m+1)^d \leq 2^d n^d$ for all $m\leq n$.
Consequently, applying  Proposition \ref{prop:rates-reg} with $c=2^d$ and $\beta:= d$ 
we find 
\begin{equation}
\label{eq:riskplus}
   \RP{\Lc}{h_{D,\cA_D}^+} - \RxB{\Lc}{P} \leq c_{d,\alpha }\ln(n  )  \,  n^{-2\alpha \gamma } 
\end{equation}   
with probability $P^n$ at least $1- 2^dn^{1+d} e^{-n^{d \gamma}}$, where 
$c_{d,\alpha }>0$ is a constant only depending on $d$, $\alpha$, and $|f^*_{L,P}|_\alpha$.
Combining this with \eqref{ineq-inflat-bound-2} we then obtain \eqref{rates-good}.

Finally, inequality \eqref{rates-good} for the 
bad interpolating histogram rule 
follows analogously, since in this case
Lemma \ref{lem:segments-of-gh} shows 
 \begin{align*}
||\bhn  - \fpd||_{L_2(P_X)}^2
  \leq 
  4 P_X(D_X^{+t}\setminus \D) 
+
   \RP {L} {h_{D,\ca A_D}^+} - \RPB {L} \, .
 \end{align*}

\section{Learning properties of approximating neural networks} 
\label{app-dnn-proofs}

\subsection{Auxiliary Results on Functions that can be represented by DNNs}

In this section we present some results on algebraic properties of the set of 
functions that can be represented by DNNs. We particularly  focus on the
network sizes required to perform   algebraic transformations of such functions.

To this end, recall that throughout this work we solely consider the 
ReLU-activation function $\sigma := |\cdot|_+$ and its shifted extensions \eqref{eq:shifted-activation}. 
Given an input dimension $d$, a depth $L\geq 2$, and a
width vector $(p_1,\dots,p_{L-1}) \in \mbn^{L-1}$,
 a function $f\in \cA_{p_1,\dots,p_{L-1}}$ is then of the form 
 \eqref{eq:basic-rep}, i.e.
 \begin{align*}
  f(x) &= H_L\circ H_{L-1}\circ \dots\circ H_1(x)  \;, \quad x \in \mbr^{p_0} \;,%\no \\
%  &= A^{(L)}\sigma_{b^{(L-1)}}A^{(L-1)}\sigma_{b^{(L-2)}} \cdot \dots\cdot A^{(2)}\sigma_{b^{(1)}}A^{(1)} (x)  +  b^{(L)} \;, 
\end{align*} 
 where each layer $H_l$, $l=1,\dots, L$,
 is of the form \eqref{eq:layer-repres}, where we drop the index for the activation to ease notation. Specifically,  
 each layer can be represented by a $p_l \times p_{l-1}$
 weight matrix $A^{(l)}$ with $p_0 := d$ and $p_L := 1$ and a shift vector $b^{(l)}\in \R^{p_l}$, 
and the last layer $H_L$ has the identity as an activation function. 
In the following, we thus 
 say that the network $f$ is represented by $(\frk A, \frk B)$, where $\frk A:=  (A^{(1)}, \dots,  A^{(L)})$
and $\frk B:=  (b^{(1)}, \dots,  b^{(L)})$. 
For later use we emphasize that $p_L = 1$ implies $b^{(L)}\in \R$.
Moreover note
 that 
each pair $(\frk A, \frk B)$ determines a neural network,
but in general, a neural network, if viewed as a function,  can be described by more than one such pair.

Now, our first lemma describes the changes in the representation when manipulating 
a single neural network.

\begin{lemma}\label{result:dnn-algebra-1}
%\fix{\noteis{Habe das alte lemma E.1 in zwei zerlegt, damit es etwas aufger\"aumter aussieht.}}
Let $d\geq 1$, $L\geq 2$,  and  $p:=(p_1,\dots,p_{L-1}) \in \mbn^{L-1}$. 
Moreover, let 
 $f\in \archx{p}$ be a neural network with
 representation $\frk A:=  (A^{(1)}, \dots,  A^{(L)})$
and $\frk B:=  (b^{(1)}, \dots,  b^{(L)})$.
Then the following statements hold true:
\begin{enumerate}
 \item For all $\alpha \in \mbr$ and $c \in \mbr$ we have 
$\alpha  f + c \in \archx{p}$  with 
% 
% Then for $\alpha \in \mbr$ and $c \in \mbr$, the function $\alpha  f + c \in \archx{d,m,1}$ has 
representation 
\[ \bigl(A^{(1)}, \dots,  A^{(L-1)}, \alpha A^{(L)}\bigr)   
\qquad\qquad
\mbox{ and }
\qquad \qquad
\bigl(b^{(1)}, \dots,  b^{(L-1)},\alpha b^{(L)}+c\bigr) \, .\]
\item We have $\relu f\in \archx{p,1}$ 
with representation 
\[   \bigl(A^{(1)}, \dots,  A^{(L)}, 1\bigr)  
\qquad\qquad
\mbox{ and }
\qquad \qquad
\bigl(b^{(1)}, \dots,  b^{(L)},  0\bigr)    \;.\]
\end{enumerate}
\end{lemma}

\begin{proofof}{Lemma \ref{result:dnn-algebra-1}}
\ada i This immediately follows from the representation \eqref{eq:basic-rep}
and the fact that $H_L$ does not have an activation function.

\ada {ii} Let $\tilde H_1, \dots, \tilde H_{L+1}$ be the
layers of the neural network $\tilde f$ given by the new representation. 
Then we have $H_l = \tilde H_l$ for all $l=1,\dots, L-1$ as well
as $\tilde H_L = |H_L|_+$ and $\tilde H_{L+1} = \id_\R$.
Applying the  representation \eqref{eq:basic-rep} for $f$ and $\tilde f$ then gives the assertion.
\end{proofof}

Our next lemma describes a possible representation of 
the sum of two nets with the same depth $L$.

\begin{lemma}\label{result:dnn-algebra-2}
Let $d\geq 1$, $L\geq 2$,  and   $\markone p:=(\markone p_1,\dots,\markone p_{L-1}) \in \mbn^{L-1}$ and
$\marktwo p:=(\marktwo p_1,\dots,\marktwo p_{L-1}) \in \mbn^{L-1}$ be two width vectors.
Then for all $\markone f\in \archx {\markone p}$ and $\marktwo f\in \archx {\marktwo p}$
we have 
%\fix{\noteis{Ich habe mal die Bezeichnungen der beiden Netze mit Macros gemacht, siehe locmacros.tex. Dann können wir optisch rumspielen, um zu sehen, was am besten aussieht. }}
\begin{align*}
 \markone f + \marktwo f \in \archx{\markone p + \marktwo p}\, .
\end{align*}
In addition, if $(\markone {\frk A},\markone {\frk B})$ and 
$(\marktwo {\frk A},\marktwo {\frk B})$ are representations of $\markone f$ and $\marktwo f$, then $\markone f + \marktwo f$ has the representation 
$\frk A:=  (A^{(1)}, \dots,  A^{(L)})$
and $\frk B:=  (b^{(1)}, \dots,  b^{(L)})$ defined  by
\begin{align*}
 A^{(1)} &:= \biggl( \begin{array}{c}\markone{A}^{(1)}\\ \marktwo{A}^{(1)} \end{array} \biggr) 
 \in \R^{(\markone m_1+\marktwo m_1)\times d}\, ,  
 &  b^{(1)} &:= \biggl( \begin{array}{c}\markone{b}^{(1)}\\ \marktwo{b}^{(1)} \end{array} \biggr) 
 \in \R^{\markone m_1+\marktwo m_1} \, ,
\end{align*}
as well as 
\begin{align*}
 A^{(l)} &:= \biggl( \begin{array}{cc}\markone{A}^{(l)} & 0\\ 0& \marktwo{A}^{(l)} \end{array} \biggr) 
 \in \R^{(\markone m_l+\marktwo m_l)\times (\markone m_{l-1}+\marktwo m_{l-1})}  \, ,
 &  b^{(l)} &:= \biggl( \begin{array}{c}\markone{b}^{(l)}\\ \marktwo{b}^{(l)} \end{array} \biggr) 
 \in \R^{\markone m_l+\marktwo m_l} \, ,&   
\end{align*}
 for all $l=2,\dots, L-1$ and 
  \begin{align*}
 A^{(L)} &:= \bigl( \begin{array}{cc}\markone{A}^{(L)}  & \marktwo{A}^{(L)}\end{array} \bigr) 
 \in \R^{\markone m_L+\marktwo m_L}  \, ,
 &  b^{(L)} &:= \markone{b}^{(L)} +\marktwo{b}^{(L)} 
 \in \R\, . \\
\end{align*}
\end{lemma}

\begin{proofof}{Lemma \ref{result:dnn-algebra-2}}
Let $\markone H_{1}, \dots , \markone H_L$ be the layers of $\markone f$ and 
$\marktwo H_{1}, \dots , \marktwo H_L$ be the layers of $\marktwo f$.
For $l =1,\dots,L$, we further  introduce the concatenation of layers
\[  \markone W_l:= \markone H_{l}\circ \dots \circ \markone H_1 
\qquad\qquad 
\mbox{ and }
\qquad\qquad
 \marktwo W_l:= \marktwo H_{l}\circ \dots \circ \marktwo H_1 
\, .   \]
Moreover, for $l =1,\dots,L$, let $H_l$ be the layer given by $A^{(l)}$
and $b^{(l)}$ and $W_l:= H_{l}\circ \dots \circ H_1 $.
Since the last layers of $\markone f$ and $\marktwo f$ do not have an activation function,
we then find 
\begin{align*}
  (\markone f+ \marktwo f)(x) &=  \markone A^{(L)} \cdot \markone  W_{L-1}(x)  + \markone b^{(L)}    +    \marktwo A^{(L)} \cdot \marktwo  W_{L-1}(x) + \marktwo b^{(L)} \\
  &=   \bigl( \begin{array}{cc}\markone{A}^{(L)}  & \marktwo{A}^{(L)}\end{array} \bigr)  \cdot  
      \biggl( \begin{array}{c}\markone  W_{L-1}(x)  \\\marktwo  W_{L-1}(x)  \end{array} \biggr)    + \markone b^{(L)} + \marktwo b^{(L)}\\
&=   {A}^{(L)}  \cdot
      \biggl( \begin{array}{c}\markone  W_{L-1}(x)  \\\marktwo  W_{L-1}(x)  \end{array} \biggr)      +   b^{(L)} 
\end{align*}
for all $x\in \R^d$.
Similarly,  for all $l=2,\dots,L-1$ and  all $x\in \R^d$ we have 
\begin{align*}
\biggl( \begin{array}{c}\markone  W_{l}(x)  \\\marktwo  W_{l}(x)  \end{array} \biggr)   
   = \biggl( \begin{array}{c}  \markone H_l \circ  \markone  W_{l-1}(x)  \\  \marktwo H_l \circ \marktwo  W_{l-1}(x)  \end{array} \biggr) 
   &= \biggl( \begin{array}{c} \bigl| \markone A^{(l)} \cdot \markone  W_{l-1}(x) + \markone b^{(l)} \bigr|_+ \\ \bigl| \marktwo A^{(l)} \cdot  \marktwo  W_{l-1}(x) + \marktwo b^{(l)} \bigr|_+ \end{array} \biggr) \\
   &= \left | \biggl( \begin{array}{cc}\markone{A}^{(l)} & 0\\ 0& \marktwo{A}^{(l)} \end{array} \biggr) 
    \biggl( \begin{array}{c}\markone  W_{l-1}(x)  \\ \marktwo  W_{l-1}(x)  \end{array} \biggr)   +   
    \biggl( \begin{array}{c}\markone  b^{(l)}  \\ \marktwo  b^{(l)}  \end{array} \biggr) \right |_+ \\
    &=  \biggl|{A}^{(l)}  \cdot
      \biggl( \begin{array}{c}\markone  W_{l-1}(x)  \\\marktwo  W_{l-1}(x)  \end{array} \biggr)      +   b^{(l)}  \biggr|_+  \, .
\end{align*}
Finally, for the first layer and  all $x\in \R^d$ we obtain
\begin{align*}
\biggl( \begin{array}{c}\markone  W_{1}(x)  \\\marktwo  W_{1}(x)  \end{array} \biggr)   
   = \biggl( \begin{array}{c}  \markone H_1 (x)  \\  \marktwo H_1 (x)  \end{array} \biggr) 
   = \biggl( \begin{array}{c} \bigl| \markone A^{(1)} \cdot x + \markone b^{(1)} \bigr|_+ \\ \bigl| \marktwo A^{(1)} \cdot x + \marktwo b^{(1)} \bigr|_+ \end{array} \biggr) 
   &= \left | \biggl( \begin{array}{cc}\markone{A}^{(1)}  \\   \marktwo{A}^{(1)} \end{array} \biggr) 
    \cdot x  +   
    \biggl( \begin{array}{c}\markone  b^{(1)}  \\ \marktwo  b^{(1)}  \end{array} \biggr) \right |_+ \\
    &=  \bigl|{A}^{(1)} 
      \cdot x     +   b^{(l)}  \bigr|_+  \, .
\end{align*}
Combining these results gives $W_l =(\markone  W_{l}, \marktwo  W_{l} )^T$ for all $l=1,\dots,L$, i.e.~we have found the assertion.
\end{proofof}

%%%%%%%%%%%%%%%%%%%%%%%%%%%%%%%%%%%%%%%%%%%%%%%%%%%%%%%%%%%%%%%%%%%%%%%%%%%%%%%%%%%%%%%%%%%%%%%%%%
%%%%%%%%%%%%%%%%%%%%%%%%%%%%%%%%%%%%%%%%%%%%%%%%%%%%%%%%%%%%%%%%%%%%%%%%%%%%%%%%%%%%%%%%%%%%%%%%%%

\subsection{Approximating Step Functions by DNNs}
\label{app-neuron-lego}

In this section we collect the main pieces to approximate histograms with DNNs. 
The first lemma, which  is a longer and more detailed version of Lemma \ref{result:one-bump-d}, shows how to approximate an indicator 
function on a multidimensional interval by a small ReLU-DNN with two \emph{hidden} layers.

\begin{lemma}\label{result:one-bump-d-v2}
Let $d\geq 1$ and let $z_1 = (z_{1,1},\dots,z_{1,d})\in \Rd$  and $z_2 = (z_{2,1},\dots,z_{2,d})\in \Rd$ be two vectors with 
$z_{1} < z_{2}$. Moreover,  let $\e>0$ satisfy
\begin{displaymath}
   \e < \min \Bigl\{ \frac{z_{2,i} - z_{1,i}} 2: i=1,\dots,d\Bigr\}
\end{displaymath}
and define 
\begin{align*}
 A^{(1)} &:= \frac 1 \e \biggl( \begin{array}{c} -I_d \\ I_d   \end{array} \biggr)
 &
 b^{(1)} &:= \frac 1 \e \biggl( \begin{array}{c} z_1 + \e \\ -z_2 - \e   \end{array} \biggr) \\
 A^{(2)} &:= (-1, -1, \dots, -1) \in \R^{2d}
  &
 b^{(2)} &:= 1 \\
 A^{(3)} &:= 1 
  &
 b^{(3)} &:= 0 \, ,
\end{align*}
where $I_d$ denotes the $d$-dimensional identity matrix, and
$A^{(3)}, b^{(2)}, b^{(3)}\in \R$.
Then the neural network $f_\e:\R^d\to \R$ given by the  representation 
$\frk A:=  (A^{(1)}, A^{(2)}, A^{(3)})$
and 
$\frk B:=  (b^{(1)}, b^{(2)}, b^{(3)})$
satisfies $f_\e \in \cA_{2d,1}$ and 
\begin{align}\label{result:one-bump-d-v2-e1}
 \{ f_\e >0\} &\subset (z_{1}, z_{2})\, ,  \\ \label{result:one-bump-d-v2-e2}
 \{ f_\e = 1\} & = [z_{1}+\e, z_{2}-\e]\, , \\ \label{result:one-bump-d-v2-e3}
%  \qquad \mbox{and} \qquad
 \{ f_\e < 0 \} &= \{ f_\e > 1\} = \emptyset\, .
\end{align}
\end{lemma}

\begin{proofof}{Lemma \ref{result:one-bump-d-v2}}
Let $H_1, H_2, H_3$ be the layers of $f_\e$. Then we have $H_3 = \id_\R$
and if $h_1^{(1)}, \dots,  h_d^{(1)}, h_1^{(2)},\dots, h_d^{(2)}$ denote the $2d$
component functions of $H_1$, that is 
\begin{displaymath}
	   H_1(x) = \Bigl(h_1^{(1)}(x), \dots,  h_d^{(1)}(x), h_1^{(2)}(x),\dots, h_d^{(2)}(x)\Bigr)^T\, , \qquad \qquad x\in \Rd,
\end{displaymath}
we thus find 
\begin{align} \nonumber
   f_\e(x) = H_3 \circ H_2 \circ H_1(x)
=  H_2 \circ H_1(x)
% &= \biggl| \,  \sum_{i=1}^d a_i h_i^{(1)}(x)  + \sum_{i=1}^d a_{d+i} h_i^{(2)}(x)    + b \, \biggr|_+ \\ \nonumber
&=  \biggl| \, - \sum_{i=1}^d  h_i^{(1)}(x)  - \sum_{i=1}^d   h_i^{(2)}(x)    +  1 \, \biggr|_+ \\ \label{result:one-bump-d-h0}
&=  \biggl| \,   \sum_{i=1}^d  \bigl(1 - h_i^{(1)}(x) - h_i^{(2)}(x) \bigr)      - d + 1 \, \biggr|_+ 
\end{align}
for all $x\in \R^d$.
Therefore, we first investigate the functions $1- h_i^{(1)} -  h_i^{(2)}$. To this end, let us fix an $i\in \{1,\dots,d\}$
and an $x=(x_1,\dots,x_d)\in \Rd$. Then we obviously have 
\begin{align*}
 h_i^{(1)}(x) 
%  &= \brelu{\langle  a_i^{(1)}, x\rangle +  b_i^{(1)} } 
 = \Brelu{- \frac {x_i} \e +    \frac {z_{1,i} + \e}\e  }
%  = \frac 1\e \cdot \brelu{-x_i + z_{1,i} + \e}
 =
 \begin{cases}
 \frac{-x_i + z_{1,i}+\e }\e  & \mbox{ if } x_i \leq z_{1,i}+\e\\
  0 & \mbox{ else, }
 \end{cases}
 \end{align*}
 and 
\begin{align*} 
  h_i^{(2)}(x) 
%  &= \brelu{\langle  a_i^{(2)}, x\rangle +  b_i^{(2)} } 
 = \Brelu{\frac {x_i} \e -   \frac {z_{2,i} - \e}\e  }
 = 
  \begin{cases}
 \frac{x_i - z_{2,i}+\e }\e  & \mbox{ if } x_i \geq z_{2,i}-\e\\
  0 & \mbox{ else. }
 \end{cases}
\end{align*}
Since $z_{1,i}+\e <  z_{2,i}-\e$, we 
consequently find 
\begin{displaymath}
 1- h_i^{(1)}(x) -  h_i^{(2)}(x)
  = 
  \begin{cases}
   \frac{x_i - z_{1,i}}\e  & \mbox{ if } x_i \leq z_{1,i}+\e\\
  1 & \mbox{ if }  x_i\in [z_{1,i}+\e,  z_{2,i}-\e]\\
 \frac{z_{2,i} - x_i }\e  & \mbox{ if } x_i \geq z_{2,i}-\e\, .
 \end{cases}
\end{displaymath}
In particular, we have 
\begin{align} \label{result:one-bump-d-h1}
 \bigl\{ 1- h_i^{(1)} -  h_i^{(2)} > 0\bigr\} 
 & = \bigl\{ (x_1,\dots,x_d)\in \Rd: x_i \in (z_{1,i}, z_{2,i})  \bigr\} \, ,\\  \label{result:one-bump-d-h2}
 \bigl\{ 1- h_i^{(1)} -  h_i^{(2)} = 1\bigr\}
 & = \bigl\{ (x_1,\dots,x_d)\in \Rd: x_i \in [z_{1,i}+\e, z_{2,i}-\e]  \bigr\} \, ,\\  \label{result:one-bump-d-h3}
 \bigl\{ 1- h_i^{(1)} -  h_i^{(2)} > 1\bigr\}
 & = \emptyset\, .
\end{align}
Combining  our initial equation \eqref{result:one-bump-d-h0} with \eqref{result:one-bump-d-h2} and  \eqref{result:one-bump-d-h3}  yields
\begin{align*}
 \{ f_\e = 1\}
 = \Biggl\{ \biggl| \,   \sum_{i=1}^d  \bigl(1 - h_i^{(1)} - h_i^{(2)} \bigr)      - d + 1 \, \biggr|_+     = 1 \Biggr\} 
 &= \biggl\{    \sum_{i=1}^d  \bigl(1 - h_i^{(1)} - h_i^{(2)} \bigr)            = d \biggr\} \\
 & = [z_{1}+\e, z_{2}-\e] \, ,
\end{align*}
i.e.~we have found \eqref{result:one-bump-d-v2-e2}
Net we will verify \eqref{result:one-bump-d-v2-e1}
to this end, we first note that 
 \eqref{result:one-bump-d-h0} 
% with  \eqref{result:one-bump-d-h1} 
gives 
\begin{align}\nonumber
 \{ f_\e >0\} 
 &=   \Biggl\{ \biggl| \,   \sum_{i=1}^d  \bigl(1 - h_i^{(1)} - h_i^{(2)} \bigr)      - d + 1 \, \biggr|_+   > 0 \Biggr\} \\  \label{result:one-bump-d-h4}
 &= \biggl\{    \sum_{i=1}^d  \bigl(1 - h_i^{(1)} - h_i^{(2)} \bigr)        >  d - 1 \biggr\} 
%  &\subset \bigcap_{i=1}^d   \bigl\{ 1- h_i^{(1)} -  h_i^{(2)} > 0\bigr\}\\
%  & = (z_{1}, z_{2})
 \, .
\end{align}
Our next intermediate goal is to show 
\begin{align}\label{result:one-bump-d-h5}
 \biggl\{    \sum_{i=1}^d  \bigl(1 - h_i^{(1)} - h_i^{(2)} \bigr)        >  d - 1 \biggr\} \subset \bigcap_{i=1}^d   \bigl\{ 1- h_i^{(1)} -  h_i^{(2)} > 0\bigr\} \, .
\end{align}
To this end, we assume the converse, i.e.~there is an
$x\in \Rd$ and an $i_0\in\{1,\dots,d\}$ with 
\begin{displaymath}
 \sum_{i=1}^d  \bigl(1 - h_i^{(1)}(x) - h_i^{(2)}(x) \bigr)      >  d - 1 
 \qquad \mbox{ and } \qquad
 1 - h_{i_0}^{(1)}(x) - h_{i_0}^{(2)}(x) \leq 0\, .
\end{displaymath}
Without loss of generality we may assume that $i_0 = d$. Then combining both inequalities we find 
\begin{displaymath}
 d-1 
 \,<\, 
 \sum_{i=1}^{d-1}   \bigl(1 - h_i^{(1)}(x) - h_i^{(2)}(x) \bigr)  +  \bigl(1 - h_d^{(1)}(x) - h_d^{(2)}(x) \bigr) 
 \,\leq \,\sum_{i=1}^{d-1}   \bigl(1 - h_i^{(1)}(x) - h_i^{(2)}(x) \bigr) \, ,
\end{displaymath}
and this shows that there is also  an $i\in \{1,\dots,d-1\}$ with 
$1 - h_i^{(1)}(x) - h_i^{(2)}(x) > 1$.
This contradicts  \eqref{result:one-bump-d-h3}, and hence we have shown 
\eqref{result:one-bump-d-h5}.
Now combining \eqref{result:one-bump-d-h4} with \eqref{result:one-bump-d-h5}
 and  \eqref{result:one-bump-d-h1} 
we obtain 
\begin{align*}
  \{ f_\e >0\}  \subset   \bigcap_{i=1}^d   \bigl\{ 1- h_i^{(1)} -  h_i^{(2)} > 0\bigr\} = (z_{1}, z_{2}) \, ,
\end{align*}
i.e.~we have found \eqref{result:one-bump-d-v2-e1}.
Finally, the equation $\{ f_\e > 1\} = \emptyset$ immediately follows from combining \eqref{result:one-bump-d-h0}  and \eqref{result:one-bump-d-h3},
and $\{ f_\e < 0\} = \emptyset$ is a direct consequence of 
\eqref{result:one-bump-d-h0}.
\end{proofof}

Our next goal is to describe how well the function 
$f_\e$ found in Lemma \ref{result:one-bump-d-v2}
approximates the indicator function $\eins_{[z_1,z_2]}$.
To this end, we first recall 
a well-known estimate
on $\inorm\cdot$-covering numbers of cuboids in the following lemma.
We include its proof for the sake of completeness.

\begin{lemma}\label{lemma:cover-of-cuboid}
 Let $s_1,\dots,s_d >0$, $s_{\mathrm{min}}:= \min\{s_1,\dots,s_d\}$,  $z\in \Rd$,  and
 \begin{align*}
  A:= \bigl\{ x\in \Rd:  z_i \leq  x_i \leq z_i + s_i\bigr\}\, .
 \end{align*}
Then for all $\e\in (0,s_{\mathrm{min}}]$ we have 
\begin{align*}
 \ca N(A, \inorm\cdot, \e) \leq \Bigl(\frac 32\Bigr)^d \biggl( \prod_{i=1}^d s_i \biggr)\cdot \e^{-d}\, .
\end{align*}
\end{lemma}

\begin{proofof}{Lemma \ref{lemma:cover-of-cuboid}}
 Let us fix an $i\in \{1,\dots,d\}$. Since $\e\leq s_i$, we then 
 need at most $\lceil \frac{s_i}{2\e}\rceil$ closed intervals of length
 $2\e$ to cover the interval $[z_i, z_i + s_i]$. From this it is easy to conclude 
 that 
 \begin{align*}
  \ca N(A, \inorm\cdot, \e) 
  \leq 
  \prod_{i=1}^d \Bigl\lceil \frac{s_i}{2\e}\Bigr\rceil
  \leq 
  \prod_{i=1}^d \Bigl( \frac{s_i}{2\e} + 1\Bigr)
    \leq 
  \prod_{i=1}^d \Bigl( \frac{s_i}{2\e} + \frac{s_i}{\e}\Bigr)
  =
  \Bigl(\frac 32\Bigr)^d \biggl( \prod_{i=1}^d s_i \biggr)\cdot \e^{-d}\, ,
 \end{align*}
 and hence we have shown the assertion.
\end{proofof}

Now, 
the next lemma describes the announced description of the approximation error.

\begin{lemma}
\label{lemma:aprox-error-bump}
%\fix{\eqref{lemma:aprox-error-bump-e2} should give us the required extra error bound. Example: 
%For the uniform distribution the right hand side of \eqref{lemma:aprox-error-bump-e2} behaves like $O(\e)$. I guess, that in general we will need to have a $\varphi$
%with $\e^{-d+1} \cdot \varphi(\e)\to 0$ for $\e\to 0$. Dann w\"ahlen wir $\e$ klein genug \dots}
 Let $z_1, z_2\in [-1,1]^d$, and $\e>0$ as in Lemma \ref{result:one-bump-d-v2}.
 Moreover, let $A\subset [-1,1]^d$ be a subset satisfying 
 $(z_1,z_2) \subset A\subset [z_1,z_2]$. Then the neural network $f_\e \in \archx{2d,1}$ constructed in Lemma \ref{result:one-bump-d-v2} satisfies 
 \begin{align}\label{lemma:aprox-error-bump-e1}
  \{ f_\e = \eins_{A} \} 
  =   [z_1 + \e , z_2 - \e ] \cup (X\setminus A) \;.
 \end{align}
Moreover, if $A$ is a cube of side length $s>0$, that is
$z_{2,i}-z_{1,i} = s$ for all $i=1,\dots,d$,
and
we have a distribution $P_X$ on $[-1,1]^d$ that satisfies 
Assumption \ref{ass:proba-2} for some $\varphi: \mbr_+ \to \mbr_+$, then 
we further  have 
\begin{align}\label{lemma:aprox-error-bump-e2}
 P_X \bigl( \{ f_\e \neq  \eins_{A} \}   \bigr)
 \leq 3d \cdot 
 \Bigl(\frac {3s}2\Bigr)^{d-1} 
 \cdot \e^{-d+1} \cdot \varphi(2\e)\, .
\end{align}
\end{lemma}

\begin{proofof}{Lemma \ref{lemma:aprox-error-bump}}
By \eqref{result:one-bump-d-v2-e2}
and
\eqref{result:one-bump-d-v2-e1}
we find the inclusions 
$\{ f_\e = 1\} =  [z_1 + \e , z_2 - \e ] \subset A$
and 
$\{ f_\e > 0 \} \subset (z_1  , z_2 ) \subset A$. Using 
$\{ f_\e < 0\} = \emptyset$, which is known by  \eqref{result:one-bump-d-v2-e3},
we then obtain
\begin{align*}
 \{ f_\e = \eins_{A} \} 
 &=  \bigl(A \cap  \{ f_\e = 1\}\bigr) \cup \bigl((X\setminus A) \cap  \{ f_\e = 0 \}\bigr) \\
  &=  \bigl(A \cap  \{ f_\e = 1\}\bigr) \cup \bigl((X\setminus A) \cap  (X\setminus\{ f_\e > 0 \})\bigr) \\
&=   [z_1 + \e , z_2 - \e ] \cup (X\setminus A) \;,
\end{align*}
i.e.~we have shown \eqref{lemma:aprox-error-bump-e1}. 
Now, to establish \eqref{lemma:aprox-error-bump-e2}, we first note that 
\eqref{lemma:aprox-error-bump-e1} together with  $A\subset [z_1,z_2]$ implies 
\begin{align*}
 \{ f_\e \neq \eins_{A} \}  
 = 
 \bigl(X\setminus  [z_1 + \e , z_2 - \e ]\bigr) \cap   A
 \subset 
 [z_1,z_2] \setminus (z_1 + \e , z_2 - \e )\, .
\end{align*}
To further bound $[z_1,z_2] \setminus (z_1 + \e , z_2 - \e )$ we define 
\begin{align*}
 S_i^- 
 &:= \bigl\{ x: x_i \in [z_{1,i}, z_{1,i} + \e] \mbox{ and } x_j \in [z_{1,j},z_{2,j}] \mbox{ for all } j\neq i  \bigr\}\, , \\
  S_i^+ 
 &:= \bigl\{ x: x_i \in [z_{2,i} - \e, z_{2,i}] \mbox{ and } x_j \in [z_{1,j},z_{2,j}] \mbox{ for all } j\neq i  \bigr\}
\end{align*}
Then we have $[z_1,z_2] \setminus (z_1 + \e , z_2 - \e ) \subset 
S_1^-\cup \dots\cup S_d^- \cup S_1^+\cup \dots\cup S_d^+$, and hence we obtain
\begin{align}\label{lemma:aprox-error-bump-h1}
P_X \bigl( \{ f_\e \neq  \eins_{A} \}   \bigr)
 \leq
 \sum_{i=1}^d P_X(S_i^-) +  \sum_{i=1}^d P_X(S_i^+)\, . 
\end{align}
Now observe that since $A$ is a cube with side length $s$, the sets 
$S_i^-$ and $S_i^+$ are cuboids with 
side lengths $s_1,\dots,s_d$, where 
$s_i = \e$ and $s_j = s$ for all $j\neq i$.
Applying Lemma \ref{lemma:cover-of-cuboid}   then shows 
\begin{align*}
 \ca N(S_i^{\pm 1}, \inorm\cdot, \e) \leq \Bigl(\frac 32\Bigr)^d 
 s^{d-1}\cdot \e^{-d+1}\, ,
\end{align*}
and combining with Assumption \ref{ass:proba-2} we obtain 
\begin{align*}
 P_X(S_i^{\pm 1}) \leq \Bigl(\frac 32\Bigr)^d 
 s^{d-1}\cdot \e^{-d+1} \cdot \varphi(2\e)\, .
\end{align*}
Inserting this estimate into \eqref{lemma:aprox-error-bump-h1}
yields \eqref{lemma:aprox-error-bump-e2}.
\end{proofof}

As a second step in our construction presented in Subsection \ref{subsec:eps-inflate-histo-dnn} we combine
Lemmas \ref{result:dnn-algebra-1} and \ref{result:dnn-algebra-2} 
 with Lemma \ref{result:one-bump-d-v2} to approximate step-functions 
on cubic partitions by ReLU-DNNs with two hidden  layers.

\begin{proposition}\label{result:DNN-aprox-HR}
%\fix{\noteis{Achtung, in Proposition \ref{result:DNN-aprox-HR} hat sich noch die Aussage etwas ge\"andert: aus $X\setminus (z_i^{(1)},z_i^{(2)})$
%ist $X\setminus A_i$ geworden! Dies hat nat\"urlich nur Auswirkungen auf einer 
%Lebesgue-, also auch $P_X$-Nullmenge. Au\ss erdem ist noch $\inorm{f_\e}$-Schranke hinzugekommen, da ich glaube, 
%dass wir sie noch brauchen werden und der blaue text weiter unten.}}
Let $A_1,\dots,A_k$ be mutually disjoint subsets of $X:= [-1,1]^d$ such that for each $i\in \{1,\dots,k\}$
there exist $z_i^-, z_i^+\in X$ with $z_i^-< z_i^+$ and  $(z_i^-, z_i^+) \subset A_i \subset [z_i^-, z_i^+]$.
Moreover, let  $z_{i,j}^\pm$ be the $j$-th coordinate of $z_i^\pm$.
Then for all $g:X\to \R$ of the form
\begin{equation}
\label{eq:repr-f}
 g = \sum_{i=1}^k \a_i \eins_{A_i} 
\end{equation}
with $\a_i \in\R$,  all $\e>0$ satisfying
\begin{displaymath}
 \e < \min \Bigl\{ \frac{z_{i,j}^+ - z_{i,j}^-} 2: i=1\dots,k \mbox{ and } j=1,\dots,d\Bigr\}
\end{displaymath}
and all
$m_1 \geq  2d  k$ and $m_2\geq k$, 
there exists a neural network $f_\e\in \archx{m_1,m_2}$ such that 
\begin{displaymath}
 \{ f_\e = g \} = \bigcup_{i=1}^k \;  [z_{i}^- + \e , z_{i}^+ - \e ] \cup (X\setminus A_i )  \;.
\end{displaymath}
and $\inorm {f_\e} = \max\{|\a_1|,\dots,|\a_k|\}$. 
In addition, if $A_1,\dots,A_k$ are cubes of side length $s>0$, 
i.e.~$z_i^+-z_i^- = (s,\dots, s)\in \R^d$ for all $i=1,\dots,k$, and 
$P_X$ is a distribution on $[-1,1]^d$ that satisfies 
Assumption \ref{ass:proba-2} for some $\varphi: \mbr_+ \to \mbr_+$, then 
we further  have 
\begin{align*} 
 P_X \bigl( \{ f_\e \neq g \}   \bigr)
 \leq 3d k\cdot 
 \Bigl(\frac {3s}2\Bigr)^{d-1} 
 \cdot \e^{-d+1} \cdot \varphi(2\e)\, .
\end{align*}
\end{proposition}

\begin{proofof}{Proposition \ref{result:DNN-aprox-HR}}
Since $\archx{2dk, k}\subset \archx{m_1,m_2}$ it suffices 
to find an $f_\e$
with the desired properties in $\archx{2dk, k}$
%For $i=1,\dots,k$ and $\e > 0$, we introduce the sets 
%\[  I_{i, \e} := \{  (x_1,\dots,x_d) \in \mbr^d : x_j \in [z_{i,j}^- + \e , z_{i,j}^+ - \e ]  \} \;. \]
By assumption and Lemma \ref{result:one-bump-d-v2}, we find, for all $\e >0$ and $i=1,\dots, k$,
a neural network   $f^{(\e)}_i   \in \cA_{2d,1}$, and Lemma 
\ref{lemma:aprox-error-bump} shows that 
\begin{align*}
 \{ f^{(\e)}_i = \eins_{A_i} \} 
&=   [z_{i}^- + \e , z_{i}^+ - \e ] \cup (X\setminus A_i) \;.
\end{align*}
Moreover, for any $\alpha_i \in \mbr$, Lemma \ref{result:dnn-algebra-1} ensures $\alpha_i f^{(\e)}_i \in \cA_{2d,1}$ with 
\[  \{\alpha_i   f^{(\e)}_i = \alpha_i  \eins_{A_i} \} = [z_{i}^- + \e , z_{i}^+ - \e ] \cup (X\setminus  A_i)   \;.  \]
Now, applying Lemma \ref{result:dnn-algebra-2}  shows that  
\[ f_\e := \sum_{i=1}^k \alpha_i   f^{(\e)}_i  \] 
belongs to $\cA_{2kd,k}$, and
since we have 
\begin{align}\nonumber
 \{ f^{(\e)}_i \neq 0  \} \cap \{ f^{(\e)}_l \neq 0  \}
 = 
 \{ f^{(\e)}_i > 0  \} \cap \{ f^{(\e)}_l > 0  \}
 \subset
 (z_{i}^-  , z_{i}^+ )  \cap (z_{l}^-  , z_{l}^+ )
 &\subset 
 A_i \cap A_l \\ \label{result:DNN-aprox-HR-h1}
 &= \emptyset
\end{align}
for all $i\neq l$, our previous considerations give us
\[
\{ f_\e = g \} = \bigcup_{i=1}^k \;   [z_{i}^- + \e , z_{i}^+ - \e ]\cup (X\setminus A_i) \;.
\]
Finally, the identity $\inorm {f_\e} = \max\{|\a_1|,\dots,|\a_k|\}$
follows from \eqref{result:DNN-aprox-HR-h1} and 
$\inorm{f_i^{(\e)}} = |\a_i|$ for all $i=1,\dots,k$
and the bound on $ P_X \bigl( \{ f_\e \neq  \eins_{A} \}   \bigr)$ 
is a direct consequence of 
\eqref{lemma:aprox-error-bump-e2}.
\end{proofof}

%%%%%%%%%%%%%%%%%%%%%%%%%%%%%%%%%%%%%%%%%%%%%%%%%%%%%%%%%%%%%%%%%%%%%%%%%%%%%%%%%%%%%%%%%%%%%%%%%%%%%%%%%%%%%%%%%%%%%%%%%%%%%%%%%%%%%%%%%%%%%%%%%%
%%%%%%%%%%%%%%%%%%%%%%%%%%%%%%%%%%%%%%%%%%%%%%%%%%%%%%%%%%%%%%%%%%%%%%%%%%%%%%%%%%%%%%%%%%%%%%%%%%%%%%%%%%%%%%%%%%%%%%%%%%%%%%%%%%%%%%%%%%%%%%%%%%

\subsection{Proof of Main Theorem \ref{results:main-dnn}}
\label{sec:proof-thm2}

%\fix{{\notice check if reasonable }}
Throughout this section we assume that the general  assumptions of Theorem \ref{results:main-dnn} are satisfied. 
In particular,  $D \in (X \times Y)^n$ is an i.i.d.~sample of size $n \geq 1$ and 
$D_X:=\{x_1^*, ..., x_{m_n}^*\}\in \potm$ is  the set of input observations.
Moreover,  $(s_n)_{n \in \mbn}$ is a sequence with 
$s_n \rightarrow 0$, $s_n^d > 2^d/n$ and $\frac{\ln (n s_n^d)}{n s_n^d}\to 0$ as $n \to \infty$. 
In addition, we let $(\rho_n)_{n \in \mbn}$ be a non-negative sequence with $\rho^d_n \leq 2^d/n$ and 
$\rho_n^{-d} \varphi( \rho_n ) \to 0$  for $n\to \infty$. Finally, let  $(\e_n)_{n \in \mbn}$ and  $(\delta_n)_{n \in \mbn}$ be positive 
sequences with $\e_n = \delta_n = \rho_n/2$.

We firstly show our claim for the good interpolating DNN from Example \ref{good-DNN}, having representation 
\[  \gdnn =  h^{+,(\e_n)}_{D, \cA_D}    + \sum_{i=1}^m b^+_i \eins^{(\delta_n)}_{x_i^* + t B_\infty  } \]
with $t=\min\{r, \rho_n\} $ and associated $\cH_{\cA}^{(\epsilon)}$-part 
\[  h^{+,(\e_n)}_{D, \cA_D} := \sum_{j\in J} c_j^+ \eins^{(\e_n)}_{A_j}  \;. \] 
We split the excess risk into three different terms

\begin{align}
\label{eq:ex-11}
\cR_{L,P}(\gdnn)-\cR^*_{L,P} &= \cR_{L,P}(\gdnn)-  \cR_{L,P}(h^{+, (\e )}_{D, \cA_D} )  \no \\
   & +   \cR_{L,P}(h^{+, (\e )}_{D, \cA_D} )   -  \cR_{L,P}(h^{+}_{D, \cA_D} )    +  \cR_{L,P}(h^{+}_{D, \cA_D} ) -  \cR^*_{L,P} \;. 
\end{align}
Convergence of the the first term follows from Lemma \ref{result:classifier-diff} and by exploiting Assumption \ref{ass:proba-2}. We obtain 

\begin{align}
\label{eq:t1}
 \cR_{L,P}(\gdnn)-  \cR_{L,P}(h^{+, (\e_n )}_{D, \cA_D} ) &\leq  4\;P_X\bigl( \{ \gdnn \neq h^{+, (\e_n )}_{D, \cA_D} \}\bigr)   \no  \\
 &\leq 4\;P_X\bigl( D_X^{+t} \bigr) \no  \\
&\leq  4\;P_X\bigl( D_X^{+\rho_n } \bigr) \no  \\
&\leq 4 n\varphi( \rho_n ) \no \\
&\leq 4\cdot 2^d \rho_n^{-d} \varphi( \rho_n )  \;. 
\end{align}
Hence, by our assumption on $\rho_n$ may we conclude 
\begin{align*}
%\label{eq:ex-71}
 \cR_{L,P}(\gdnn)-  \cR_{L,P}(h^{+, (\e_n )}_{D, \cA_D} )  \to 0 \;,
\end{align*} 
in probability for $|D|\to \infty$.

For bounding the second term in \eqref{eq:ex-11} we remind that  $|J|\leq \left(\frac{2}{s_n}\right)^d $. 
Lemma \ref{result:classifier-diff} and Proposition \ref{result:DNN-aprox-HR} yield\footnote{This is justified since $\e_n = \rho_n/2 \leq n^{-1/d} < s_n/2 $. } 
\begin{align}
\label{eq:t2}
  \cR_{L,P}(h^{+, (\e_n )}_{D, \cA_D} )   -  \cR_{L,P}(h^{+}_{D, \cA_D} )  &\leq  4\;P_X\bigl( \{ h^{+,(\e_n)}_{D, \cA_D} \neq h^+_{D, \cA_D} \}\bigr)   \no   \\
  &\leq 12\cdot d |J| \cdot  \Bigl(\frac {3s_n}2\Bigr)^{d-1}  \cdot \e_n^{-d+1} \cdot \varphi(2\e_n) \no \\
  &\leq 12 \cdot  d \left(\frac{2}{s_n}\right)^d \cdot  \Bigl(\frac {3s_n}2\Bigr)^{d-1}  \cdot \e_n^{-d+1} \cdot \varphi(2\e_n) \no \\
%&=   24\cdot  d \cdot  3^{d-1} s_n^{-1}   \cdot \e_n^{-d+1} \cdot \varphi(2\e_n) \no  \\
&\leq 4\cdot  d \cdot  6^{d}    \cdot \rho_n^{-d} \cdot \varphi(\rho_n)   \;.
\end{align}
Hence, our assumption on $\rho_n$ ensures 
\begin{align*}
%\label{eq:ex-41}
\cR_{L,P}(h^{+, (\e_n )}_{D, \cA_D} )   -  \cR_{L,P}(h^{+}_{D, \cA_D} )  \to 0 \;,  
\end{align*}
in probability for $|D|\to \infty$. 
Finally, convergence of the last term in \eqref{eq:ex-11} is easily derived with the help of Proposition \ref{prop:cons-reg} and we conclude that 
\[  ||\gdnn  - \fpb||^2_{L_2(P_X)} = \cR_{L,P}(\gdnn)-\cR^*_{L,P} \to 0 \;,   \] 
in probability for $|D|\to \infty$.

We now turn to considering the bad interpolating DNN from Example \ref{good-DNN}. 
Since we have $\bdnn (x) \in [-1,1]$ and $f_{L,P}^\dagger (x) \in [-1,1]$ for all $x \in X$, 
Assumption \ref{ass:proba-2} gives  
\begin{align}
\label{eq:ex-71}
\int_{D_X^{+t}} ( \bdnn (x) - f_{L,P}^\dagger (x) )^2\; dP_X &\leq 4P_X(D_X^{+\rho_n}) \leq   4 n \varphi(\rho_n) \leq   4 \rho_n^{-d} \varphi(\rho_n) \;.
\end{align}
Moreover, for all $x \in X\setminus D_X^{+t}$ we have $\bdnn (x) = -h_{D, \cA_D}^{+, (\e_n)}$ and   $f_{L,P}^\dagger (x)= -f^*_{L,P}(x)$. Hence 
\begin{align*}
%\label{eq:ex-81}
\int_{X\setminus  D_X^{+t}} ( \bdnn (x) - f_{L,P}^\dagger (x) )^2\; dP_X &\leq \cR_{L,P}(h_{D, \cA_D}^{+, (\e_n)}) - \cR^*_{L,P} \;.
\end{align*}
Combining both considerations with the first part of the proof shows then in probability for $|D|\to \infty$ 
\[ ||\bdnn  - \fpd||_{L_2(P_X)} \to 0 \;. \]
Finally, since $|J| \leq (\frac{2}{s_n})^d \leq n$, Proposition \ref{result:DNN-aprox-HR} shows that $\gbdnn \in \cA_{4dn, 2n}$.

%%%%%%%%%%%%%%%%%%%%%%%%%%%%%%%%%%%%%%%%%%%%%%%%%%%%%%%%%%%%%%%%%%%%%%%%%%%%%%%%%%%%%%%%%%%%%%%%%%%%%%
%%%%%%%%%%%%%%%%%%%%%%%%%%%%%%%%%%%%%%%%%%%%%%%%%%%%%%%%%%%%%%%%%%%%%%%%%%%%%%%%%%%%%%%%%%%%%%%%%%%%%%

\subsection{Proof of Main Theorem \ref{results:main-dnn-2}}
\label{sec:proof-thm3}

Let all assumptions of Theorem \ref{results:main-dnn-2} be satisfied. Moreover, we let  $(\e_n)_{n \in \mbn}$ and  $(\delta_n)_{n \in \mbn}$ be positive 
sequences with $\e_n = \delta_n = \rho_n/2$. We prove the result for the good interpolating DNN by reconsidering \eqref{eq:ex-11}. Indeed, by our 
assumption $\rho_n^{-d} \varphi(\rho_n) \leq \ln(n) n^{-2/3}$ and thus \eqref{eq:t1} leads to 
\[  \cR_{L,P}(\gdnn)-  \cR_{L,P}(h^{+, (\e_n )}_{D, \cA_D} ) \leq 4 \rho_n^{-d} \varphi( \rho_n ) 
 \leq 4 \ln(n) n^{-2/3} \leq 4 \ln(n) n^{-2\alpha \gamma} \;. \]
Moreover, \eqref{eq:t2} gives 
\[  \cR_{L,P}(h^{+, (\e_n )}_{D, \cA_D} )   -  \cR_{L,P}(h^{+}_{D, \cA_D} ) \leq  4\cdot  d \cdot  6^{d}    \cdot \rho_n^{-d} \cdot \varphi(\rho_n)  
\leq 16\cdot  d \cdot  6^{d}  \ln(n) n^{-2\alpha \gamma} \;.  \]
Finally, \eqref{eq:riskplus} shows with probability $P^n$ at least $1- 2^dn^{1+d} e^{-n^{d \gamma}}$ 
\begin{equation}
\label{eq:risk-hplus}
   \cR_{L,P}(h^{+}_{D, \cA_D} ) -  \cR^*_{L,P}  \leq c_{d,\alpha }\ln(n  )  \,  n^{-2\alpha \gamma }  \;,   
\end{equation}   
where $c_{d,\alpha }>0$ is a constant only depending on $d$, $\alpha$, and $|f^*_{L,P}|_\alpha$. Collecting the above considerations shows 
the first part of the theorem.

Now, coming to the bad interpolating DNN, we derive from \eqref{eq:ex-71} 
\[ \int_{D_X^{+t}} ( \bdnn (x) - f_{L,P}^\dagger (x) )^2\; dP_X \leq 4P_X(D_X^{+\rho_n}) \leq 4 \ln(n) n^{-2\alpha \gamma} \;.  \]
Moreover, combining the results from \eqref{eq:t2} and \eqref{eq:risk-hplus}  gives with probability $P^n$ at least $1- 2^dn^{1+d} e^{-n^{d \gamma}}$  
\begin{align*}
\cR_{L,P}(h_{D, \cA_D}^{+, (\e_n)}) - \cR^*_{L,P}  &\leq 4\cdot  d \cdot  6^{d}    \cdot  \ln(n) n^{-2/3} +  c_{d,\alpha }\ln(n  )  \,  n^{-2\alpha \gamma }  \\
&\leq c'_{d,\alpha } \ln(n  )  \,  n^{-2\alpha \gamma }    \;,
\end{align*}
where $ c'_{d,\alpha }= 4\cdot  d \cdot  6^{d} +  c_{d,\alpha } $. Thus, with probability $P^n$ at least $1- 2^dn^{1+d} e^{-n^{d \gamma}}$  
\[ ||\bdnn  - \fpd||_{L_2(P_X)}^2 \leq   c''_{d,\alpha } \ln(n  )  \,  n^{-2\alpha \gamma }  \;, \]
where $c''_{d,\alpha } = 4 + c'_{d,\alpha }$.

Finally, we have $|J| \leq (\frac{2}{s_n})^d = 2^d n^{\frac{d}{2\alpha +d }}\leq n$, provided $n \geq n_{d, \alpha}$, for some $ n_{d, \alpha} \in \mbn$, depending 
on $d$ and $\alpha$. Proposition \ref{result:DNN-aprox-HR} shows then that $\gbdnn \in \cA_{4dn, 2n}$.

\section{Uniform bounds for histograms based on data-dependent partitions}
\label{app:histo-random}
%builds the foundation for the proofs of the learning properties for the inflated histogram rules and is of independent interest.  

The main purpose of this section is to firstly present a general variance improved oracle inequality  
for bounding the excess risk with respect to a broad class of loss functions. 
We apply this result to the special case of the least squares loss and to 
histogram rules that choose their cubic partitions in a certain, data-dependent way. In particular, we give an optimized uniform bound that crucially 
relies on an explicit capacity bound, expressed in terms of the covering number. 
This is a necessary step to provide in 
Section \ref{sec:uniform-histos} the learning properties of histograms based on data-dependent cubic partitions.

%%%%%%%%%%%%%%%%%%%%%%%%%%%%%%%%%%%%%%%%%%%%%%%%%%%%%%%%%%%%%%%%%%%%%%%%%%%%%%%%%%%%%%%%%%%%%%%%%%%%%%%%%%%
%%%%%%%%%%%%%%%%%%%%%%%%%%%%%%%%%%%%%%%%%%%%%%%%%%%%%%%%%%%%%%%%%%%%%%%%%%%%%%%%%%%%%%%%%%%%%%%%%%%%%%%%%%%

\subsection{A Generic Oracle Inequality for Empirical Risk Minimization}
\label{sec:genericERM}

If not stated otherwise, we assume throughout this subsection that $X$ is an arbitrary non-empty set
that is equipped with some $\sigma$-algebra. We write $\cL_\infty$ for the corresponding set 
of all bounded, measurable functions $f:X\to \R$. Moreover, $Y\subset \R$ is assumed to be measurable.
Following \cite[Definition 2.18]{StCh08} we say that a measurable
loss  $L:X\times Y\times \R\to [0,\infty)$ is   locally Lipschitz continuous 
if for all $a\geq 0$ there exists a constant $c_{a}\geq 0$ such that %for all $t,t'\in [-a,a]$ we have
\begin{equation}\label{loss:lipschitz-loss-def}
\sup_{\substack{x\in X\\ y\in Y}}  \bigl| L(x,y, t) - L(x,y,t')\bigr| \, \leq \, c_{a} \, |t-t'|\, , \qquad\qquad t,t'\in [-a,a]\, .
\end{equation}
Moreover, for $a\geq 0$, the smallest such constant $c_{a}$ is denoted by $|L|_{a,1}$.

In addition, we need to recall the notion  of covering numbers, which is recalled in the following definition.

\begin{definition}
Let $(T,d)$ be a metric space and $\e>0$.
We call $S\subset T$ an $\e$-net of $T$ if for all $t\in T$ there exists 
an $s\in S$ with $d(s,t)\leq \e$.
Moreover, 
the $\e$-covering number
of $T$   is defined by
$$
\ca N(T,d,\e) 
:=
\inf\biggl\{
n \geq 1: \exists\, s_{1},\dots,s_{n}\in T \mbox{ such that } T\subset \bigcup_{i=1}^{n} B_{d}(s_{i},\e) 
\biggr\} \, ,
$$ 
where $\inf\emptyset := \infty$ and
$B_{d}(s,\e):=\{t\in T: d(t,s)\leq \e\}$ denotes the closed ball with center $s\in T$ and radius $\e$.

Moreover, if $(T,d)$ is a subspace of a normed space $(E,\snorm {\cdot})$ and 
the metric is given by $d(x,x') = \snorm{x-x'}$, $x,x'\in T$, 
we write $\ca N(T,\snorm {\cdot},\e):= \ca N(T,d,\e)$. 
\end{definition}

Finally, we need to fix some notation related to empirical risk minimization. To this end, we fix a loss function
$L:X\times Y\times \R\to [0,\infty)$ and an 
$\ca F\subset \sL \infty X$. Given a distribution $P$ on $X\times Y$, we denote the smallest 
possible risk attained by functions in $\ca F$ by $\RPxB L {\ca F}$, that is 
\begin{align*}
 \RPxB L {\ca F} := \inf_{f\in \ca F} \RP L f\, .
\end{align*}
Finally, following \cite[Definition 6.2]{StCh08}, we say that an ERM method $D\mapsto f_D$ with respect to $L$ and 
$\ca F$ is measurable, if
for all $n\geq 1$ the map 
\begin{align*}
 (X\times Y)^n \times X & \to \R \\
 (D,x)& \mapsto f_D(x)
\end{align*}
is measurable with respect to the universal completion of the product $\sigma$-algebra of the product space
$ (X\times Y)^n \times X$.
 Recall from \cite[Lemma 6.17]{StCh08}
that for closed, separable $\ca F \subset \sL \infty X$ for which there exists an ERM, there also 
exists a  measurable ERM. Moreover, in this case the map 
\begin{align*}
 (X\times Y)^n & \to [0,\infty] \\
 (D,x)& \mapsto \RP L {f_D}
\end{align*}
is also measurable with respect to the universal completion of the product $\sigma$-algebra of 
$ (X\times Y)^n$, see  \cite[Lemma 6.3]{StCh08}. In the following, we thus assume that 
$ (X\times Y)^n$ is equipped with this universal completion.

With the help of these notion we can now state the generic oracle inequality for empirical risk minimizers.

\begin{theorem}
\label{theo:generic-oracle}
 Let $L:X\times Y\times \R\to [0,\infty)$ be a locally Lipschitz continuous loss, 
%  with $|L|_{1,M}\leq 1$, 
 $\ca F\subset \sL \infty X$ be  a closed, separable set  satisfying $\inorm f\leq M$ for a suitable constant $M>0$ 
and all $f\in \ca F$, and $P$ be a distribution on $X\times Y$ that has a Bayes decision function $f_{L,P}^{*}$ with $\RP L \fpb < \infty$.
Assume that there exist constants $B>0$, 
$\vt\in [0,1]$, and $V\geq B^{2-\vt}$ such that for all measurable $f: X \to [-M,M]$ we have 
\begin{eqnarray}\label{sup-bound}
\inorm {L\circ f- L\circ \fpb} & \leq & B\, ,\\ \label{var-bound}
\Ex_{P} \bigl( L\circ f - L\circ \fpb    \bigr)^{2} & \leq & V \cdot  \bigl(\Ex_{P} (L\circ f - L\circ \fpb)    \bigr)^{\vt}\, .
\end{eqnarray}
Then, for all measurable empirical risk minimization algorithms $D\mapsto f_D$, 
 all $n\geq 1$, $\tau>0$, and all $\e>0$ we have 
\begin{eqnarray*}
\RP L {f_D} - \RPB L 
&\leq & 4\bigl( \RPxB L {\ca F}-\RPB L\bigr) + 5\, |L|_{M,1} \cdot \e \\
&&\quad +
2 \biggl(\frac{16 V\bigl(\tau +1+ \ln  \ca N(\ca F,\inorm\cdot,\e) \bigr)  }n\biggr)^{\frac 1{2-\vt}} 
\end{eqnarray*}
with probability $P^n$ not less than $1- e^{-\tau}$.
\end{theorem}

\begin{proof}
We first note that \eqref{sup-bound} ensures $\RP L {f_D} - \RPB L\leq B$
and since we have additionally assumed 
 $V\geq B^{2-\vt}$, we see that  it suffices to consider sample sizes $n\geq 16\tau$.

Given an  $f\in \ca F$, we define $h_f:=L\circ f-L\circ \fpb$. 
Let us now 
 fix an $\fO\in \ca F$ and a data set  $D\in (X\times Y)^n$. Since $f_D$ is an empirical risk minimizer, we have
$\RD L {f_D}\leq \RD L \fO$, and hence we find $\Ex_D h_{f_D} \leq \Ex_D h_\fO$. As a consequence, we obtain
\begin{eqnarray}\nonumber
 \RP L {f_D} - \RP L \fO
& = &
\Ex_P h_{f_D}  - \Ex_P h_\fO \\ \label{conc-adv:erm-h1}
&\leq & 
\Ex_P h_{f_D} - \Ex_D h_{f_D} + \Ex_D h_\fO - \Ex_P h_\fO\, .
\end{eqnarray}

To bound the first difference in \eqref{conc-adv:erm-h1}
we first observe that 
for $f,f'\in \ca F$, 
$x\in X$, and $y\in Y$ the local Lipschitz continuity of $L$ %with $|L|_{M,1}\leq 1$
gives
\begin{align*}
 \bigl| h_f(x,y) - h_{f'}(x,y)  \bigr|
 = \bigl| L(x,y,f(x)) -  L(x,y,f'(x)) \bigr| \leq |L|_{M,1} \cdot \bigl| f(x) - f'(x)   \bigr|\, , 
\end{align*}
and thus we have $\inorm{h_f-h_{f'}}\leq |L|_{M,1} \cdot \inorm{f-f'}$ for all $f,f'\in \ca F$.
Now, let $\ca C\subset \ca F$ be a minimal $\e$-net   of $\ca F$ with respect to  $\inorm\cdot$.
For a data set $D\in (X\times Y)^n$ there then exists an $f\in \ca C$ such that 
$\inorm{f-f_D} \leq \e$, and hence $\inorm{h_{f_D} - h_f}\leq |L|_{M,1} \cdot \e$. This yields
\begin{align} \nonumber
\max\Bigl\{ \bigl|\Ex_P h_{f_D} - \Ex_P h_f\bigr| \, ,\bigl|\Ex_D h_{f_D} - \Ex_D h_f\bigr| \Bigr\}
 &\leq \max\Bigl\{  \Ex_P |h_{f_D} - h_f| \, , \Ex_D |h_{f_D} - h_f| \Bigr\} \\ \nonumber
 &\leq \inorm{h_{f_D} - h_f} \\ \label{eps-approx}
 &\leq |L|_{M,1} \cdot\e\, .
\end{align}
For $f\in \ca C$ and $r>0$ we now define the function 
$$
g_{f,r}:= \frac {\Ex_P h_f- h_{f}}{\Ex_P h_f +r}\, .
$$
It is easy to see that both $\Ex_P g_{f,r}=0$ and $\inorm {g_{f,r}}\leq 2Br^{-1}$ hold. In addition, in the case $\vt>0$ and $b:= \Ex_P h_f\neq 0$, setting 
$q:= \frac 2 {2-\vt}$, $q':= \frac 2\vt$, and $a:= r$  in the second inequality of 
\cite[Lemma 7.1]{StCh08} shows
$$%\begin{equation}\label{concentration:model-selct-hxx}
\Ex_P  g_{f,r}^2
\leq 
\frac{\Ex_P  h_f^2}{(\Ex_P h_f+r)^2} 
\leq 
\frac{(2-\vt)^{2-\vt}\vt^\vt\, \Ex_P  h_f^2}{ 4 r^{2-\vt}(\Ex_P h_f)^\vt} 
\leq
 V{r^{\vt -2}} \, .
$$%\end{equation} 
Furthermore, in the case $\vt>0$ and $\Ex_P h_f=0$, the variance bound (\ref{var-bound}) gives
$\Ex_P  h_f^2=0$, and hence 
we have  $\Ex_P  g_{f,r}^2 \leq V{r^{\vt-2}}$. %(\ref{concentration:model-selct-hxx}).
Finally, in the case $\vt = 0$, we have  
$\Ex_P  g_{f,r}^2 \leq \Ex_P  h_f^2 \, r^{-2} \leq V{r^{\vt-2}}$. In summary, we 
we have thus found
\begin{align*}
 \Ex_P  g_{f,r}^2 \leq V{r^{\vt-2}}
\end{align*}
in all cases.
By applying  Bernstein's inequality in the form of \cite[Theorem 6.12]{StCh08}
in combination with a union bound we thus find
\begin{equation}\label{concentration:model-selct-h1}
P^n
\biggl(
 D\in (X\times Y)^n  : \sup_{f\in \ca C} \Ex_D g_{f,r} < \sqrt{\frac{2V\tau }{nr^{2-\vt}}}+ \frac {4B\tau}{3n r}
\biggr)
\geq
1-|\ca C|\,e^{-\tau} \quad
\end{equation}
for all $r>0$. Let us now pick a data set $D\in (X\times Y)^n$  that satisfies the above inequality, that is 
$$
\sup_{f\in \ca C}\Ex_D g_{f,r} <\sqrt{\frac{2V\tau }{nr^{2-\vt}}}+ \frac {4B\tau}{3n r}\, .
$$
For an $f\in \ca C$ with $\inorm{f-f_D} \leq \e$, Inequality \eqref{eps-approx} together with the definition of 
$g_{f,r}$ then gives
\begin{align} \nonumber
 \Ex_P h_{f_D} - \Ex_D h_{f_D} 
 &\leq 
 \Ex_P h_{f} - \Ex_D h_{f} + 2\,|L|_{M,1} \cdot\e \\   \nonumber
 & <
 \Ex_P h_{f}\biggl(\sqrt{\frac{2V\tau }{nr^{2-\vt}}}+ \frac {4B\tau}{3n r}\biggr) +     \sqrt{\frac{2V\tau r^\vt}{n}}+ \frac {4B\tau}{3n} + 2\,|L|_{M,1} \cdot\e \\ \label{oracle-h1}
 & \leq 
   \bigl( \Ex_P h_{f_D}+\e\bigr)\biggl(\sqrt{\frac{2V\tau }{nr^{2-\vt}}}+ \frac {4B\tau}{3n r}\biggr) +     \sqrt{\frac{2V\tau r^\vt}{n}}+ \frac {4B\tau}{3n} + 2\, |L|_{M,1} \cdot\e\, .
\end{align}

Our next goal is 
to estimate the second difference 
\eqref{conc-adv:erm-h1}, that is  $\Ex_D h_\fO - \Ex_P h_\fO$.
Let us first consider 
 case $\vt>0$.
Here, we have both $\inorm{ h_\fO - \Ex_P h_\fO} \leq 2B$ and 
$$
\Ex_P ( h_\fO - \Ex_P h_\fO)^2\leq \Ex_P h_\fO^2\leq V (\Ex_P h_\fO)^\vt\, .
$$
Furthermore, setting $q:= \frac 2 {2-\vt}$, $q':= \frac 2\vt$, $a:=\bigl(\frac{2^{1-\vt}\vt^\vt  V\tau}{n} \bigr)^{1/2}$, and $b:= \bigl(\frac{2\Ex_P h_\fO}{\vt}\bigr)^{\vt/2}$ in 
\cite[Lemma 7.1]{StCh08} yields
$$
\sqrt {\frac{2\tau V (\Ex_P h_\fO)^\vt}n} 
\leq 
\biggl(1-\frac \vt 2\biggr)\biggl(\frac{2^{1-\vt}\vt^\vt  V\tau}n\biggr)^{\frac 1{2-\vt}} +  \Ex_P h_\fO 
\leq  
\biggl(\frac{2 V\tau}n\biggr)^{\frac 1{2-\vt}} +  \Ex_P h_\fO ,
$$
By another application of Bernstein's inequality we consequently find that 
\begin{equation}\label{concentration:model-selection-h1}
 \Ex_D h_\fO - \Ex_P h_\fO<   \Ex_P h_\fO + \biggl(\frac{2 V\tau}n\biggr)^{\frac 1{2-\vt}}  +   \frac {4B\tau}{3n}     
\end{equation}
holds with probability $P^n$ not less than $1-e^{-\tau}$.
Finally, in the case  $\vt=0$, Hoeffding's inequality in combination with 
$\inorm {h_{f_0}}\leq B\leq \sqrt V$ also yields \eqref{concentration:model-selection-h1}.

To finish the proof, we now combine  (\ref{conc-adv:erm-h1}), \eqref{concentration:model-selct-h1},
\eqref{oracle-h1},  and 
(\ref{concentration:model-selection-h1}).  As a result we see that  
\begin{align*}
\Ex_P h_{f_D}   &< 2 \Ex_P h_\fO +  \bigl(\Ex_P h_{f_D} + \e\bigr)\biggl(\sqrt{\frac{2V\tau }{nr^{2-\vt}}}+ \frac {4B\tau}{3n r}\biggr)
+    
\sqrt{\frac{2V\tau r^\vt}{n}} \\
&\quad +
\Bigl(\frac{2 V\tau}n\Bigr)^{\frac 1{2-\vt}}     
+   \frac {8B\tau}{3n} + 2\, |L|_{M,1} \cdot \e 
\end{align*}
holds with probability $P^n$ not less than $1-(1+|\ca C|)e^{-\tau}$.
In the following, we fix a data set $D$, for which this inequality holds. 
Defining 
$$
r:= \Bigl(\frac {16V\tau}n \Bigr)^{1/(2-\vt)}\, , 
$$
 a simple calculation then shows 
both
$$
\sqrt{\frac{2V\tau }{nr^{2-\vt}}} = \frac 1 {\sqrt 8}
\qquad \mbox{ and } \qquad 
\sqrt{\frac{2V\tau r^\vt }{n}} = \frac r{\sqrt 8}\, .
$$
Moreover,  $V\geq B^{2-\vt}$ together with  $n\geq 16\tau$ gives
\[ 
\frac {4B\tau}{3n r}
=
\frac 1 {12} \cdot \frac {16\tau}{n} \cdot \frac B r 
\Leq
\frac 1 {12} \cdot \Bigl(\frac {16\tau}{n}\Bigr)^{\frac 1 {2-\vt}} \cdot \frac {V^{\frac 1 {2-\vt}}} r 
=
\frac 1 {12}
\qquad
\mbox{ and }
\qquad 
\frac {8B\tau}{3n} \leq \frac r6\, .
\]
Finally, we have 
\begin{align*}
 \Bigl(\frac{2 V\tau}n\Bigr)^{\frac 1{2-\vt}} = 8^{-\frac 1{2-\vt}} \Bigl(\frac{16 V\tau}n\Bigr)^{\frac 1{2-\vt}} 
 \leq \frac r{\sqrt 8}\, .
\end{align*}
Inserting these estimates in our inequality on 
$\Ex_Ph_{f_D}$ gives
\begin{align*}
\Ex_P h_{f_D}   
&< 2 \Ex_P h_\fO +  \bigl(\Ex_P h_{f_D} + \e\bigr)\biggl(\frac 1 {\sqrt 8} + \frac 1{12}\biggr)
+    
\frac r {\sqrt 8}   
+
\frac r {\sqrt 8}     
+   \frac r6 + 2\, |L|_{M,1} \cdot\e \\
& =  2\Ex_P h_\fO + \frac {6+\sqrt 2}{12\sqrt 2} \cdot  \Ex_P h_{f_D}  
+  
\frac{6+\sqrt 2}{6\sqrt 2} \cdot r
+ \frac {6+ 25\sqrt 2}{12\sqrt 2} \cdot |L|_{M,1} \cdot \e \, ,
\end{align*}
and by elementary 
transformations we thus  conclude that
\begin{align*}
 \RP L {f_D} - \RPB L  
= \Ex_P h_{f_D}   
&< \frac{24\sqrt 2}{11\sqrt 2-6} \cdot  \Ex_P h_\fO + 
\frac{12+2\sqrt 2}{11\sqrt 2-6} \cdot r
+ \frac {6+ 25\sqrt 2}{11\sqrt 2-6} \cdot |L|_{M,1} \cdot\e \\
& <  4 \,\Ex_P h_\fO + 
2  r
+ 5 \, |L|_{M,1} \cdot\e \\
& = 4 \bigl( \RP L \fO - \RPB L   \bigr) + 
2\,  \Bigl(\frac {16V\tau}n \Bigr)^{1/(2-\vt)}
+ 5 \, |L|_{M,1} \cdot\e
\end{align*}
Now the assertion follows by a simple algebraic transformation of $\tau$
and taking the infimum over all $\fO\in \ca F$.
\end{proof}

If we have an upper bound on the covering numbers occurring in Theorem \ref{theo:generic-oracle},
then we can optimize the right hand side of its oracle inequality with respect to $\e$.
The following corollary executes this idea for the least squares loss and 
histogram rules that choose their cubic partitions in a certain, data-dependent way.

\begin{corollary}
\label{prop:oracle-partitioning}
Let $Y = [-M,M]$  and let $L$ be the least squares loss. 
% Assume the distribution $P$ on $X \times Y$ satisfies   
% the assumptions of Theorem \ref{theo:generic-oracle}.  
For $K<\infty$ and $A<\infty$ let $\cA_1, \dots,\cA_K$  be  finite Partitions of $X$, satisfying $|\cA_i|\leq A$ for any $i=1,\dots,K$. 
Moreover, let $D \mapsto h_{D, \ca A_D }$ be an algorithm, that first chooses a partition $\ca A_D$ from 
$\cA_1, \dots, \cA_K$  and then computes the corresponding 
 $\cA_D$-histogram. 
Then, for all $n\geq 1$ and $\tau >0$, we have 
\begin{align*}
% \sup_{i=1,...,K}(
\RP L {h_{D, \cA_D }} - \RPB L
&\leq  4\sup_{i=1,\dots,K}\bigl( \RPxB L {\cH_{\cA_i}}-\RPB L\bigr) \\
& \quad +
1024 \,\frac{ \tau M^2 }n  +  512 \,\frac{  A M^2  }n\left( 1+\ln \left(\frac{n}{A} \right) \right) 
\end{align*}
with probability $P^n$ not less than $1- Ke^{-\tau}$. 
%Here, the constant $c_{1/2,\vt}$ is given in \eqref{eq:cp}.
\end{corollary}

\begin{proof}[Proof of Corollary \ref{prop:oracle-partitioning}]
Since $L$ is the least squares loss, the assumptions
\eqref{sup-bound} and \eqref{var-bound} 
of Theorem \ref{theo:generic-oracle} are satisfied with $\vartheta  = 1$, $B=4M^2$, and $V=16M^2$. 
Moreover, our assumption $Y\subset [-M,M]$ ensures that $L$ is locally Lipschitz 
continuous with $|L|_{M,1} \leq 4M$.

Now, for a fixed $i\in \{1,\dots,K\}$ we  recall that the histogram rule $D\mapsto h_{D,\ca A_i}$
is an empirical risk minimizer over the hyptheses class 
% 
%  Recall the associated class of $\cA$-histograms 
\begin{equation*}
 \cH_{\cA_i} =\left\{  \sum_{j \in J}c_j \eins_{A_j}\;: \; c_j \in Y \right\} \;,
\end{equation*} 
where $\cA_i=(A_j)_{j \in J}$. Moreover, 
for any  $\e>0$, the $\e$-covering number of $\cH_{\cA_i}$ satisfies 
\begin{equation}
\label{eq:cov-histo}
  \cN(\cH_{\cA_i}, ||\cdot||_\infty, \e) \leq (2M/\e)^{|\cA_i|} \leq (2M/\e)^{A} \;. 
\end{equation}  
For   $n\geq 1$, $\tau>0$, and $\e>0$ Theorem \ref{theo:generic-oracle} thus gives 
\begin{eqnarray*}
\RP L {h_{D,\ca A_i}} - \RPB L 
\leq  4\bigl( \RPxB L {\cH_{\cA_i}}-\RPB L\bigr) + 20 M \cdot \e 
+
\frac{512 M^2\bigl(2\tau +  A \ln(\frac {2M} \e) \bigr)   }n 
\end{eqnarray*}
with probability $P^n$ not less than $1- e^{-\tau}$.

Next we optimize this bound over $\e>0$. To this end, we consider
the  strongly convex function 
\begin{align*}
 h(\e) = \alpha   \e + \beta \ln(\gamma/\e)\, , 
\end{align*}
where $\alpha:= 20 M$, $\beta := \frac{512AM^2}{n}$, and $\gamma:= 2M$.  
Then a simple calculation shows that $h$ has a minimum  at $\e^* := \frac \beta\alpha$,
giving 
\begin{align*}
  h(\e^*) 
  = 
  \beta \biggl(1+\ln\Bigl(\frac{\alpha \gamma}\beta\Bigr)   \biggr)
  =
   \frac{512AM^2}{n}\biggl(1+\ln\Bigl(\frac{40  n M^2}{512 AM^2}\Bigr)   \biggr)
 \leq 
   \frac{512AM^2}{n}\biggl(1+\ln\Bigl(\frac{ n}{ A}\Bigr)   \biggr) \, .
\end{align*}

Inserting this estimate in our above oracle inequality 
obtained from Theorem \ref{theo:generic-oracle},
using the fact that 
\begin{align*}
   \RP L {h_{D, \cA_D }} - \RPB L \leq \sup_{i=1,\dots,K} \RP L {h_{D,\ca A_i}} - \RPB L \, ,
\end{align*}
and finally applying a simple union bound then gives the assertion.
\end{proof}

%%%%%%%%%%%%%%%%%%%%%%%%%%%%%%%%%%%%%%%%%%%%%%%%%%%%%%%%%%%%%%%%%%%%%%%%%%%%%%%%%%%%%%%%%%%%%%%%%%%%%%%%%%%
%%%%%%%%%%%%%%%%%%%%%%%%%%%%%%%%%%%%%%%%%%%%%%%%%%%%%%%%%%%%%%%%%%%%%%%%%%%%%%%%%%%%%%%%%%%%%%%%%%%%%%%%%%%

\subsection{Learning Properties of Histograms}
\label{sec:uniform-histos}

The first lemma describes how well the infinite sample histogram rules  defined in \eqref{eq:inf-hist} can approximate the 
least squares Bayes risk.

\begin{lemma}[Approximation Error]
\label{lem:approx-error-reg}
Let $L$ be the least squares loss, $X:= [-1,1]^d$, $Y= [-1,1]$, and $P$ 
be a distribution on $X\times Y$. 
Then, for all 
$\e > 0$, there exists an $s_\e >0$ such that for 
any cubic partition $\ca A$ of $X$ with width $s \in (0, s_\e]$ one has 
\[   \cR_{L,P}(  h_{P,\cA}) - \cR_{L,P}^*  < \e \;.  \]
Moreover,  if $\fpb$ is $\alpha$- H\"older continuous for some $\alpha \in (0,1]$, then for all $s\in (0,1]$ 
and all cubic partitions $\ca A$ of $X$ with width $s$ 
we have 
\[ \cR_{L,P}(  h_{P,\cA}) - \cR_{L,P}^*  \leq  |\fpb|_\alpha^2\cdot  s^{2\alpha}\;.  \]
\end{lemma}

\begin{proofof}{Lemma \ref{lem:approx-error-reg}}
For the proof of the  first assertion we fix an $\e>0$. Then  recall that there exists a
continuous  function $f:\R^d\to \R$ with compact support such that 
\begin{align}\label{lem:approx-error-reg-h1}
   \snorm{\fpb - f}_2 \leq \e\, ,
\end{align}
see e.g.~\cite[Theorem 29.14 and Lemma 26.2]{Bauer01}. Moreover, since $\inorm \fpb \leq 1$,  we can assume
without loss of generality
 that $\inorm f\leq 1$.
Now, since $f$ is continuous and has compact support, $f$ is uniformly continuous, 
and hence there exists a $\delta \in (0,1]$ such that for all $x,x'\in X$ with $\inorm{x-x'}\leq \delta$ we have 
\begin{align}\label{lem:approx-error-reg-h2}
   \bigl|f(x)-f(x')\bigr| \leq \e\, .
\end{align}
We define  $s_\e := \delta$. Now,  we fix a cubic partition $\ca A = (A_j)_{j\in J}$ of width $s>0$ for some $s\in (0,s_\e]$.
For $x\in X$ with $P_X(A(x)) > 0$ we then have 
\begin{displaymath}
   h_{P,\cA}(x) = \frac{1}{P_X(A(x))} \int_{A(x)}f^*_{L,P}\,  dP_X  \, .
\end{displaymath}
For such $x$ we then define
\begin{align}\label{lem:approx-error-reg-hx3}
   \bar f(x): = \frac{1}{P_X(A(x))} \int_{A(x)}f \, dP_X  \, .
\end{align}
For the remaining $x\in X$ we simply set $\bar f(x) := 0$. With these preparations we then have 
\begin{align}\label{lem:approx-error-reg-h3}
   \snorm{h_{P,\cA} - \fpb}_2 \leq \snorm{h_{P,\cA} - \bar f}_2 + \snorm{\bar f-f}_2 + \snorm{f - \fpb }_2\, .
 \end{align}
Clearly, \eqref{lem:approx-error-reg-h1} shows that   the third term is bounded by $\e$.
Let us now consider the second term. Here we first note that for an $x\in X$ with $P_X(A(x)) > 0$ we have 
\begin{align} \nonumber
   \bigl|  f(x)-\bar f(x)\bigr|
& =  \frac{1}{P_X(A(x))} \biggl|\int_{A(x)}  f(x) - f(x') \, dP_X(x')  \biggr|  \\ \nonumber
& \leq \frac{1}{P_X(A(x))} \int_{A(x)} \bigl| f(x) - f(x') \bigr| \, dP_X(x')  \\ \label{lem:approx-error-reg-hx1}
& \leq \e\, ,
\end{align}
where in the last step we used \eqref{lem:approx-error-reg-h2}.
Consequently, we obtain 
\begin{align}\label{lem:approx-error-reg-hx2}
   \snorm{f-\bar f}_2^2   \nonumber
&= \sum_{j\in J: P_X(A_j) > 0}\, \int_{A_j}\bigl|  f(x)-\bar f(x)\bigr|^2 \, dP_X(x') \\ \nonumber
&\leq   \sum_{j\in J: P_X(A_j) > 0} \e^2 \cdot P_X(A_j) \\
&\leq \e^2 \, .
\end{align}
In other words, the second term is bounded by $\e$, too. Let us finally consider the first term.
Here we have
\begin{align*}
\snorm{h_{P,\cA} - \bar f}_2^2
&=  \sum_{j\in J: P_X(A_j) > 0} \int_{A_j} \bigl|  h_{P,\cA} - \bar f\bigr|^2 \, dP_X \\
&= \sum_{j\in J: P_X(A_j) > 0} \int_{A_j} \biggl|    \frac{1}{P_X(A_j)} \int_{A_j}\fpb \, dP_X  -  \frac{1}{P_X(A_j)} \int_{A_j}f \, dP_X  \biggr|^2   \, dP_X \\
&=  \sum_{j\in J: P_X(A_j) > 0} \,\, \biggl|   \int_{A_j}  \fpb \, dP_X  - \int_{A_j}  f \, dP_X   \, dP_X   \biggr|^2\\
&\leq \Biggl( \sum_{j\in J: P_X(A_j) > 0} \,\, \biggl|   \int_{A_j}  \fpb \, dP_X  - \int_{A_j}  f \, dP_X   \, dP_X   \biggr|  \Biggr)^2\\
&\leq \Biggl( \sum_{j\in J: P_X(A_j) > 0} \,\,    \int_{A_j} \bigl| \fpb - f\bigr| \, dP_X    \Biggr)^2\\
% &\leq  \biggl|\sum_{j\in J: P_X(A_j) > 0}     \int_{A_j}  \fpb \, dP_X  - \int_{A_j}  f \, dP_X   \, dP_X   \biggr|^2\\
& =\snorm{\fpb -f}_1^2 \\
&\leq  \snorm{\fpb -f}_2^2 \\
&\leq \e^2 \, .
\end{align*}
Consequently, the first term is bounded by $\e$, too, and hence we conclude by 
\eqref{lem:approx-error-reg-h3} that the excess risk satisfies 
\begin{displaymath}
   \RP L {h_{P,\cA}} - \RPB L =  \snorm{h_{P,\cA} - \fpb}_2^2 \leq 9\e^2\, .
\end{displaymath}
A simple variable transformation then yields the first assertion. 

To show the second assertion we first note that for all $x,x'\in X$ we have 
\begin{align*}
  | \fpb(x)  - \fpb(x')| \leq \holdernorm \fpb \cdot ||x-x'||_\infty^\alpha\, .
\end{align*}
For $s\in (0,1]$, $\e :=  \holdernorm \fpb \cdot s^\alpha$,  and $x,x'\in X$ with $||x-x'||_\infty\leq s$ we thus find 
\begin{align*}
  | \fpb(x)  - \fpb(x')| \leq     \e  \, .
\end{align*}
Now consider  $f: = \fpb$ and fix an  arbitrary cubic partition $\ca A$ of $X$ with width $s$.
Then $\bar f$ defined by \eqref{lem:approx-error-reg-hx3} is given by $\bar f = h_{P,\cA}$.
Moreover, we have 
\begin{displaymath}
 \RP L {h_{P,\cA}} - \RPB L =  \snorm{ \fpb - h_{P,\cA}}^2_2 \leq \e^2 \, ,
\end{displaymath}
where in the last step we used  \eqref{lem:approx-error-reg-hx1} and \eqref{lem:approx-error-reg-hx2}.
\end{proofof}

Based on the previous results  
we can now establish universal consistency of the empirical histogram rule $D \mapsto h_{D,\cA_D}$ for regression 
based on a cubic data-dependent partition $\cA_D$ from $\cP(X)$.

\begin{proposition}[Universal Consistency]
\label{prop:cons-reg}
Let $L$ be the least squares loss, $X:= [-1,1]^d$, $Y= [-1,1]$, $P$ 
be a distribution on $X\times Y$, and $D \in (X \times Y)^n$ be an i.i.d. sample of size $n \geq 1$ drawn from $P$ with $|D_X|=m_n$. 
Suppose that $(s_n)_{n \in \mbn}$ is a sequence with 
$s_n \to 0$ as well as $\frac{\ln (n s_n^d)}{n s_n^d}\to 0$ as $n \to \infty$. 
%\fix{\notice{Cor. \ref{prop:oracle-partitioning} shows: instead of $\frac{\log(n)}{ns_n^d} \to 0 $ we need now $ns_n^d \to \infty $}}
Assume further that $\pi_{m_n, s_n}$ is an $m_n$-sample cubic partitioning rule of width $s_n \in (0,1]$, satisfying  
$|\Image(\pi_{m_n, s_n})| \leq c n^\beta $, for  some $c<\infty$ and some $\beta>0$
that are independent of $n$.
Denoting $\cA_D:=\pi_{m_n, s_n}(D_X)$, we have  
\[   \RP{\Lc}{h_{D,\cA_D}} \to \RxB{\Lc}{P}  \]
in probability as $n\to \infty$. 
\end{proposition}

\begin{proofof}{Proposition \ref{prop:cons-reg}}
Note that for any  $\e>0$ and for any $\cA \in \Image(\pi_{m_n, s_n})$, the $\e$-covering number of $\cH_\cA$ satisfies 
\begin{equation}
\label{eq:cov-numb}
  \cN(\cH_\cA, ||\cdot||_\infty, \e) \leq (2/\e)^{|\cA|} \;,  
\end{equation}  
with $|\cA|\leq (2/s_n)^d$. 
Let us write $\ca P_n:= \Image(\pi_{1, s_n}) \cup \dots\cup \Image(\pi_{n, s_n})$.
Applying Corollary \ref{prop:oracle-partitioning} 
with $A:= (2/s_n)^d$ and $K:= |\ca P_n| \leq c n^{1+\beta}$ 
gives, for all $\tau \geq 1$ and $n\geq 1$, with probability $P^n$ at least $1-2c n^{1+\beta} e^{-\tau}$ that 
\begin{align*}
\RP{\Lc}{h_{D,\cA_D}} - \RxB{\Lc}{P} 
% &\leq \sup_{\cA \in \ca P_n} \cR_{L,P}(h_{D, \cA}) - \cR_{L,P}^*  \no\\
&\leq 
4\sup_{\cA \in \ca P_n}\bigl( \RPxB L {\cH_{\cA}}-\RPB L\bigr) \no  \\
&  \;\;\; + 1024 \,\frac{ \tau  }n  +  512 \,\frac{  2^d  }{n s_n^d}\left( 1+\ln \left(\frac{n s_n^d}{2^d} \right) \right) \;.
\end{align*}
Now, for all $\e>0$, 
Lemma \ref{lem:approx-error-reg} guarantees the existence of an $s_\e >0$ 
such that for any cubic partition  $\cA$ of width 
$s_n \in (0, s_\e]$ we have 
\begin{align}
\label{eq:approx-reg}
  \cR_{L,P}(  h_{P,\cA}) -  \cR_{L,P}^* & < \e \;.
\end{align} 
Since we assumed $s_n \to 0$ we conclude that the latter inequality 
holds for all sufficiently large $n$.
Combining both bounds we find  for all sufficiently large $n$ that  with probability  $P^n$ at least $1-2c n^{1+\beta} e^{-\tau}$ it holds
\begin{align*}
\RP{\Lc}{h_{D,\cA_D}} - \RxB{\Lc}{P} \leq 
4\e +  1024 \,\frac{ \tau  }n  + c_d \, \frac{\ln (n s_n^d)}{n s_n^d} \;,
\end{align*}
where $c_d=1024 \cdot 2^d$. Finally, choosing 
$\tau = (\beta + 2) \log(n)$, 
the result follows by remembering that by assumption
$\frac{\ln (n s_n^d)}{n s_n^d} \to 0$.
\end{proofof}

We now come to our second main contribution of this section, namely the derivation of learning rates for  
the empirical histogram rule $D \mapsto h_{D,\cA_D}$ for regression 
based on a cubic data-dependent partition $\cA_D$ from $\cP(X)$.

\begin{proposition}[Learning Rates]
\label{prop:rates-reg}
Let $L$ be the least squares loss, $X:= [-1,1]^d$, $Y= [-1,1]$, $P$ 
be a distribution on $X\times Y$, and $D \in (X \times Y)^n$ be an i.i.d. sample of size $n \geq 1$ drawn from $P$ with $|D_X|=m_n$. 
Assume the Bayes decision function $f^*_{L,P}$ is $\alpha$-H\"older continuous for some $\alpha \in (0,1]$. 
Suppose further  that $(s_n)_{n \in \mbn}$ is a sequence satisfying 
%$(s_n)_{n \in \mbn}$ be a sequence chosen according to 
\[  s_n = n^{-\gamma}\;, \quad \gamma = \frac{1}{2\alpha + d}   \;.\] 
Assume further that $\pi_{m_n, s_n}$ is an $m_n$-sample cubic partitioning rule of width $s_n \in (0,1]$, satisfying  
$|\Image(\pi_{m_n, s_n})| \leq c n^\beta $, for  some $c<\infty$ and some $\beta>0$
that are independent of $n$.
 Denoting $\cA_D:=\pi_{m_n, s_n}(D_X)$,  the excess risk then satisfies 
for all  $n\geq 1$ the inequality
 \[   \RP{\Lc}{h_{D,\cA_D}} - \RxB{\Lc}{P} \leq c_{d,\alpha }\ln(n  ) \left( \frac 1 n \right)^{2\alpha \gamma } \]
with probability $P^n$ at least $1- cn^{1+\beta} e^{-n^{d \gamma}}$, where 
$c_{d,\alpha }>0$ is a constant only depending on $d$, $\alpha$, and $|f^*_{L,P}|_\alpha$.
% , and with $\gamma' = \frac{d}{2\alpha + d}$. 
\end{proposition}

\begin{proofof}{Proposition \ref{prop:rates-reg}}  
If the Bayes decision function $f^*_{L,P}$ is $\alpha$-H\"older continuous, Lemma \ref{lem:approx-error-reg} gives us for all $n\geq1$ that 
\begin{align}\label{eq:approx-reg-2}
\cR_{L,P}(  h_{P,\cA}) - \cR_{L,P}^*  < |f^*_{L,P}|_\alpha s_n^{2\alpha}\;.
\end{align}
Repeating the proof  of Proposition \ref{prop:cons-reg} by replacing 
\eqref{eq:approx-reg} with \eqref{eq:approx-reg-2}
shows that for  all $\tau \geq 1$ and  $n\geq 1$ we have 
\begin{align}
\label{eq:rate}
\RP{\Lc}{h_{D,\cA_D}} - \RxB{\Lc}{P} &\leq 4|f^*_{L,P}|_\alpha s_n^{2\alpha}  + 1024 \,\frac{ \tau  }n  + 1024\cdot 2^d \, \frac{\ln (n s_n^d)}{n s_n^d}  \;.
\end{align}
with probability $P^n$ not less than $1-2cn^{1+\beta} e^{-\tau} $.
Using the definition of $s_n$ and setting $\tau_n: = ns_n^{2\alpha } = n^{\frac{d}{2\alpha + d}}$
then gives the assertion.
\end{proofof}

%\end{appendix}

%%%%%%%%%% FINAL CHECKS %%%%%%%%%%%%%%%%%%%%%%%%%%%%%%%%%%%%%%%%%%%%%%

\checknbnotes
%\checknbdrafts

%%%% GOOD JOB !

\end{document}